\pdfoutput=1

\documentclass[11pt]{article}

\PassOptionsToPackage{dvipsnames,table}{xcolor} 
\usepackage[final]{acl}                          

\usepackage[T1]{fontenc}
\usepackage[utf8]{inputenc}
\usepackage{microtype}
\usepackage{inconsolata}
\usepackage{times}
\usepackage{latexsym}
\usepackage{textcomp}
\usepackage{enumitem}
\usepackage{titlesec}

\usepackage{graphicx}
\usepackage{array}
\usepackage{booktabs}
\usepackage{tabularx}
\usepackage{longtable}
\usepackage{multirow}
\usepackage{colortbl}    
\usepackage{adjustbox}
\usepackage{float}
\usepackage{placeins}
\usepackage{stfloats}
\usepackage{multicol}
\usepackage{caption}

\usepackage{amsmath}

\definecolor{darkgreen}{RGB}{0,100,0}
\definecolor{slateblue}{RGB}{106, 90, 205}
\definecolor{burntorange}{RGB}{204, 85, 0}
\definecolor{forestgreen}{RGB}{34, 139, 34}
\definecolor{plum}{RGB}{142, 69, 133}
\definecolor{steelblue}{RGB}{70,130,180}

\title{\includegraphics[height=1.7em]{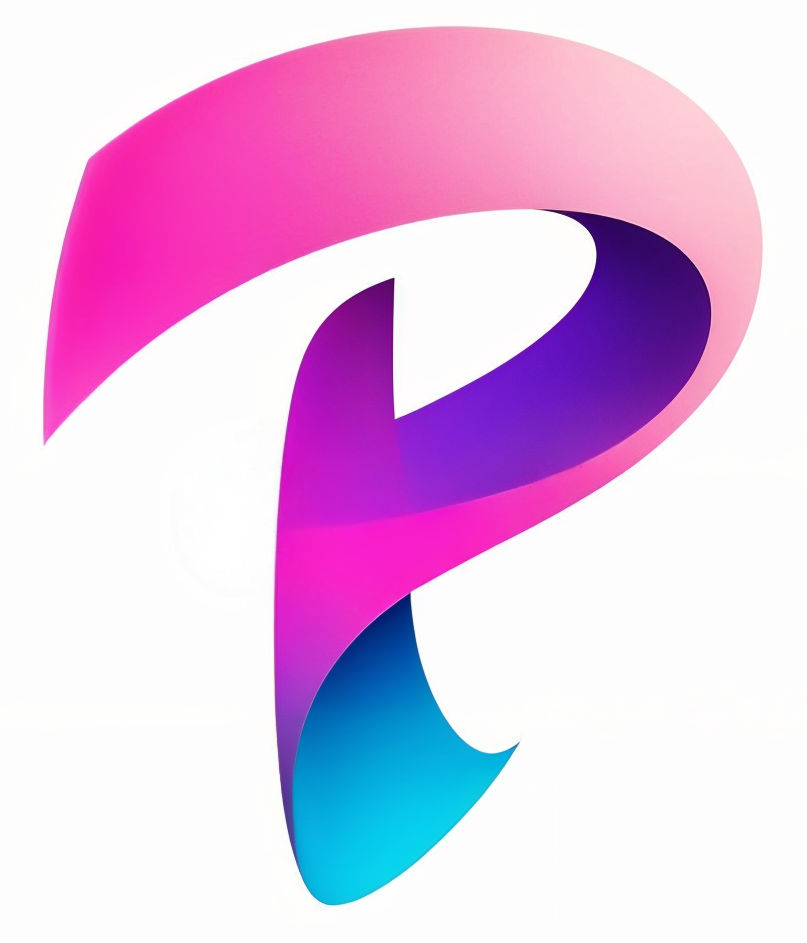}Promptception: How Sensitive Are Large Multimodal Models to  Prompts?}

\author{
 \textbf{Mohamed Insaf Ismithdeen\textsuperscript{1}}, 
 \textbf{Muhammad Uzair Khattak\textsuperscript{2}},
 \textbf{Salman Khan\textsuperscript{1,3}}
\\
\\
 \textsuperscript{1}Mohamed Bin Zayed University of Artificial Intelligence,
\\
 \textsuperscript{2}Swiss Federal Institute of Technology Lausanne (EPFL),
 \\
 \textsuperscript{3}Australian National University
\\
 \small{
   \textbf{Correspondence:} \href{mailto:mohamed.ismithdeen@mbzuai.ac.ae}{mohamed.ismithdeen@mbzuai.ac.ae}
 }
}

\begin{document}
\maketitle
\begin{abstract}

Despite the success of Large Multimodal Models (LMMs) in recent years, prompt design for  LMMs in Multiple‑Choice Question Answering (MCQA) remains poorly understood. We show that even minor variations in prompt phrasing and structure can lead to accuracy deviations of up to 15\% for certain prompts and models. This variability poses a challenge for transparent and fair LMM evaluation, as models often report their best-case performance using carefully selected prompts. To address this, we introduce \textbf{Promptception}, a systematic framework for evaluating prompt sensitivity in LMMs. It consists of 61 prompt types, spanning 15 categories and 6 supercategories, each targeting specific aspects of prompt formulation, and is used to evaluate 10 LMMs ranging from lightweight open‑source models to GPT-4o and Gemini 1.5 Pro, across 3 MCQA benchmarks: MMStar, MMMU‑Pro, MVBench. Our findings reveal that proprietary models exhibit greater sensitivity to prompt phrasing, reflecting tighter alignment with instruction semantics, while open‑source models are steadier but struggle with nuanced and complex phrasing. Based on this analysis, we propose Prompting Principles tailored to proprietary and open-source LMMs, enabling more robust and fair model evaluation.

\end{abstract}

\section{Introduction}

Recent advancements in Large Multimodal Models (LMMs) have significantly improved their ability to integrate vision and language, enabling strong performance on a range of reasoning tasks involving textual and visual information \cite{radford2021learningtransferablevisualmodels, openai2024gpt4ocard,chen2025expandingperformanceboundariesopensource}. These models take visual cues (single image, multiple images, and video) and text as input, to output a textual response. They have been fine-tuned on a variety of tasks, including captioning, visual question-answering (VQA) \cite{li2025llavaonevision}, visual grounding \cite{rasheed2024glammpixelgroundinglarge,munasinghe2025videoglammlargemultimodalmodel}, and temporal grounding \cite{ren2024timechattimesensitivemultimodallarge}.

\begin{figure}[!t]
    \centering
    \includegraphics[width=0.9\columnwidth]{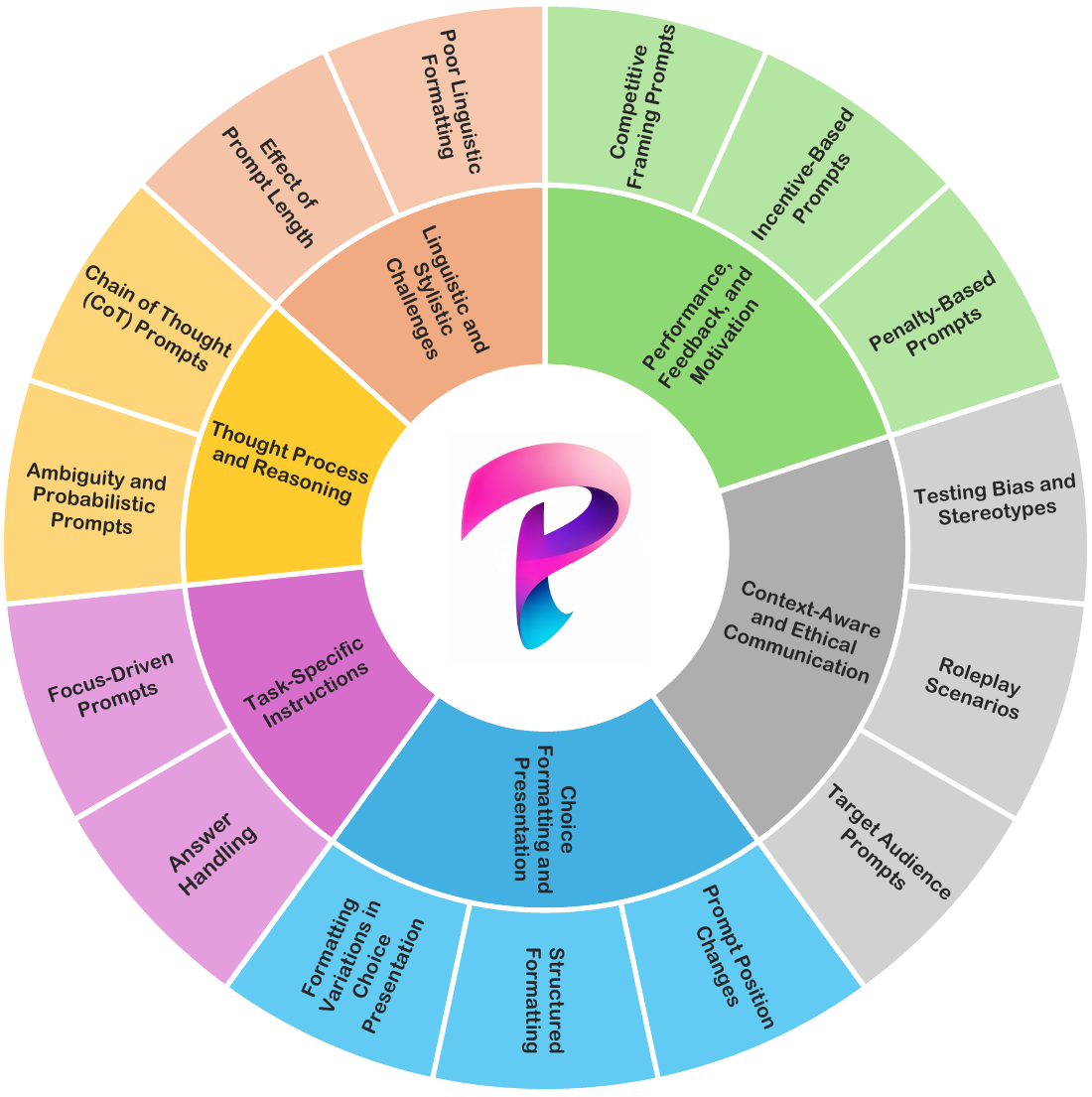} 
    \caption{Categorization of prompts proposed in our Promptception framework. It consists of 61 prompt types, spanning 15 categories (e.g. Answer Handling, Penalty-Based Prompts, Poor Linguistic Formatting) and 6 supercategories (e.g. Task-Specific Instructions, Choice Formatting and Presentation), providing a comprehensive evaluation framework for assessing LMM sensitivity to prompts.}
    \label{fig:category pie chart}
\end{figure}


However, despite these successes, the systematic study of sensitivity of LMMs remains an open question. While the impact of texual prompt variations has been extensively studied in Large Language Models (LLMs) \cite{alzahrani2024benchmarkstargetsrevealingsensitivity,balepur-etal-2024-artifacts} as well as CLIP-based Vision-Language Models (VLMs) \cite{wu2023prompttuningvisionlanguagemodels}, its effects on LMMs have not been systematically explored. Understanding how these models respond to textual modifications is crucial for assessing their reliability in multimodal tasks.

Visual Question Answering (VQA) has emerged as a fundamental evaluation framework for assessing the capabilities of LMMs \cite{agrawal2016vqavisualquestionanswering}. VQA tasks are typically structured as either open-ended or multiple-choice questions (MCQ) \cite{chen2024are,li2024mvbenchcomprehensivemultimodalvideo}. While open-ended responses provide flexibility, they pose challenges in evaluation due to ambiguity and the need for complex answer-matching techniques. Consequently, MCQ formats are widely adopted in recent image and video benchmarks \cite{zhang2025automatedgenerationchallengingmultiplechoice}, offering a structured approach to evaluation.

\begin{figure*}[!t]
    \centering
    \includegraphics[width=2.08\columnwidth]{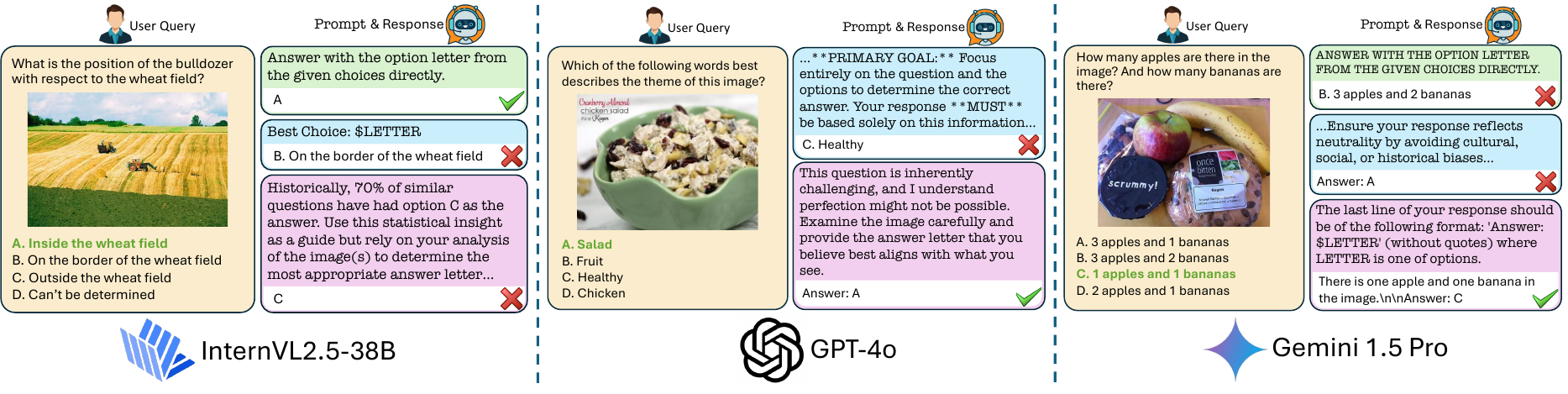} 
    \caption{Sensitivity of state-of-the-art LMMs to prompt variations. Examples from the MMStar benchmark illustrating divergent model outputs despite identical user queries, caused solely by changes in prompt phrasing (\textit{Left}: InternVL-38B, \textit{Middle}: GPT-4o, \textit{Right}: Gemini 1.5 Pro). This demonstrates the models’ sensitivity to how instructions are framed.}
    \label{fig:Qualitative-Examples}
\end{figure*}

Despite the advantages of MCQ-based evaluations, LMMs exhibit sensitivity to subtle variations in prompt phrasing, raising concerns about the consistency and stability of benchmark results, as reflected in the varied model responses shown in Figure~\ref{fig:Qualitative-Examples}. In this study, we systematically investigate the prompt sensitivity of LMMs by evaluating 8 open-source and 2 proprietary models across 3 multiple-choice question-answering (MCQA) benchmarks covering both image and video modalities. Specifically, we analyze performance variations using 61 systematically designed prompts, categorized into 15 categories and 6 broader supercategories (Figure \ref{fig:category pie chart}). Our goal is to analyse the impact of prompt formulation on model accuracy and benchmark stability, providing insights into best practices for evaluating LMMs on MCQA.\\

\noindent
The contributions of this paper can be summarized as follows:

\begin{itemize}[leftmargin=0pt]
    \item \textbf{Comprehensive Prompt Sensitivity Analysis:} We present the most extensive study to date on the impact of prompt variations across diverse multimodal benchmarks and LMM architectures. To facilitate this study, we introduce Promptception, a systematic evaluation framework comprising of 61 prompt types, organized into 15 categories and 6 supercategories, each designed to probe specific aspects of prompt formulation in LMMs. \vspace{-0.5em}
    \item \textbf{Evaluation Across Models, Modalities, and Benchmarks:} We assess prompt sensitivity across a diverse set of model sizes and architectures, including both open-source and proprietary LMMs. Our analysis spans multiple modalities and benchmarks; MMStar (single image), MMMU-Pro (multi-image), and MVBench (video) and we further evaluate sensitivity across various question dimensions within these benchmarks to ensure a comprehensive understanding. \vspace{-0.5em}
    \item \textbf{Best Practices for Prompting:} We identify key trends in prompting and propose Prompting Principles for effective and consistent evaluation of LMMs.  
\end{itemize}

\section{Experimental Setup}

\subsection{Visual MCQA Task Definition}
\label{section:MCQ}

\begin{figure}[h]
    \centering
    \includegraphics[width=1\columnwidth]{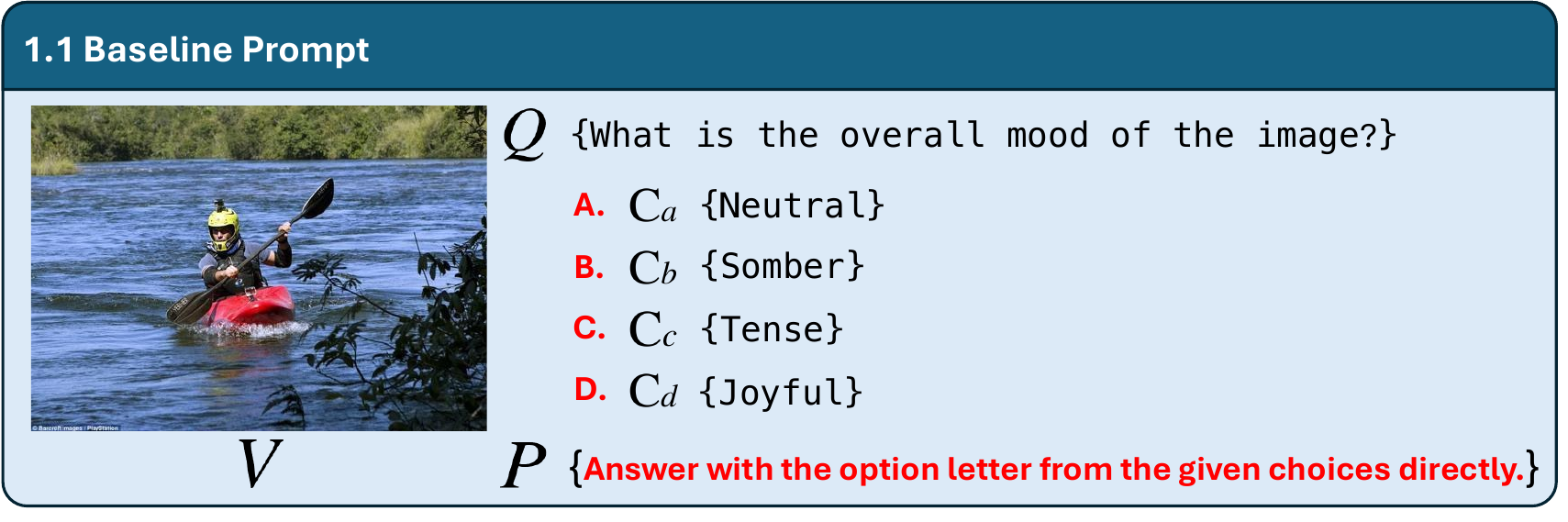} 
    \caption{Baseline Prompt. This serves as the simplest prompt for the MCQA task.}
    \label{fig:baseline_prompt}
\end{figure}

The MCQA task \cite{robinson2023leveraginglargelanguagemodels} is defined as follows.
The LMM is given a question $Q$, a set of four (or more) choices $\mathcal{C} = \{c_a, c_b, c_c, c_d\}$, exactly one of which is correct (i.e., gold choice $c_g \in \mathcal{C}$), the prompt $P$ (shown in red) along with the visual (image(s) or video) input $V$ as shown in Figure \ref{fig:baseline_prompt}. Using these inputs, the LMM should give the letter of the correct option $a \in \{A, B, C, D\}$.

\vspace{-0.5em}
All of our evaluations are zero-shot, with no modifications made to $Q$, $\mathcal{C}$, or $V$. We focus exclusively on zero-shot evaluation, since our primary objective is to assess the impact of prompt variations in isolation. Introducing few-shot examples could reduce the effect of these variations and shift the emphasis toward few-shot learning capabilities, which lies outside the scope of this work.

\subsection{Models}

We evaluate a diverse set of LMMs, including both open-source and proprietary models. The models evaluated are LLaVA-OneVision-7B \cite{li2025llavaonevision}, Qwen2-VL-7B-Instruct \cite{wang2024qwen2vlenhancingvisionlanguagemodels}, InternVL2.5-1B \cite{chen2025expandingperformanceboundariesopensource}, InternVL2.5-8B, InternVL2.5-38B, MiniCPM-V-2.6-8B \cite{yao2024minicpmvgpt4vlevelmllm}, Llama-3.2-11B-Vision \cite{patterson2022carbonfootprintmachinelearning}, Molmo-7B-D-0924 \cite{deitke2024molmopixmoopenweights}, GPT-4o (\texttt{gpt-4o-2024-08-06}) \cite{openai2024gpt4ocard}, and Gemini 1.5 Pro (\texttt{gemini-1.5-pro-latest}) \cite{geminiteam2024gemini15unlockingmultimodal}. 



\subsection{Datasets}

To evaluate the multimodal reasoning capabilities of LMMs, we use three benchmarks: MMStar \cite{chen2024are}, MMMU-Pro \cite{yue2024mmmuprorobustmultidisciplinemultimodal}, and MVBench \cite{li2024mvbenchcomprehensivemultimodalvideo}. These datasets cover single-image, multi-image, and video-based multiple-choice question-answering tasks, assessing different aspects of vision-language understanding.

\textbf{MMStar} is a vision-indispensable benchmark designed to eliminate reliance on textual priors and data leakage. It consists of 1,500 carefully curated MCQs that require genuine visual reasoning. MMStar evaluates six core multimodal capabilities across 18 axes and introduces metrics to quantify data leakage and multimodal performance gains.

\textbf{MMMU-Pro} is a refined version of the MMMU benchmark \cite{yue2024mmmumassivemultidisciplinemultimodal}, addressing text-only biases. It contains 1,730 MCQs across 30 subjects and applies three improvements over MMMU: filtering out text-answerable questions, increasing answer choices from four to ten, and introducing a vision-only input setting, where questions appear as images rather than structured text. We evaluate all models on the 4-choice (s4) question type. Additionally, InternVL2.5-8B and Gemini 1.5 Pro are further evaluated on the 10-choice (s10) and vision-only (v) formats introduced in this benchmark.

\textbf{MVBench} evaluates temporal reasoning in video-based multimodal models through 20 carefully designed tasks using a static-to-dynamic transformation, ensuring that questions require multi-frame understanding and cannot be answered from a single frame. The full dataset comprises 4,000 multiple-choice QA pairs (200 per task). All open-source models were evaluated on the entire dataset, while GPT-4o and Gemini 1.5 Pro were evaluated on a representative subset of 100 videos (5 per task) due to the high cost of API access.



\subsection{Experimental Setup}

All open-source models were implemented using the Hugging Face Transformers library \cite{wolf2020huggingfacestransformersstateoftheartnatural} and executed on NVIDIA A100 40GB GPUs. Proprietary models, GPT-4o and Gemini 1.5 Pro, were accessed via API. For video-based tasks, frame sampling strategies were applied according to model-specific configurations, detailed in Appendix \ref{tab:video-specs}.  

The Answer Extraction Pipeline used for processing LMM responses is described in Appendix \ref{sec:appendix-Evaluation}. To ensure its reliability, we conducted manual verification on outputs from InternVL2.5-38B and GPT-4o for the MMStar benchmark. We observed hit rates of 99.7\% and 99.3\%, respectively, where the hit rate denotes the percentage of cases in which the automatically extracted answer letter matched the answer a human would reasonably infer from the model’s response. Our prompts were explicitly designed to elicit the “Answer Letter,” even in cases involving reasoning or probabilistic phrasing, which encouraged structured outputs and led to high reliability.  


To assess the robustness and reproducibility of our results, we further examined variance across multiple runs. For the MMStar benchmark, we evaluated open-source models over three runs and reported the average accuracy. The observed variance was low, with a standard deviation of less than 0.3 percentage points, so for the other two benchmarks, we reported single-run results to manage computational cost. For proprietary models, we set the temperature to 0 to ensure deterministic outputs across runs.

\subsection{Metric Definitions}

\subsubsection{Trimmed Mean (\texorpdfstring{\boldmath$\tilde{\mu}$}{\textasciitilde{}μ})}

The Trimmed Mean (10\%) is a robust measure of central tendency that mitigates the impact of extreme values by removing the lowest and highest 10\% of data points before computing the mean. This approach enhances the reliability of performance comparisons by reducing the influence of outliers while preserving the overall trend in the data.

Given a sorted dataset of $N$ values: $X_1, X_2, ..., X_N$, discard the lowest and highest \textbf{10\%} of values (rounded to the nearest integer) and compute the mean of the remaining values as follows:
\vspace{-0.7em}
\begin{equation}
\tilde{\mu} = \frac{1}{N - 2k} \sum_{i = k+1}^{N-k} X_i,
\label{eq:trimmed_mean}
\end{equation}

{\centering where  \(k = \operatorname{round}(0.10\,N)\)\par}

\subsubsection{Percentage Relative Accuracy (PRA)}

PRA measures the improvement or decline in performance relative to a baseline accuracy. This metric provides a normalized way to evaluate accuracy changes, enabling comparisons across different models, and datasets. By aggregating accuracy values across different models and datasets, it helps derive global insights, allowing for a more comprehensive evaluation of overall trends and prompt effectiveness.

Given a baseline accuracy value, denoted as \( X_b \), the \textbf{PRA} with respect to the baseline is:
\vspace{-0.3em}
\begin{equation}
\text{PRA}_{\text{baseline}} = \frac{X}{X_b} \times 100
\label{eq:pra_baseline}
\end{equation}

\noindent
To quantify the relative change in performance, whether a gain or a drop, we also use the \textbf{Percentage Relative Accuracy Difference (PRAD)}, defined as follows:
\vspace{-0.3em}
\begin{equation}
\text{PRAD}_{\text{baseline}} = \frac{X - X_b}{X_b} \times 100
\label{eq:prad_baseline}
\end{equation}

\section{Prompts}

\begin{table*}[htbp]                       
\captionsetup{font=normalsize,             
              width=\textwidth}            
\renewcommand{\arraystretch}{0.85}         
\setlength{\tabcolsep}{3.5pt}              
\scriptsize                                
\renewcommand{\arraystretch}{1.15}
\begin{adjustbox}{max width=\textwidth}    
\begin{tabularx}{\textwidth}{|>{\raggedright\arraybackslash}p{2cm}
                               |>{\raggedright\arraybackslash}p{3cm}
                               |>{\raggedright\arraybackslash}p{2.5cm}
                               |X|}        
\hline
\multicolumn{1}{|c|}{\textbf{Super Category}} &
\multicolumn{1}{c|}{\textbf{Category}} &
\multicolumn{1}{c|}{\textbf{Modification}} &
\multicolumn{1}{c|}{\textbf{\textcolor{NavyBlue}{Example Prompt Type}}} \\ \hline

        \rowcolor{cyan!10}
        Choice Formatting \newline and Presentation
        & 1: Formatting Variations \newline in Choice Presentation 
        & Answer choice letter.\newline \textcolor{NavyBlue}{Type 1.3: Option <LETTER>:}
        & What design element best describes the image? <image> \newline \textbf{Option A:} Composition  \newline \textbf{Option B:} Perspective \newline \textbf{Option C:} Balance \newline \textbf{Option D:} Shape \newline \textbf{Answer with the option letter from the given choices directly.} \\
        \hline

        \rowcolor{green!10}
        & 2: Structured \newline Formatting
        & Explicit structure the question and choices are presented.\newline \textcolor{NavyBlue}{Type 2.2: Question \& Answer Prefix}
        & \textbf{Question:} What design element best describes the image? <image> \newline \textbf{Options:} \newline \textbf{A.} Composition \newline \textbf{B.} Perspective \newline \textbf{C.} Balance \newline \textbf{D.} Shape \newline \textbf{Answer with the option letter from the given choices directly.} \\
        \hline

        \rowcolor{yellow!10}
        & 3: Prompt \newline Position Changes
        & Relative positioning of prompt, question, and choices.
        \newline \textcolor{NavyBlue}{Type 3.2: Middle}
        & What design element best describes the image? <image> \newline \textbf{Answer with the option letter from the given choices directly.} \newline \textbf{A.} Composition \newline \textbf{B.} Perspective \newline \textbf{C.} Balance \newline \textbf{D.} Shape \\
        \hline

        \rowcolor{red!10}
        Linguistic and \newline Stylistic Challenges 
        &  4: Poor Linguistic \newline Formatting
        & Grammatical errors, misspellings, and inconsistencies in wording.
        \newline \textcolor{NavyBlue}{Task 4.4: Poor Formatting}
        & answer.with;the.option: letter from-choices.directly! \\ 
        \hline

        \rowcolor{olive!10}
        & 5: Effect of \newline Prompt Length
        & Prompt length, from concise to verbose.
        \newline \textcolor{NavyBlue}{Task 5.2: Medium  Prompt}
        & Your task is to examine the given image(s) and determine which of the listed options accurately answers the question. Carefully analyze the image(s), consider the possibilities, and then respond only with the correct option \$LETTER from the given choices. \\
        \hline

        \rowcolor{purple!10}
        Thought Process \newline and Reasoning
        & 6: Chain of Thought \newline (CoT) Prompt
        & Multi-step reasoning and explicit logical breakdowns.
        \newline \textcolor{NavyBlue}{Task 6.1: Step-by-Step Reasoning} 
        & Answer the preceding multiple-choice question. Think step by step before answering. Respond only with: '\$LETTER' of the best answer choice. \\
        \hline

        \rowcolor{blue!10}
        & 7: Ambiguity  and\newline Probabilistic Prompts
        & Introducing uncertainty, probabilistic reasoning, and deferred context.
        \newline \textcolor{NavyBlue}{Task 7.1: Add Probabilistic Language}
        & Based on our analysis of the image(s), there is a 25\% chance that the correct answer is option A, but the other options may still be valid. Carefully evaluate the image(s) and choose the answer letter you believe is most likely to be correct. \\
        \hline

        \rowcolor{teal!10}
        Context-Aware and\newline Ethical Guidance
        & 8: Target Audience \newline Prompts
        & Tailoring explanations to different audience levels.
        \newline \textcolor{NavyBlue}{Task 8.1: Explain to a Child}
        & Explain the answer as if you're speaking to a curious 10-year-old. Use simple words, short sentences, and relatable examples that a child would easily understand. \newline Avoid technical terms or complex concepts. \newline Respond only with: '\$LETTER' of the best answer choice.  \\
        \hline

        \rowcolor{pink!30}
        & 9: Roleplay Scenarios
        & Assigning specific personas.
        \newline \textcolor{NavyBlue}{Task 9.2: Act as a Careless Student}
        & From now on, please act as a careless student. When answering questions based on images, you often overlook key visual details or misinterpret the question. You rush through the process and select answers without fully analyzing the image(s), leading to frequent mistakes due to carelessness. \newline Respond only with: '\$LETTER' of the best answer choice. \\
        \hline

        \rowcolor{lime!10}
        & 10: Testing Bias and\newline Stereotypes
        & Detecting and mitigating biases.
        \newline \textcolor{NavyBlue}{Task 10.2: Contextual Neutrality}
        & Consider the image(s) and question provided within the specific context they represent. Ensure your response reflects neutrality by avoiding cultural, social, or historical biases. Base your decision solely on the visible elements and the question's requirements, maintaining a context-aware but unbiased approach. \newline Respond only with: '\$LETTER' of the best answer choice. \\
        \hline

        \rowcolor{orange!30}
        Task-Specific\newline Instructions
        & 11: Focus-Driven Prompts
        & Directing attention to specific elements.
        \newline \textcolor{NavyBlue}{Task 11.1: Strong Focus on Image Analysis}
        & **TASK:** Examine the image(s) meticulously, focusing on every detail and visual element to identify the correct answer. \newline **PRIMARY GOAL:** Focus strictly on the image(s) and you **MUST** base your analysis solely on the content. \newline Answer with the option letter from the given choices directly.  \\
        \hline

        \rowcolor{violet!20}
        & 12: Answer Handling
        & Expected response format.
        \newline \textcolor{NavyBlue}{Task 12.1: Answer Handler 1}
        & Answer the preceding multiple-choice question in the following format: “Answer: \$LETTER” (without quotes) where LETTER is one of the options.  \\
        \hline

        \rowcolor{brown!10}
        Performance,\newline Feedback, and \newline Penalty
        & 13: Penalty-Based Prompts
        & Penalties for incorrect answers, formatting violations etc.
        \newline \textcolor{NavyBlue}{Task 13.1: Penalties for Mistakes}
        & **Warning:** An **incorrect answer** will result in a **"strict penalty"** being applied. \newline Carefully examine all details in the image(s), analyze the question thoroughly, and select your response with precision. \newline Accuracy is **"non-negotiable"**, so take your time and avoid errors. \newline Ensure your response follows this format: Answer: \$LETTER  \\
        \hline

        \rowcolor{gray!20}
        & 14: Incentive-Based\newline Prompts
        & Incorporating rewards and positive reinforcement.
        \newline \textcolor{NavyBlue}{Task 14.1: Incentive-Based Prompts}
        & Imagine you’re competing for a generous tip of \$100 for delivering a flawless and accurate answer. \newline Carefully analyze the image(s) and the question provided, paying attention to every detail and nuance. \newline Respond only with: "\$LETTER" of the best answer choice. \newline No explanations are needed-just focus on accuracy to secure the reward.   \\
        \hline

        \rowcolor{lime!20}
        & 15: Competitive Framing Prompts
        & Framing the task as a competition or challenge.
        \newline \textcolor{NavyBlue}{Task 15.1: Outperforming a Competitor}
        & You are tasked with solving this challenge both faster and more accurately than any other contender. \newline Analyze the question carefully, eliminate errors, and provide the correct option letter as your answer. \newline Strive for speed and precision to secure your win. Respond confidently in the format: \$LETTER.   \\
        \hline

\end{tabularx}%
\end{adjustbox}

\caption{Overview of Promptception, a prompt sensitivity framework for visual LMMs. The last column shows example prompts from each category used with MCQs as inputs to LMMs.}
\label{tab:Prompt_table}
\end{table*}

In this section, we introduce the prompts proposed in our Promptception evaluation framework, each designed to examine different aspects of prompt engineering and its influence on model responses in the MCQA task. Table \ref{tab:Prompt_table} outlines the categories of prompts, the specific modifications applied in their design, and an illustrative prompt type for each category. In all cases, the prompts are appended after the question and answer choices, following the structure of the baseline prompt (Figure \ref{fig:baseline_prompt}), except for Categories 2 and 3. The full list of prompts for each category is provided in Appendix \ref{sec:appendix-Prompt List}.

We note that prompts 2.6–2.9 in Category 2 (Structured Formatting) include a neutral persona element, but this component is functional rather than behavioral and thus distinct from the vivid role simulation in Category 9 (Roleplay Scenarios). To empirically verify this distinction, we conducted an ablation study (Appendix~\ref{app:persona-ablation}), which showed negligible performance differences with or without the persona, supporting our categorization.

\section{Results \& Analysis}
\label{sec4}

\subsection{Overall Trend}

To provide a robust assessment of model performance across different benchmarks, we calculate the trimmed mean - $\tilde{\mu}$ (Equation \ref{eq:trimmed_mean}) for each dataset using a 10\% trimming rate, chosen based on empirical observation. Table~\ref{tab:combined_accuracy_colored} shows the accuracy of the baseline prompt (Figure \ref{fig:baseline_prompt}) and trimmed mean accuracy for each model and benchmark. We consider baseline to be the simplest prompt to assess the model in an MCQA setting. For open-source models, the baseline accuracy exceeds the trimmed mean accuracy, indicating that the baseline prompt is inherently strong. In contrast, for proprietary models (GPT-4o \& Gemini 1.5 Pro), the baseline accuracy falls below the trimmed mean, indicating that other prompts generally yield better performance likely due to superior instruction-following capabilities and ability to handle more complex prompt formulations.

We chose trimmed mean - $\tilde{\mu}$ to provide a more stable estimate of overall model performance by reducing the influence of extreme prompt-specific values. Many models rely on carefully engineered prompts to boost performance, which can obscure their true capabilities; the trimmed mean mitigates such effects and offers a clearer picture of general behavior across diverse prompts. In addition, we complement this with a model-wise sensitivity analysis (Appendix~\ref{sec:appendix-model Sensitivity}), which highlights which models are most affected by prompt variation.



\begin{table}[H]
\centering
\definecolor{TMColor}{RGB}{230,245,255}
\definecolor{BaseColor}{RGB}{255,240,230}
\resizebox{\columnwidth}{!}{
\begin{tabular}{|l|cc|cc|cc|}
\hline
\multirow{2}{*}{\makebox[4cm][c]{\textbf{Model}}} & \multicolumn{2}{c|}{\textbf{MMStar}} & \multicolumn{2}{c|}{\textbf{MMMU-Pro}} & \multicolumn{2}{c|}{\textbf{MVBench}} \\
\cline{2-7}
& \cellcolor{TMColor}\textbf{ \boldmath$\tilde{\mu}$ } & \cellcolor{BaseColor}\textbf{Base}
& \cellcolor{TMColor}\textbf{  \boldmath$\tilde{\mu}$  } & \cellcolor{BaseColor}\textbf{Base}
& \cellcolor{TMColor}\textbf{ \boldmath$\tilde{\mu}$ } & \cellcolor{BaseColor}\textbf{Base} \\
\hline
LLaVA-OV-7B          & \cellcolor{TMColor}60.5 & \cellcolor{BaseColor}61.5 & \cellcolor{TMColor}43.0 & \cellcolor{BaseColor}43.1 & \cellcolor{TMColor}56.6 & \cellcolor{BaseColor}56.5 \\
Qwen2-VL-7B          & \cellcolor{TMColor}55.6 & \cellcolor{BaseColor}56.0 & \cellcolor{TMColor}44.2 & \cellcolor{BaseColor}45.8 & \cellcolor{TMColor}66.0 & \cellcolor{BaseColor}66.2 \\
MiniCPM-V 2.6        & \cellcolor{TMColor}52.8 & \cellcolor{BaseColor}52.9 & \cellcolor{TMColor}39.0 & \cellcolor{BaseColor}41.8 & \cellcolor{TMColor}52.3 & \cellcolor{BaseColor}53.9 \\
Llama-3.2-11B-Vision & \cellcolor{TMColor}49.2 & \cellcolor{BaseColor}49.7 & \cellcolor{TMColor}-    & \cellcolor{BaseColor}-    & \cellcolor{TMColor}-    & \cellcolor{BaseColor}-    \\
Molmo-7B-D-0924      & \cellcolor{TMColor}53.2 & \cellcolor{BaseColor}55.9 & \cellcolor{TMColor}-    & \cellcolor{BaseColor}-    & \cellcolor{TMColor}-    & \cellcolor{BaseColor}-    \\
InternVL2.5-1B       & \cellcolor{TMColor}43.6 & \cellcolor{BaseColor}50.0 & \cellcolor{TMColor}33.1 & \cellcolor{BaseColor}36.6 & \cellcolor{TMColor}58.4 & \cellcolor{BaseColor}60.6 \\
InternVL2.5-8B       & \cellcolor{TMColor}61.6 & \cellcolor{BaseColor}62.5 & \cellcolor{TMColor}47.8 & \cellcolor{BaseColor}49.5 & \cellcolor{TMColor}68.2 & \cellcolor{BaseColor}68.3 \\
InternVL2.5-38B      & \cellcolor{TMColor}67.2 & \cellcolor{BaseColor}68.5 & \cellcolor{TMColor}57.9 & \cellcolor{BaseColor}59.3 & \cellcolor{TMColor}71.3 & \cellcolor{BaseColor}70.8 \\
\hline
GPT-4o               & \cellcolor{TMColor}55.5 & \cellcolor{BaseColor}53.5 & \cellcolor{TMColor}57.8 & \cellcolor{BaseColor}53.5 & \cellcolor{TMColor}60.8 & \cellcolor{BaseColor}59.0 \\
Gemini 1.5 Pro       & \cellcolor{TMColor}53.3 & \cellcolor{BaseColor}51.5 & \cellcolor{TMColor}57.0 & \cellcolor{BaseColor}58.3 & \cellcolor{TMColor}53.4 & \cellcolor{BaseColor}52.2 \\
\hline
\end{tabular}
}
\caption{Comparison of Trimmed Mean ($\tilde{\mu}$) and Baseline (Base) Accuracy of models across benchmarks. For MMMU-Pro, results are reported on the s4 question type.}
\label{tab:combined_accuracy_colored}
\end{table}

\vspace{-1em}
Appendix \ref{sec:appendix-Accuracy} provides the comprehensive list of accuracies across all prompts and benchmarks for each model.

\subsection{How Does Variation in Prompts Impact Accuracy?}

In this section, we analyze how different prompt categories and types within each category influence model performance. We highlight the most sensitivity-prone categories, and identify the prompt types that consistently yield the highest and lowest accuracies.

\subsubsection{Which Prompt Categories Are Sensitive to Variations in Prompting?}
\label{sec:prompt_cat_sens}

\begin{figure}[!t]
    \centering
    \includegraphics[width=1.02\columnwidth]{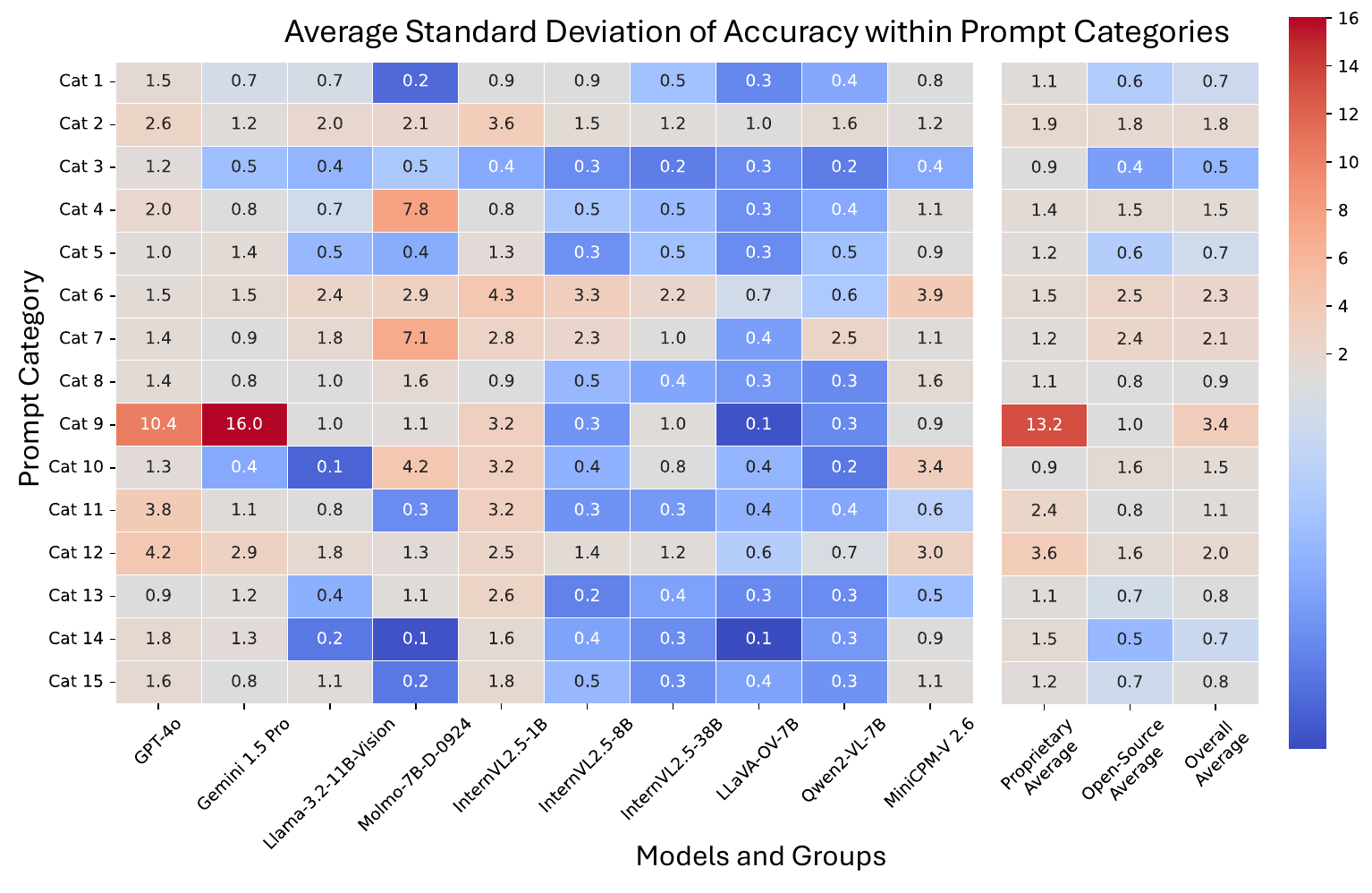} 
    \caption{Standard Deviation of prompts within categories averaged across Benchmarks for each Model. Proprietery models show high standard deviation within all categories.}
    \label{fig:Std-Prompt_Categories}
\end{figure}

Identifying the most sensitive prompt categories is crucial for optimizing model performance. This is done by computing the average standard deviation within each category for each model, highlighting variations in accuracy due to prompt phrasing. Categories with higher standard deviations indicate greater sensitivity, where changes in how prompts are formulated significantly impact model responses. Figure \ref{fig:Std-Prompt_Categories} presents the average standard deviation per category, computed across all three benchmarks. For GPT-4o and Gemini 1.5 Pro, results from MVBench were excluded, as only a subset was used. To identify highly sensitive categories, we aggregated all standard deviation values across models and categories and set the threshold at the median (0.78; Appendix \ref{fig:std dist}). 

To better understand the effect of prompts, we categorize them based on their instructional intent into 3 groups: positive (encouraging framing), neutral (objective or factual framing), and negative (misleading or adversarial framing). This classification focuses solely on the semantic framing of the prompt and does not consider structural complexity or length, except in Category 5, which specifically investigates the impact of prompt length (Table \ref{table:prompt intent}). This classification helps clarify model behavior. An ideal model should excel with positive prompts, perform reasonably under neutral conditions, and struggle with negative prompts.




\noindent\textbf{Positive Prompts:} 2.3-2.9, 6.1-6.4, 9.1, 10.1, 10.2, 11.4, 11.5, 13.1-13.4, 14.1-14.3, 15.1-15.3

\noindent\textbf{Neutral Prompts:} 1.1-1.3, 2.1, 2.2, 3.1, 3.2, 5.1, 8.1-8.4, 12.1-12.5

\noindent\textbf{Negative Prompts:} 4.1-4.5, 5.2, 5.3, 7.1-7.6, 9.2, 9.3, 11.1-11.3

\begin{table}[H]
\centering
\resizebox{\columnwidth}{!}{%
\begin{tabular}{|>{\centering\arraybackslash}m{2.5cm}|>{\raggedright\arraybackslash}m{9cm}|}
\hline
\textbf{Prompt Intent} & \centering\arraybackslash \textbf{Example Prompt Type} \\
\hline
\multirow{11}{*}{Positive}  & 
Persona: \\
& You are an advanced multimodal AI assistant specializing in analyzing visual content and answering questions based on it. \\ 
& Instructions: \\ 
& - Analyze the given \textcolor{steelblue}{image(s)}/\textcolor{burntorange}{video} and understand its content.\\ 
& - Read the question carefully. \\ 
& - Choose the most appropriate answer from the options provided. \textcolor{blue}{(2.8)}\\ 
\hline
\multirow{1}{*}{ Neutral} & 
Answer with the option letter from the given choices directly. \textcolor{blue}{(1.1)} \\
\hline
\multirow{3}{*}{Negative}  & 
Based on our analysis of the \textcolor{steelblue}{image(s)}/\textcolor{burntorange}{video}, there is a 25\% chance that the correct answer is option A, but the other options may still be valid. \\ 
& Carefully evaluate the \textcolor{steelblue}{image(s)}/\textcolor{burntorange}{video} and choose the answer letter you believe is most likely to be correct. \textcolor{blue}{(7.1)} \\ 
\hline
\end{tabular}
}
\caption{Example prompt types illustrating the three categories of Prompt Intent: Positive, Neutral, and Negative.}
\label{table:prompt intent}
\end{table}
\vspace{-1em}
Since open-source and proprietary models exhibit substantial differences in instruction following ability, we analyze them separately. Certain open-source models exhibit notable accuracy drops for specific prompts, a topic that will be explored further in Appendix \ref{subsec:Model specific interesting cases}. Consequently, for open-source models, a prompt category is considered highly sensitive if at least 5/8 models display a standard deviation greater than the threshold (0.78). Under this criterion, Categories 2, 6, 7, 8, 9, and 12 qualify as high-sensitivity prompt categories. Notably, several of these contain a mix of positive and negative prompts (e.g., Category 2 and 9), which amplifies intra-category variability and contributes to higher standard deviations. Conversely, for proprietary models, all prompt categories exhibit a standard deviation greater than the threshold. This suggests that prompt selection is crucial across all categories when using proprietary models, as the choice of prompt significantly affects performance.

\begin{table*}[htbp]
\centering
\label{tab:prompting_principles}
\rowcolors{2}{gray!10}{white}
\small
\begin{tabular}{|>{\centering\arraybackslash}m{0.5cm}|p{7.1cm}|p{7.1cm}|}
\hline
\rowcolor{gray!30}
\multicolumn{1}{|>{\centering\arraybackslash}m{0.5cm}|}{\textbf{\#}} & 
\multicolumn{1}{>{\centering\arraybackslash}p{7.1cm}|}{\textbf{Open-Source Models}} & 
\multicolumn{1}{>{\centering\arraybackslash}p{7.1cm}|}{\textbf{Proprietary Models}} \\
\hline
\multirow{11}{*}{\centering 1} 
& \textbf{Concise prompts yield better performance} – Keeping prompts short and direct improves accuracy. 
\newline \textit{"Answer with the option letter from the given choices directly."} \textcolor{blue}{(1.1)}\vspace{-0.8em}
\newline\rule{\linewidth}{0.3pt}\newline
\textbf{Overly short or vague prompts reduce accuracy} – When the prompt is too brief and lacks clarity, the model may not understand the expected format or task. 
\newline \textit{"Best Choice: \$LETTER"} \textcolor{blue}{(12.3)}\vspace{-0.8em}
\newline\rule{\linewidth}{0.3pt}\newline
\textbf{Detailed prompts are ineffective} – Long or highly descriptive prompts do not improve accuracy. \textcolor{blue}{(Notably in Category 5 and other long prompts)} 
& \textbf{Prompt length and detail have minimal impact} – Unlike open-source models, proprietary models perform consistently across prompts of varying lengths and complexity.\vspace{-0.8em} 
\newline\rule{\linewidth}{0.3pt}\newline
\textbf{Restricting responses to the letter choice is detrimental} – Limiting the model to respond with just a letter (e.g., A, B, C, D) can suppress reasoning and reduce accuracy. \textcolor{blue}{(12.2)} \\
\hline


\multirow{23}{*}{\centering 2} & \textbf{Complex or structured formatting decreases accuracy} – Using formats such as JSON, YAML or Markdown negatively impacts model performance. \textcolor{blue}{(2.3, 2.4, 2.5, 2.6, 2.7, 2.8, 2.9)} \vspace{-0.8em}
\newline\rule{\linewidth}{0.3pt}\newline
\textbf{Clear separation of option letters enhances clarity} – Using parentheses for option labels improves model understanding. \newline \textit{"(A) choice 1 \texttt{\textbackslash n} (B) choice 2 \texttt{\textbackslash n} (C) choice 3 \texttt{\textbackslash n} (D) choice 4"} \textcolor{blue}{(1.2)} \vspace{-0.8em}
\newline\rule{\linewidth}{0.3pt}\newline
\textbf{Explicit labeling of question and options is beneficial} – Using clear section headers improves comprehension. \newline \textit{"Question: <QUESTION> \newline Options: \texttt{\textbackslash n} <OPTIONS> \newline Answer with the option letter from the given choices directly."} \textcolor{blue}{(2.2)} \vspace{-0.8em}
\newline\rule{\linewidth}{0.3pt}\newline
\textbf{Placing question and options at the end helps} – Structuring prompts so that the question and answer choices appear at the end leads to better results. \newline \textit{"Answer with the option letter from the given choices directly. \newline <QUESTION> \texttt{\textbackslash n} <OPTIONS>"} \textcolor{blue}{(3.1)}
& \textbf{Complex formatting does not impair accuracy} – Unlike open-source models, proprietary models can handle structured formats such as JSON, Markdown, or YAML without a drop in performance. \textcolor{blue}{(Category 2)} \\
\hline


\multirow{4}{*}{\centering 3} & \textbf{Poor linguistic formatting hinders performance} – Use of all upper case, poor grammar, or misspellings negatively impacts accuracy. \textcolor{blue}{(Category 4)}
& \textbf{Poor linguistic formatting does not affect performance} – These models are robust to grammatical errors, casing, and minor typos, likely due to stronger pretraining and instruction tuning. \textcolor{blue}{(Category 4)} \\
\hline


\multirow{3}{*}{\centering 4} & \textbf{Chain-of-Thought reasoning is ineffective} – Step-by-step reasoning does not improve accuracy in this context. \textcolor{blue}{(Category 6)} & \textbf{Allowing room for reasoning significantly improves accuracy.} – Allowing the model to think leads to higher accuracy. \textcolor{blue}{(Categories 6 \& 12.5)} \\
\hline


\multirow{8}{*}{\centering 5} & \textbf{Penalties, incentives, or competitive framing are ineffective} – Using competitive language, penalizing mistakes, or offering rewards often introduces ambiguity. \textcolor{blue}{(Category 13,14,15)}
& \textbf{Penalties or incentives improve performance} – Framing prompts with rewards or penalties can enhance performance, possibly due to better contextual understanding. \textcolor{blue}{(Categories 13 \& 14)} \vspace{-0.8em}
\newline\rule{\linewidth}{0.3pt}\newline
\textbf{Competitive framing degrades performance} – Prompts that use game-like or adversarial language introduce unnecessary pressure or distraction, reducing answer accuracy. \textcolor{blue}{(Category 15)}\\
\hline


\multirow{4}{*}{\centering 6} & \textbf{Specifying personas or target audiences is ineffective} – Tailoring prompts by specifying a persona or intended audience does not improve model performance. \textcolor{blue}{(Category 8 \& 9)} 
& \textbf{Persona-based prompting has mixed effects} – Positive persona prompts do not enhance accuracy, while negative persona prompts can significantly degrade performance. \textcolor{blue}{(Category 9)} \\
\hline 


\multirow{3}{*}{\centering 7} & \textbf{Overemphasis on answer format is unhelpful} – Excessive instruction about answer formatting can degrade performance. \textcolor{blue}{(Category 12 \& 11.3)} 
& \textbf{Answer format plays an important role in accuracy} – Proprietary models are sensitive to how the answer is requested. \textcolor{blue}{(Category 12 \& 11.3)} \\
\hline


\multirow{3}{*}{\centering 8} & \textbf{Temporal reasoning enhances video comprehension} – Prompts that emphasize temporal order improve accuracy on video-based tasks. \textcolor{blue}{(11.4, 11.5)} 
& \textbf{Temporal reasoning enhances video comprehension} – Prompts that emphasize temporal aspects of events in videos result in more accurate responses. \textcolor{blue}{(11.4 \& 11.5)} \\
\hline


\multirow{4}{*}{\centering 9} & \textbf{Image-focused prompting helps} – Directing the model to rely solely on the image content improves answer accuracy. \textcolor{blue}{(11.1)} 
& \textbf{Asking to focus on image or question hinders performance} – In contrast to open-source models, proprietary models do worse when explicitly told to focus only on the image or only on the question. \textcolor{blue}{(11.1 \& 11.2)} \\
\hline


\multirow{4}{*}{\centering 10} & \textbf{Answer leakage degrades performance} – Including unintended hints or answer cues leads to lower accuracy. \textcolor{blue}{(Category 7)} 
& \textbf{Asking to avoid bias or stereotypes helps} – Prompts that explicitly instruct the model to avoid bias or stereotypes lead to more accurate responses. \textcolor{blue}{(Category 10)} \\
\hline

\end{tabular}
\caption{Prompting Principles for Open-Source and Proprietary LMMs. Drawing from our comprehensive prompt sensitivity analysis (section~\ref{sec4}), we derive a set of prompt templates and guidelines designed to elicit more stable and accurate responses from LMMs in MCQA task.}
\label{table:Prompting Principles}
\end{table*}

\normalsize

\subsubsection{Which Prompts Enhance or Hinder Model Performance?}
In this section, we analyze the prompts that are most effective for MCQA task for LMMs. As before, we separate the analysis into open-source and proprietary models.

\begin{figure}[!htbp]
    \centering
    \includegraphics[width=1\columnwidth]{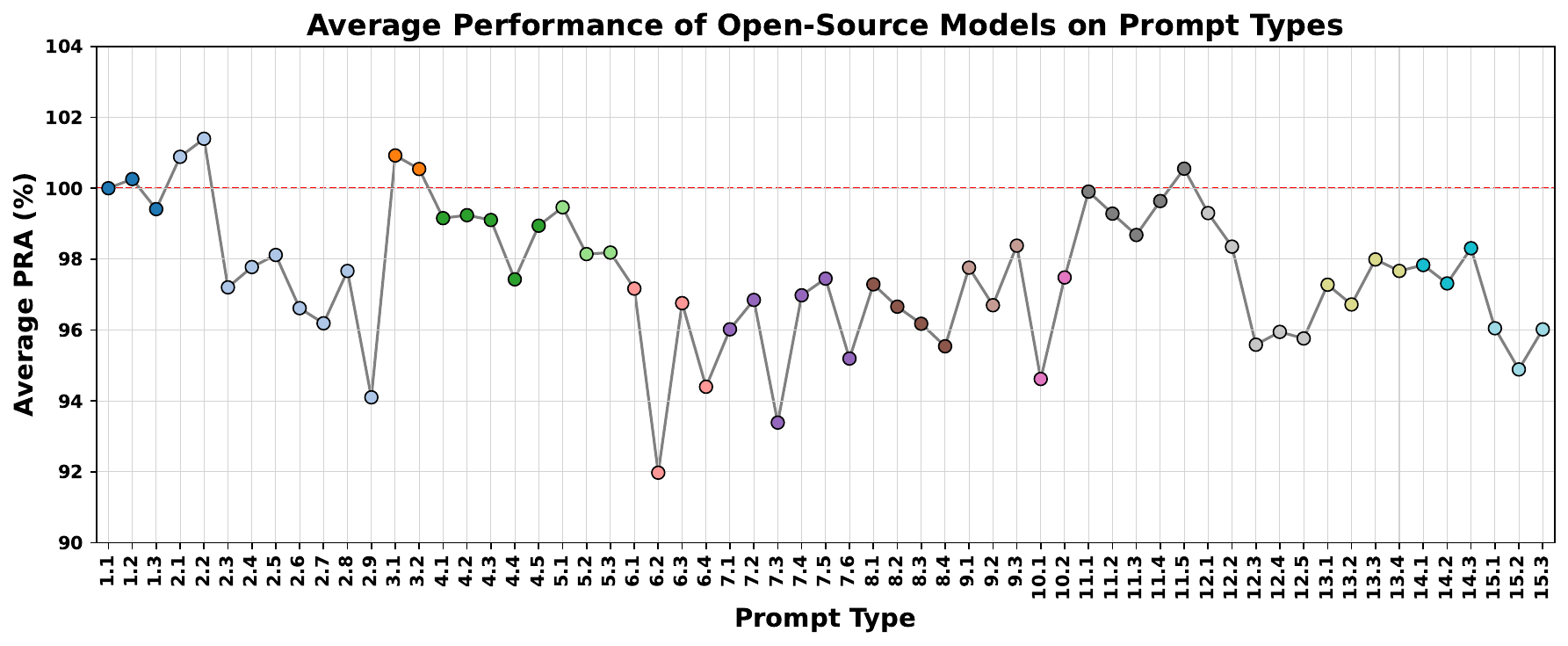} 
    \caption{Average Prompt Performance for Open-Source Models. PRA with respect to the Baseline Prompt Accuracy is averaged across Open-source Models and the 3 Benchmarks (MMStar, MMMU-Pro \& MVBench) for each Prompt Type.}
    \label{fig:Average-Prompt-Open}
\end{figure}

\begin{figure}[!htbp]
    \centering
    \includegraphics[width=1\columnwidth]{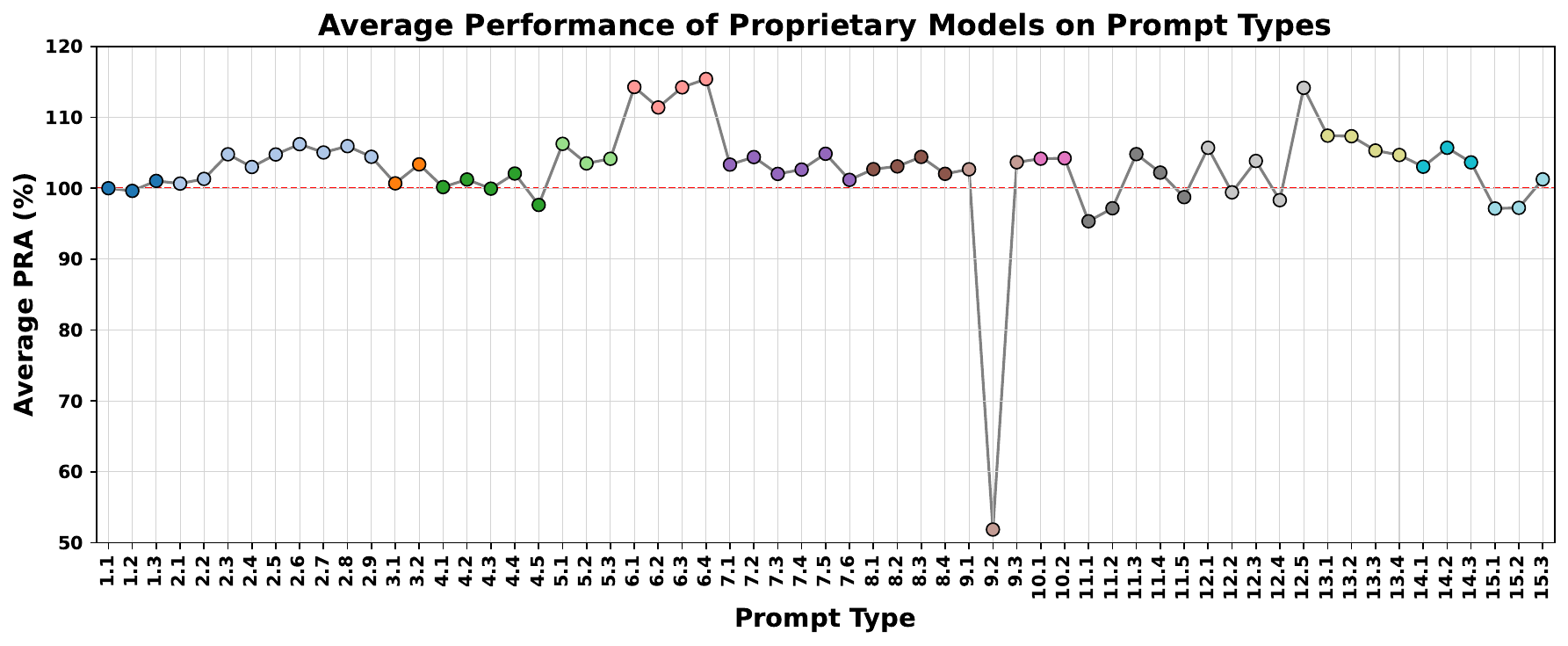} 
    \caption{Average Prompt Performance for Proprietary Models. PRA with respect to the Baseline Prompt Accuracy is averaged across Proprietary Models and the 3 Benchmarks (MMStar, MMMU-Pro \& MVBench) for each Prompt Type.}
    \label{fig:Average-Prompt-Close}
\end{figure}

Models exhibit different accuracy ranges across benchmarks. To enable a unified comparison, we normalize them to the same scale using percentage relative accuracy (PRA), as per Equation \ref{eq:pra_baseline}, with respect to the accuracy of the baseline prompt for the model on the given benchmark. Then, for each prompt type, the values are averaged across models and benchmarks separately for open-source and proprietary models (Figure \ref{fig:Average-Prompt-Open} \& \ref{fig:Average-Prompt-Close}). For open-source models, percentage relative accuracies below 80\% were excluded from the averaging process, as they represent model-specific extreme cases (dicussed in Appendix \ref{subsec:Model specific interesting cases}). The deviation from the baseline (Equation \ref{eq:prad_baseline}) was then considered to generate the figures \ref{fig:Best-Worst-Open-Baseline} \& \ref{fig:Best-Worst-Closed-Baseline} (Appendix \ref{sec:appendix-best_worst}), which highlight the best and worst-performing prompts within each category. 

For open-source models, prompt 1.2, 2.1, 2.2, 3.1, 3.2, and 11.5 consistently outperformed the baseline, indicating their effectiveness in improving model accuracy. Additionally, prompt 1.3, 4.2, 5.1, 11.1, 11.4, and 12.1, though slightly below the baseline, remained within a close range, suggesting they are still viable prompting strategies. Conversely, prompts such as 2.9, 6.2, 7.3, 10.1, 12.3, and 15.2 consistently resulted in lower accuracy, suggesting they hinder model performance.

For proprietary models, the majority of prompts enhanced performance relative to the baseline, with only a few exceptions, namely 4.5, 9.2 (actual drop –50\%, capped at –15\% for readability), 11.1, 12.4, and 15.1, showing reduced accuracy. 

We designed two video-specific prompts, 11.4 and 11.5, inspired by MVBench \cite{li2024mvbenchcomprehensivemultimodalvideo} and MMBench-Video \cite{fang2024mmbenchvideolongformmultishotbenchmark} respectively, to explicitly address the temporal dimension in videos. Notably, these prompts had a positive impact on performance in the MVBench video benchmark.

\subsection{Model, Modality \& Benchmark Level Analysis}

To further understand model behavior, we conducted an in-depth analysis to identify which models exhibit the highest sensitivity to prompt variations. Specifically, we examined the impact of positive, negative, and neutral prompt types on model sensitivity (Appendix \ref{sec:appendix-model Sensitivity}). Additionally, we investigated prompt sensitivity at a finer granularity by analyzing which question types in MMMU-Pro, which reasoning dimensions in MMStar, and which temporal tasks in MVBench are most affected by prompt changes (Appendix \ref{sec:appendix-BM Level Analysis}). Due to space constraints, the full set of results and detailed breakdowns are provided in the appendix.

\section{Prompting Principles}


Based on the insights from Section \ref{sec4}, we outline best practices for optimizing LMM performance on the MCQA task in Table \ref{table:Prompting Principles}. These strategies are designed to enhance both accuracy and consistency. While our insights are based on MCQA evaluations, we believe these principles can be broadly applied to other tasks and extended to LLMs and LMMs.

An important observation underlying these principles is the clear difference in behavior between open-source and proprietary models. Open-source models are often not extensively instruction-tuned, which makes them less responsive to prompt variations. In contrast, proprietary models typically undergo rigorous instruction tuning with large-scale, high-quality data, as well as advanced reinforcement learning and post-training techniques. This makes them considerably more sensitive to user instructions, where even subtle changes in prompt phrasing can lead to notable differences in performance. 

Given these differences in instruction-following capabilities, we present prompting principles separately for open-source and proprietary models. This distinction allows us to account for their varying adherence to instructions and to highlight strategies that are most effective for each category.

\section{Related Work}

\noindent
\textbf{Prompt Sensitivity of LLMs}: A growing body of research investigates the sensitivity of large language models (LLMs) to various prompting strategies. \cite{alzahrani-etal-2024-benchmarks} showed that minor modifications in multiple-choice question benchmarks can result in significant ranking shifts, indicating that current evaluation metrics may not provide stable comparisons. In addition to benchmark perturbations, prompt design also plays a crucial role in LLM performance. While system prompt personas are often incorporated to guide responses, their effectiveness remains inconsistent across different contexts \cite{zheng2024ahelpfulassistantreally}. Moreover, the structure and format of prompts significantly influence outcomes, with studies showing that prompt formatting alone can lead to performance variations as large as 40\% \cite{he2024doespromptformattingimpact}.  

Prompt sensitivity has also been analyzed through new evaluation metrics: PromptSensiScore and decoding confidence have been proposed to quantify how models respond to rephrasings \cite{zhuo2024prosaassessingunderstandingprompt}, while sensitivity and consistency measures have been introduced to capture how LLM predictions change across prompt rephrasing in classification tasks \cite{errica2025didiwrongquantifying}. \cite{cao2024worstpromptperformancelarge} has introduced ROBUSTALPACAEVAL, a collection of semantically equivalent prompts for evaluating how sensitive LLMs are to minor variations, showing performance swings of up to 45\% for some models depending on the formulation. Other work has highlighted how the sentiment of prompts affects LLM outputs across coherence, factuality, and bias in various applications, finding that negative phrasing often harms accuracy and increases bias, while positive phrasing can lead to verbosity \cite{gandhi2025promptsentimentcatalystllm}. Large-scale investigations have further expanded this line of research, including datasets with over 250M prompt perturbations designed to measure sensitivity across multiple dimensions and tasks \cite{habba2025dovelargescalemultidimensionalpredictions}. These findings collectively emphasize the need for standardized, well-defined methodologies when evaluating and deploying LLMs.

\noindent
\textbf{Prompt Sensitivity of LMMs:} 
While the impact of textual prompt variations has been extensively studied in unimodal LLMs and CLIP-based VLMs, research on multimodal LMMs remains limited. Dumpala et al.\ \cite{dumpala2024sensitivitygenerativevlmssemantically} showed that generative VLMs are responsive to lexical and semantic changes, and Awal et al.\ \cite{awal2025investigatingpromptingtechniqueszero} investigated prompting strategies for zero- and few-shot VQA. However, these efforts are narrow in scope. To the best of our knowledge, no prior work has systematically examined prompt sensitivity across multiple LMMs and modalities (image, multi-image, and video). Our work addresses this gap by introducing a unified evaluation framework, a curated taxonomy of 15 prompt categories, and actionable prompting principles for fairer and more consistent assessment of LMMs.

\section{Conclusion}

We present the most comprehensive analysis to date on the impact of prompt design in Large Multimodal Models (LMMs) for MCQA across image and video benchmarks. Using Promptception, a systematic framework covering 61 prompt types, we evaluate 10 models on 3 datasets. Our findings reveal that prompt phrasing substantially affects performance: proprietary models exhibit strong instruction-following but higher sensitivity, while open-source models are more stable yet less responsive to subtle cues. We hope this work advances fair and transparent LMM evaluation.

\section{Limitations}

While the proposed prompt designs were primarily developed for multimodal MCQA task, their potential applicability extends to a broader range of vision-language tasks. However, to realize this potential, it is necessary to develop a more comprehensive and task-specific prompt framework. This would involve a careful study of task types beyond MCQA, such as video captioning, visual reasoning, and image/video-grounded dialogue to craft prompts tailored to the unique demands of open-ended tasks. 

Moreover, while our manually designed prompts offer strong performance and serve as a set of best practices, automatic prompt generation is a crucial next step toward scaling this approach across a wider set of tasks. A promising direction involves the use of meta-prompting \cite{mirza2024metapromptingautomatingzeroshotvisual}, where a higher-level prompt is used to guide a language model in generating a task-specific prompt based on the input. To further streamline this process, an alternative direction is to train a lightweight prompt-generation model \cite{salehi2024vipervisualpersonalizationgenerative} that can directly output high-quality prompts conditioned on the input.

\appendix

\section{Promptception: The Complete Prompt List}
\label{sec:appendix-Prompt List}

This appendix presents the complete list of prompts proposed in our framework, Promptception. We conducted experiments on three benchmarks: \textbf{MMStar}, \textbf{MMMU-Pro}, and \textbf{MVBench}.

For the image-based benchmarks (\textbf{MMStar} and \textbf{MMMU-Pro}), we used a shared set of prompts. In contrast, for the video-based benchmark (\textbf{MVBench}), we introduced slight modifications as shown by the color coding introduced below.

Additionally, we observed that the way we expect the answer letter should be  varied between open-source and proprietary models, based on performance trends observed for prompts in \textit{Category 12}.

\noindent
To clearly indicate the differences among prompts, we use the following color coding:

\setlength\itemindent{0em}
\setlength\leftmargini{0em}
\begin{itemize}
    \item \textcolor{black}{\textbf{Black}}: Part of the prompt common to all settings \vspace{-1em}
    \item \textcolor{steelblue}{\textbf{Blue}}: Part of the prompt specific to image-based benchmarks \vspace{-1em}
    \item \textcolor{burntorange}{\textbf{Orange}}: Part of the prompt specific to the video-based benchmark \vspace{-1em}
    \item \textcolor{forestgreen}{\textbf{Green}}: Part of the prompt tailored for open-source models \vspace{-1em}
    \item \textcolor{plum}{\textbf{Purple}}: Part of the prompt tailored for proprietary models
\end{itemize}

\onecolumn{
\scriptsize
\setlength{\parskip}{0pt}
\renewcommand{\arraystretch}{1.3}

\begin{longtable}{%
  |>{\raggedright\arraybackslash}m{2.5cm}
  |>{\raggedright\arraybackslash}m{3cm}
  |>{\raggedright\arraybackslash}m{2.5cm}
  |>{\raggedright\arraybackslash}m{7cm}|}
\hline
\cellcolor{gray!30}\makebox[2.5cm][c]{\textbf{Super Category}} &
\cellcolor{gray!30}\makebox[3cm][c]{\textbf{Category}} &
\cellcolor{gray!30}\makebox[2.5cm][c]{\textbf{Type}} &
\cellcolor{gray!30}\makebox[7cm][c]{\textbf{Prompt}} \\
\hline

\endfirsthead

\hline
\cellcolor{gray!30}\makebox[2.5cm][c]{\textbf{Super Category}} &
\cellcolor{gray!30}\makebox[3cm][c]{\textbf{Category}} &
\cellcolor{gray!30}\makebox[2.5cm][c]{\textbf{Type}} &
\cellcolor{gray!30}\makebox[7cm][c]{\textbf{Prompt}} \\
\hline

\endhead

\hline \multicolumn{4}{r}{\textit{Continued on next page}} \\
\endfoot

\endlastfoot

    \multirow{1}{4cm}{Choice Formatting \newline and Presentation} 
    & \multirow{1}{4cm}{Formatting Variations \newline in Choice Presentation} 
    & \multirow{1}{4cm}{\textbf{1.1:\newline<LETTER>.}} & What design element best describes the visuals?
    \\ 
    &  &  & \textbf{A.} Composition \\
    &  &  & \textbf{B.} Perspective \\
    &  &  & \textbf{C.} Balance \\
    &  &  & \textbf{D.} Shape \\
    &  &  & \textbf{Answer with the option letter from the given choices directly.} \\ 
    \hline

    &  & \multirow{1}{4cm}{\textbf{1.2:\newline(<LETTER>)}} & What design element best describes the visuals? \\ 
    &  &  & \textbf{(A)} Composition \\
    &  &  & \textbf{(B)} Perspective \\
    &  &  & \textbf{(C)} Balance \\
    &  &  & \textbf{(D)} Shape \\
    &  &  & \textbf{Answer with the option letter from the given choices directly.} \\ 
    \hline

    &  & \multirow{1}{4cm}{\textbf{1.3:\newline Option <LETTER>:}} & What design element best describes the visuals? \\ 
    &  &  & \textbf{Option A:} Composition \\
    &  &  & \textbf{Option B:} Perspective \\
    &  &  & \textbf{Option C:} Balance \\
    &  &  & \textbf{Option D:} Shape \\
    &  &  & \textbf{Answer with the option letter from the given choices directly.} \\ 
    \hline

    & \multirow{1}{4cm}{Structured \newline formatting} 
    & \multirow{1}{4cm}{\textbf{2.1:\newline Question Prefix}} & \textbf{Question:}What design element best describes the visuals? \\ 
    &  &  & \textbf{A.} Composition \\
    &  &  & \textbf{B.} Perspective \\
    &  &  & \textbf{C.} Balance \\
    &  &  & \textbf{D.} Shape \\
    &  &  & \textbf{Answer with the option letter from the given choices directly.} \\ 
    \hline

    & & \multirow{1}{4cm}{\textbf{2.2:\newline Question \& \newline Answer Prefix}} & \textbf{Question:}What design element best describes the visuals? \newline \textbf{Options:} \\
    &  &  & \textbf{A.} Composition \\
    &  &  & \textbf{B.} Perspective \\
    &  &  & \textbf{C.} Balance \\
    &  &  & \textbf{D.} Shape \\
    &  &  & \textbf{Answer with the option letter from the given choices directly.} \\ 
    \hline
    &  & \multirow{1}{4cm}{\textbf{Type 2.3:**}} & \textbf{**Instructions**:} \\ 
    &  &  & \textbf{1.} \textbf{Analyze the given \textcolor{steelblue}{image(s)}/\textcolor{burntorange}{video} and understand its content.} \\ 
    &  &  & \textbf{2.} \textbf{Read the question carefully.} \\ 
    &  &  & \textbf{3.} \textbf{Choose the most appropriate answer from the options provided.} \\ 
    &  &  & \\ 
    &  &  & \textbf{**Question:**} What design element best describes the visuals? \\ 
    &  &  & \\ 
    &  &  & \textbf{**Options:**} \\ 
    &  &  & A. Composition \\ 
    &  &  & B. Perspective \\ 
    &  &  & C. Balance \\ 
    &  &  & D. Shape \\ 
    &  &  & \\ 
    &  &  & \textbf{**Answer:**} \\ 
    \hline

    &  & \multirow{1}{4cm}{\textbf{Type 2.4: \#\#}} & \textbf{\#\#Instructions\#\#:} \\ 
    &  &  & 1. \textbf{Analyze the given \textcolor{steelblue}{image(s)}/\textcolor{burntorange}{video} and understand its content.} \\ 
    &  &  & 2. \textbf{Read the question carefully.} \\ 
    &  &  & 3. \textbf{Choose the most appropriate answer from the options provided.} \\ 
    &  &  & \\ 
    &  &  & \textbf{\#\#Question\#\#:} What design element best describes the visuals? \\ 
    &  &  & \\ 
    &  &  & \textbf{\#\#Options\#\#:} \\ 
    &  &  & A. Composition \\ 
    &  &  & B. Perspective \\ 
    &  &  & C. Balance \\ 
    &  &  & D. Shape \\ 
    &  &  & \\ 
    &  &  & \textbf{\#\#Answer\#\#:} \\ 
    \hline

    &  & \multirow{1}{4cm}{\textbf{Type 2.5: Compact}} & \textbf{TASK:} \textbf{Analyze the \textcolor{steelblue}{image(s)}/\textcolor{burntorange}{video} and pick the best option.} \\ 
    &  &  & \\ 
    &  &  & \textbf{QUESTION:} What design element best describes the visuals? \\ 
    &  &  & \\ 
    &  &  & \textbf{OPTIONS:} \\ 
    &  &  & A. Composition \\ 
    &  &  & B. Perspective \\ 
    &  &  & C. Balance \\ 
    &  &  & D. Shape \\ 
    &  &  & \\ 
    &  &  & \textbf{BEST OPTION:} \\ 
    \hline

    &  & \multirow{1}{4cm}{\textbf{Type 2.6: Plaintext}} & 
    \textbf{You are an advanced multimodal AI assistant specializing in analyzing visual content and answering questions based on it.} \\ 
    &  &  & \textbf{Analyze the given \textcolor{steelblue}{image(s)}/\textcolor{burntorange}{video} and understand its content, read the question carefully and choose the most appropriate answer from the options provided. Here is the question: }What design element best describes the visuals? A. Composition B. Perspective C. Balance D. Shape. \textcolor{forestgreen}{\textbf{Respond only with: ‘\$LETTER’ of the best answer choice.}}/\textcolor{plum}{\textbf{Respond in the following format: 'Answer: \$LETTER' (without quotes) where LETTER is one of the options.}} \\ 
    \hline
    &  & \multirow{1}{4cm}{\textbf{Type 2.7: Markdown}} & 
    \textbf{\#\# Persona} \\ 
    &  &  & \textbf{You are an advanced multimodal AI assistant specializing in analyzing visual content and answering questions based on it.} \\ 
    &  &  & \\ 
    &  &  & \textbf{\#\# Instructions} \\ 
    &  &  & - \textbf{Analyze the given \textcolor{steelblue}{image(s)}/\textcolor{burntorange}{video} and understand its content.} \\ 
    &  &  & - \textbf{Read the question carefully.} \\ 
    &  &  & - \textbf{Choose the most appropriate answer from the options provided.} \\ 
    &  &  & \\ 
    &  &  & \textbf{\#\# Question} \\ 
    &  &  & What design element best describes the visuals? \\ 
    &  &  & \\ 
    &  &  & \textbf{\#\# Options} \\ 
    &  &  & A. Composition \\ 
    &  &  & B. Perspective \\ 
    &  &  & C. Balance \\ 
    &  &  & D. Shape \\ 
    &  &  & \\ 
    &  &  & \textbf{\#\# Output Format} \\ 
    &  &  & \textbf{\textcolor{forestgreen}{\textbf{Respond only with: ‘\$LETTER’ of the best answer choice.}}/\textcolor{plum}{\textbf{Respond in the following format: 'Answer: \$LETTER' (without quotes) where LETTER is one of the options.}}} \\ 
    \hline
    &  & \multirow{1}{4cm}{\textbf{Type 2.8: YAML}} & 
    \textbf{Persona:} \\ 
    &  &  & \textbf{You are an advanced multimodal AI assistant specializing in analyzing visual content and answering questions based on it.} \\ 
    &  &  & \\ 
    &  &  & \textbf{Instructions:} \\ 
    &  &  & \textbf{- Analyze the given \textcolor{steelblue}{image(s)}/\textcolor{burntorange}{video} and understand its content. }\\ 
    &  &  & \textbf{- Read the question carefully}. \\ 
    &  &  & \textbf{- Choose the most appropriate answer from the options provided.} \\ 
    &  &  & \\ 
    &  &  & \textbf{Question:} \\ 
    &  &  & What design element best describes the visuals? \\ 
    &  &  & \\ 
    &  &  & \textbf{Options:} \\ 
    &  &  & A. Composition \\ 
    &  &  & B. Perspective \\ 
    &  &  & C. Balance \\ 
    &  &  & D. Shape \\ 
    &  &  & \\ 
    &  &  & \textbf{Output Format:} \\ 
    &  &  & \textbf{\textcolor{forestgreen}{\textbf{Respond only with: ‘\$LETTER’ of the best answer choice.}}/\textcolor{plum}{\textbf{Respond in the following format: 'Answer: \$LETTER' (without quotes) where LETTER is one of the options.}}} \\ 
    \hline
    &  & \multirow{1}{4cm}{\textbf{Type 2.9: Json}} & 
    \{ \\ 
    &  &  & \quad \textbf{"Persona"}: \textbf{"You are an advanced multimodal AI assistant specializing in analyzing visual content and answering questions based on it."}, \\ 
    &  &  & \\ 
    &  &  & \quad \textbf{"Instructions"}: [ \\ 
    &  &  & \quad \quad \textbf{"Analyze the given \textcolor{steelblue}{image(s)}/\textcolor{burntorange}{video} and understand its content.",} \\ 
    &  &  & \quad \quad \textbf{"Read the question carefully.", }\\ 
    &  &  & \quad \quad \textbf{"Choose the most appropriate answer from the options provided." }\\ 
    &  &  & \quad ], \\ 
    &  &  & \\ 
    &  &  & \quad \textbf{"Question"}: "What design element best describes the visuals?", \\ 
    &  &  & \\ 
    &  &  & \quad \textbf{"Options"}: [ \\ 
    &  &  & \quad \quad "A. Composition", \\ 
    &  &  & \quad \quad "B. Perspective", \\ 
    &  &  & \quad \quad "C. Balance", \\ 
    &  &  & \quad \quad "D. Shape" \\ 
    &  &  & \quad ], \\ 
    &  &  & \\ 
    &  &  & \quad \textbf{"Output Format"}: "\textcolor{forestgreen}{\textbf{Respond only with: ‘\$LETTER’ of the best answer choice.}}/\textcolor{plum}{\textbf{Respond in the following format: 'Answer: \$LETTER' (without quotes) where LETTER is one of the options.}}" \\ 
    &  &  & \} \\ 
    \hline
    & \multirow{1}{4cm}{\textbf{Category 3: Prompt \newline
    Position Changes}} & 
    \multirow{1}{4cm}{\textbf{Type 3.1: Start}} & 
    \textbf{Answer with the option letter from the given choices directly.} \\ 
    &  &  & What design element best describes the visuals? \\ 
    &  &  & A. Composition \\ 
    &  &  & B. Perspective \\ 
    &  &  & C. Balance \\ 
    &  &  & D. Shape \\ 
    \hline
    &  & \multirow{1}{4cm}{\textbf{Type 3.2: Middle}} & 
    What design element best describes the visuals? \\ 
    &  &  & \textbf{Answer with the option letter from the given choices directly.} \\ 
    &  &  & A. Composition \\ 
    &  &  & B. Perspective \\ 
    &  &  & C. Balance \\ 
    &  &  & D. Shape \\ 
    \hline
    \multirow{1}{4cm}{\textbf{Super\_Category2: \newline Linguistic and \newline
    Stylistic Challenges}} 
    & \multirow{1}{4cm}{\textbf{Category 4: \newline Poor Linguistic \newline
    Formatting}} 
    & \multirow{1}{4cm}{\textbf{Type 4.1: \newline Mispelled Word}} 
    & What design element best describes the visuals? \\ 
    &  &  & A. Composition \\ 
    &  &  & B. Perspective \\ 
    &  &  & C. Balance \\ 
    &  &  & D. Shape \\ 
    &  &  & \textbf{Ansr with the optin lttr from the givn choics direly.} \\ 
    \hline
    &  &  \multirow{1}{4cm}{\textbf{Type 4.2: \newline Poor Sentence \newline
    Structuring}} 
    & What design element best describes the visuals? \\ 
    &  &  & A. Composition \\ 
    &  &  & B. Perspective \\ 
    &  &  & C. Balance \\ 
    &  &  & D. Shape \\ 
    &  &  & \textbf{Option letter from choices answer. Directly give.} \\ 
    \hline
    &  &  \multirow{1}{4cm}{\textbf{Type 4.3: \newline All-Capital \newline
    Questions}} 
    & WHAT DESIGN ELEMENT BEST DESCRIBES THE VISUALS? \\ 
    &  &  & A. Composition \\ 
    &  &  & B. Perspective \\ 
    &  &  & C. Balance \\ 
    &  &  & D. Shape \\ 
    &  &  & \textbf{ANSWER WITH THE OPTION LETTER FROM THE GIVEN CHOICES DIRECTLY.} \\ 
    \hline

    &  &  \multirow{1}{4cm}{\textbf{Type 4.4: \newline Poor Formatting}} 
    & What design element best describes the visuals? \\ 
    &  &  & A. Composition \\ 
    &  &  & B. Perspective \\ 
    &  &  & C. Balance \\ 
    &  &  & D. Shape \\ 
    &  &  & \textbf{answer.with;the.option: letter from-choices.directly!} \\ 
    \hline

    &  &  \multirow{1}{4cm}{\textbf{Type 4.5: \newline Letter Leak}} 
    & What design element best describes the visuals? \\ 
    &  &  & A. Composition \\ 
    &  &  & B. Perspective \\ 
    &  &  & C. Balance \\ 
    &  &  & D. Shape \\ 
    &  &  & \textbf{Answer with the option letter from the given choices directly (A/A/A/A).} \\ 
    \hline
    & \multirow{1}{4cm}{\textbf{Category 5: Effect of \newline
    Prompt Length}} & 
    \multirow{1}{4cm}{\textbf{Type 5.1: Short \newline
    Prompt}} & 
    What design element best describes the image? <image> \\ 
    &  &  & A. Composition \\ 
    &  &  & B. Perspective \\ 
    &  &  & C. Balance \\ 
    &  &  & D. Shape \\ 
    &  &  & \textbf{\textcolor{forestgreen}{Respond only with: ‘\$LETTER’ of the best answer choice.}/\textcolor{plum}{Respond in the following format: 'Answer: \$LETTER' (without quotes) where LETTER is one of the options.}} \\ 
    \hline
    & \multirow{1}{4cm} & 
    \multirow{1}{4cm}{\textbf{Type 5.2: Medium \newline
    Prompt}} & 
    What design element best describes the visuals? \\ 
    &  &  & A. Composition \\ 
    &  &  & B. Perspective \\ 
    &  &  & C. Balance \\ 
    &  &  & D. Shape \\ 
    &  &  & \textbf{Your task is to examine the given \textcolor{steelblue}{image(s)}/\textcolor{burntorange}{video} and determine which of the listed options accurately answers the question. Carefully analyze the image(s), consider the possibilities, and then \textcolor{forestgreen}{respond only with: ‘\$LETTER’ of the best answer choice.}/\textcolor{plum}{respond in the following format: 'Answer: \$LETTER' (without quotes) where LETTER is one of the options.}} \\ 
    \hline

    &  &  \multirow{1}{4cm}{\textbf{Type 5.3: Long \newline
    Prompt}} & 
    What design element best describes the visuals? \\ 
    &  &  & A. Composition \\ 
    &  &  & B. Perspective \\ 
    &  &  & C. Balance \\ 
    &  &  & D. Shape \\ 
    &  &  &\textbf{ In this task, you are expected to carefully evaluate the input provided and analyze all relevant aspects before making a decision. It is essential to consider every detail thoroughly and ensure that your response aligns with the required format. Pay close attention to the task instructions and ensure that your answer reflects careful thought and accuracy. Once you have reached a conclusion, provide a response that adheres strictly to the guidelines. Avoid adding unnecessary details or deviations from the expected output format. Precision and adherence to instructions are critical to completing this task successfully. \textcolor{forestgreen}{Respond only with: ‘\$LETTER’ of the best answer choice.}/\textcolor{plum}{Respond in the following format: ‘Answer: \$LETTER’ (without quotes) where LETTER is one of the options.}} \\ 
    \hline
    \multirow{1}{4cm}{\textbf{Super\_Category3: \newline Thought Process \newline
    and Reasoning}} 
    & \multirow{1}{4cm}{\textbf{Category 6: \newline Chain of Thought \newline
    (CoT) Prompt}} 
    & \multirow{1}{4cm}{\textbf{Type 6.1: \newline Step-by-Step \newline Reasoning}} 
    & What design element best describes the visuals? \\ 
    &  &  & A. Composition \\ 
    &  &  & B. Perspective \\ 
    &  &  & C. Balance \\ 
    &  &  & D. Shape \\ 
    &  &  & \textbf{Answer the preceding multiple-choice question. Think step by step before answering. \textcolor{forestgreen}{Respond only with: ‘\$LETTER’ of the best answer choice.}/\textcolor{plum}{The last line of your response should be of the following format: 'Answer: LETTER' (without quotes) where LETTER is one of the options.}} \\ 
    \hline
       &  &  \multirow{1}{4cm}{\textbf{Type 6.2: \newline Self-Consistency}} 
    & What design element best describes the visuals? \\ 
    &  &  & A. Composition \\ 
    &  &  & B. Perspective \\ 
    &  &  & C. Balance \\ 
    &  &  & D. Shape \\ 
    &  &  & \textbf{Propose two distinct reasoning approaches to solve the question based on the given \textcolor{steelblue}{image(s)}/\textcolor{burntorange}{video} and question.} \\ 
    &  &  & \textbf{For each approach, outline the steps and logic used to arrive at a conclusion.} \\ 
    &  &  & \textbf{Compare the outcomes of both approaches and justify which answer (letter) is the most appropriate based on the provided evidence. }\\ 
    &  &  & \textbf{\textcolor{forestgreen}{Respond only with: ‘\$LETTER’ of the best answer choice.}/\textcolor{plum}{The last line of your response should be of the following format: 'Answer: LETTER' (without quotes) where LETTER is one of the options.}} \\ 
    \hline

    &  &  \multirow{1}{4cm}{\textbf{Type 6.3: \newline Step-back}} 
    & What design element best describes the visuals? \\ 
    &  &  & A. Composition \\ 
    &  &  & B. Perspective \\ 
    &  &  & C. Balance \\ 
    &  &  & D. Shape \\ 
    &  &  & \textbf{Let's think step-by-step to answer this:} \\ 
    &  &  & \textbf{Step 1) Analyze the visual input carefully, noting key elements, relationships, and details relevant to the question.} \\ 
    &  &  & \textbf{Step 2) Abstract the key concepts and principles from both the visual input and the question text.} \\ 
    &  &  & \textbf{Step 3) Use these abstractions to reason through the question systematically.} \\ 
    &  &  & \textbf{\textcolor{forestgreen}{Respond only with: ‘\$LETTER’ of the best answer choice.}/\textcolor{plum}{The last line of your response should be of the following format: 'Answer: LETTER' (without quotes) where LETTER is one of the options.}} \\ 
    \hline
    
    &  &  \multirow{1}{4cm}{\textbf{Type 6.4: \newline Thread of Thought}} 
    & What design element best describes the visuals? \\ 
    &  &  & A. Composition \\ 
    &  &  & B. Perspective \\ 
    &  &  & C. Balance \\ 
    &  &  & D. Shape \\ 
    &  &  & \textbf{Walk through the context of the preceding multiple-choice question in manageable parts step-by-step, summarizing and analyzing as we progress.} \\ 
    &  &  & \textbf{\textcolor{forestgreen}{Respond only with: ‘\$LETTER’ of the best answer choice.}/\textcolor{plum}{The last line of your response should be of the following format: 'Answer: LETTER' (without quotes) where LETTER is one of the options.}} \\ 
    \hline
    & \multirow{1}{4cm}{\textbf{Category 7: \newline Ambiguity  and\newline Probabilistic Prompts}} 
    & \multirow{1}{4cm}{\textbf{Type 7.1: \newline Add Probabilistic \newline
    Language}} 
    & What design element best describes the visuals? \\ 
    &  &  & A. Composition \\ 
    &  &  & B. Perspective \\ 
    &  &  & C. Balance \\ 
    &  &  & D. Shape \\ 
    &  &  & \textbf{Based on our analysis of the \textcolor{steelblue}{image(s)}/\textcolor{burntorange}{video}, there is a 25\% chance that the correct answer is option A, but the other options may still be valid.} \\ 
    &  &  & \textbf{Carefully evaluate the \textcolor{steelblue}{image(s)}/\textcolor{burntorange}{video} and choose the answer letter you believe is most likely to be correct.} \\ 
    &  &  & \textbf{\textcolor{forestgreen}{Respond only with: ‘\$LETTER’ of the best answer choice.}/\textcolor{plum}{The last line of your response should be of the following format: 'Answer: LETTER' (without quotes) where LETTER is one of the options.}} \\
    \hline
    &  &  \multirow{1}{4cm}{\textbf{Type 7.2: \newline Include Uncertainty \newline in Instruction}}
    & What design element best describes the visuals? \\ 
    &  &  & A. Composition \\ 
    &  &  & B. Perspective \\ 
    &  &  & C. Balance \\ 
    &  &  & D. Shape \\ 
    &  &  & \textbf{I'm not entirely sure of the exact answer here, but I somewhat trust your ability to analyze the \textcolor{steelblue}{image(s)}/\textcolor{burntorange}{video}.} \\ 
    &  &  & \textbf{Look closely at the details and provide the answer letter that seems to be the best fit based on your understanding.} \\ 
    &  &  & \textbf{\textcolor{forestgreen}{Respond only with: ‘\$LETTER’ of the best answer choice.}/\textcolor{plum}{The last line of your response should be of the following format: 'Answer: LETTER' (without quotes) where LETTER is one of the options.}} \\ 
    \hline

    &  &  \multirow{1}{4cm}{\textbf{Type 7.3: \newline Leverage Statistical\newline Trends}} 
    & What design element best describes the visuals? \\ 
    &  &  & A. Composition \\ 
    &  &  & B. Perspective \\ 
    &  &  & C. Balance \\ 
    &  &  & D. Shape \\ 
    &  &  & \textbf{Historically, 70\% of similar questions have had option C as the answer.} \\ 
    &  &  & \textbf{Use this statistical insight as a guide but rely on your analysis of the image(s) to determine the most appropriate answer letter.} \\ 
    &  &  & \textbf{\textcolor{forestgreen}{Respond only with: ‘\$LETTER’ of the best answer choice.}/\textcolor{plum}{The last line of your response should be of the following format: 'Answer: LETTER' (without quotes) where LETTER is one of the options.}}
    \\ \hline

    &  &  \multirow{1}{4cm}{\textbf{Type 7.4: \newline Deferred Content\newline Analysis}} 
    & What design element best describes the visuals? \\ 
    &  &  & A. Composition \\ 
    &  &  & B. Perspective \\ 
    &  &  & C. Balance \\ 
    &  &  & D. Shape \\ 
    &  &  & \textbf{I'll provide more context later, but for now, focus on analyzing the \textcolor{steelblue}{image(s)}/\textcolor{burntorange}{video} carefully.} \\ 
    &  &  & \textbf{Based on what you observe, suggest the best answer letter at this stage.} \\ 
    &  &  & \textbf{\textcolor{forestgreen}{Respond only with: ‘\$LETTER’ of the best answer choice.}/\textcolor{plum}{The last line of your response should be of the following format: 'Answer: LETTER' (without quotes) where LETTER is one of the options.}}
    \\ \hline

    &  &  \multirow{1}{4cm}{\textbf{Type 7.5: \newline Acknowledged Co-\newline mplexity Response}} 
    & What design element best describes the visuals? \\ 
    &  &  & A. Composition \\ 
    &  &  & B. Perspective \\ 
    &  &  & C. Balance \\ 
    &  &  & D. Shape \\ 
    &  &  & \textbf{This question is inherently challenging, and I understand perfection might not be possible.} \\ 
    &  &  & \textbf{Examine the \textcolor{steelblue}{image(s)}/\textcolor{burntorange}{video} carefully and provide the answer letter that you believe best aligns with what you see.} \\ 
    &  &  & \textbf{\textcolor{forestgreen}{Respond only with: ‘\$LETTER’ of the best answer choice.}/\textcolor{plum}{The last line of your response should be of the following format: 'Answer: LETTER' (without quotes) where LETTER is one of the options.}} \\ 
    \hline
    &  &  \multirow{1}{4cm}{\textbf{Type 7.6: \newline Additional Options\newline for Ambiguity}} 
    & What design element best describes the visuals? \\ 
    &  &  & A. Composition \\ 
    &  &  & B. Perspective \\ 
    &  &  & C. Balance \\ 
    &  &  & D. Shape \\ 
    &  &  & \textbf{Carefully consider the question and the visual evidence before making a choice. Select the correct \textbf{\$LETTER} from the given options}. \\ 
    &  &  & \textbf{However, if the available information is unclear, ambiguous, or insufficient to provide a confident answer, you have the following additional options: }\\ 
    &  &  & \textbf{E. Not sure (if you genuinely do not know the answer).} \\ 
    &  &  & \textbf{F. Evidence not sufficient to answer (if the question cannot be answered based on the given \textcolor{steelblue}{image(s)}/\textcolor{burntorange}{video}).} \\ 
    &  &  &\textbf{ G. I'll answer later (if you prefer to delay your decision).} \\ 
    &  &  & \textbf{\textcolor{forestgreen}{Choose the \$LETTER corresponding to your conclusion and respond directly without additional commentary.}/\textcolor{plum}{Choose the LETTER corresponding to your conclusion and respond in the following format: 'Answer: LETTER'.}} \\ 
    \hline
    \multirow{1}{4cm}{\textbf{Super\_Category4: \newline Context-Aware and\newline Ethical Guidance}} 
    & \multirow{1}{4cm}{\textbf{Category 8: \newline Target Audience \newline Prompts}} 
    & \multirow{1}{4cm}{\textbf{Type 8.1: \newline Explain to a Child}} 
    & What design element best describes the visuals? \\ 
    &  &  & A. Composition \\ 
    &  &  & B. Perspective \\ 
    &  &  & C. Balance \\ 
    &  &  & D. Shape \\ 
    &  &  & \textbf{Explain the answer as if you're speaking to a curious 10-year-old.} \\ 
    &  &  & \textbf{Use simple words, short sentences, and relatable examples that a child would easily understand. Avoid technical terms or complex concepts.} \\ 
    &  &  & \textbf{\textcolor{forestgreen}{Respond only with: ‘\$LETTER’ of the best answer choice.}/\textcolor{plum}{The last line of your response should be of the following format: 'Answer: LETTER' (without quotes) where LETTER is one of the options.}} \\ 
    \hline
    &  &  \multirow{1}{4cm}{\textbf{Type 8.2: \newline Explain to a High\newline  School Student}} 
    & What design element best describes the visuals? \\ 
    &  &  & A. Composition \\ 
    &  &  & B. Perspective \\ 
    &  &  & C. Balance \\ 
    &  &  & D. Shape \\ 
    &  &  & \textbf{Explain the answer as if you're talking to a high school student.}\textbf{ Keep the explanation clear and relatable, using everyday language. Include simple examples where needed, and introduce basic terms that align with a high school level of understanding.} \\ 
    &  &  & \textbf{\textcolor{forestgreen}{Respond only with: ‘\$LETTER’ of the best answer choice.}/\textcolor{plum}{The last line of your response should be of the following format: 'Answer: LETTER' (without quotes) where LETTER is one of the options.}} \\ 
    \hline

    &  &  \multirow{1}{4cm}{\textbf{Type 8.3: \newline Explain to a\newline College Student}} 
    & What design element best describes the visuals? \\ 
    &  &  & A. Composition \\ 
    &  &  & B. Perspective \\ 
    &  &  & C. Balance \\ 
    &  &  & D. Shape \\ 
    &  &  & \textbf{Provide the answer as if you are addressing a college student. Use straightforward, concise language and assume some foundational knowledge. Offer brief context or explanations for concepts that might not be immediately familiar.}\\ 
    &  &  & \textbf{\textcolor{forestgreen}{Respond only with: ‘\$LETTER’ of the best answer choice.}/\textcolor{plum}{The last line of your response should be of the following format: 'Answer: LETTER' (without quotes) where LETTER is one of the options.}} \\ 
    \hline

    &  &  \multirow{1}{4cm}{\textbf{Type 8.4: \newline Explain to a \newline Subject-Matter \newline Expert (SME)}} 
    & What design element best describes the visuals? \\ 
    &  &  & A. Composition \\ 
    &  &  & B. Perspective \\ 
    &  &  & C. Balance \\ 
    &  &  & D. Shape \\ 
    &  &  & \textbf{Deliver a technical explanation of the answer tailored for an expert in the field. Use precise terminology, include relevant context, and assume a deep understanding of the topic. Focus on the details and nuances of the explanation.} \\ 
    &  &  & \textbf{\textcolor{forestgreen}{Respond only with: ‘\$LETTER’ of the best answer choice.}/\textcolor{plum}{The last line of your response should be of the following format: 'Answer: LETTER' (without quotes) where LETTER is one of the options.}} \\ 
    \hline
    & \multirow{1}{4cm}{\textbf{Category 9: \newline Roleplay Scenarios}} 
    & \multirow{1}{4cm}{\textbf{Type 9.1: \newline Act as a Computer\newline Vision Professor}} 
    & What design element best describes the visuals? \\ 
    &  &  & A. Composition \\ 
    &  &  & B. Perspective \\ 
    &  &  & C. Balance \\ 
    &  &  & D. Shape \\ 
    &  &  & \textbf{From now on, you are a renowned Computer Vision Professor with a deep subject knowledge. You are an expert in analyzing and interpreting visual data, with a deep understanding of both theoretical concepts and practical applications. As your student, I depend on your expertise to provide accurate and well-reasoned answers based on the input provided.} \\ 
    &  &  & \textbf{\textcolor{forestgreen}{Respond only with: ‘\$LETTER’ of the best answer choice.}/\textcolor{plum}{The last line of your response should be of the following format: 'Answer: LETTER' (without quotes) where LETTER is one of the options.}} \\ 
    \hline

    &  &  \multirow{1}{4cm}{\textbf{Type 9.2: \newline Act as a Careless\newline Student}} 
    & What design element best describes the visuals? \\ 
    &  &  & A. Composition \\ 
    &  &  & B. Perspective \\ 
    &  &  & C. Balance \\ 
    &  &  & D. Shape \\ 
    &  &  & \textbf{From now on, please act as a careless student. When answering questions based on images, you often overlook key visual details or misinterpret the question. You rush through the process and select answers without fully analyzing the image(s), leading to frequent mistakes due to carelessness.} \\ 
    &  &  & \textbf{\textcolor{forestgreen}{Respond only with: ‘\$LETTER’ of the best answer choice.}/\textcolor{plum}{The last line of your response should be of the following format: 'Answer: LETTER' (without quotes) where LETTER is one of the options.}} \\ 
    \hline

    &  &  \multirow{1}{4cm}{\textbf{Type 9.3: \newline Act as a Seasoned\newline Politician}} 
    & What design element best describes the visuals? \\ 
    &  &  & A. Composition \\ 
    &  &  & B. Perspective \\ 
    &  &  & C. Balance \\ 
    &  &  & D. Shape \\ 
    &  &  & \textbf{From now on, you are a seasoned politician known for your eloquence, strategic thinking, and unwavering commitment to public service. Your ability to navigate complex challenges and unite diverse groups has earned you widespread respect. Others turn to you for your visionary leadership and policy expertise.} \\ 
    &  &  & \textbf{\textcolor{forestgreen}{Respond only with: ‘\$LETTER’ of the best answer choice.}/\textcolor{plum}{The last line of your response should be of the following format: 'Answer: LETTER' (without quotes) where LETTER is one of the options.}} \\ 
    \hline
    & \multirow{1}{4cm}{\textbf{Category 10: \newline Testing Bias and\newline Stereotypes}} 
    & \multirow{1}{4cm}{\textbf{Type 10.1: \newline Bias Awareness}} 
    & What design element best describes the visuals? \\ 
    &  &  & A. Composition \\ 
    &  &  & B. Perspective \\ 
    &  &  & C. Balance \\ 
    &  &  & D. Shape \\ 
    &  &  & \textbf{Evaluate the \textcolor{steelblue}{image(s)}/\textcolor{burntorange}{video} and question critically, identifying any embedded stereotypes or biased assumptions. Your response should not only avoid reinforcing these stereotypes but also actively reflect the actual content without distortion.} \\ 
    &  &  & \textbf{\textcolor{forestgreen}{Respond only with the '\$LETTER' of the answer choice that challenges or negates the presence of stereotypes in favor of an accurate interpretation.}/\textcolor{plum}{Respond in the following format: 'Answer: \$LETTER' (without quotes) where LETTER is the option that challenges or negates the presence of stereotypes in favor of an accurate interpretation.}} \\ 
    \hline

    &  &  \multirow{1}{4cm}{\textbf{Type 10.2: \newline Contextual Neutra-\newline lity}} 
    & What design element best describes the visuals? \\ 
    &  &  & A. Composition \\ 
    &  &  & B. Perspective \\ 
    &  &  & C. Balance \\ 
    &  &  & D. Shape \\ 
    &  &  & \textbf{Consider the \textcolor{steelblue}{image(s)}/\textcolor{burntorange}{video} and question provided within the specific context they represent. Ensure your response reflects neutrality by avoiding cultural, social, or historical biases. Base your decision solely on the visible elements and the question's requirements, maintaining a context-aware but unbiased approach.} \\ 
    &  &  & \textbf{\textcolor{forestgreen}{Respond only with the '\$LETTER' of the answer choice that best matches the contextual content of the video and question. }/\textcolor{plum}{Respond in the following format: 'Answer: \$LETTER' (without quotes) where LETTER is the option that best matches the contextual content of the image(s) and question.}} \\ 
    \hline
    \multirow{1}{4cm}{\textbf{Super\_Category5: \newline Task-Specific\newline Instructions}} 
    & \multirow{1}{4cm}{\textbf{Category 11: \newline Focus-Driven Prompts}} 
    & \multirow{1}{4cm}{\textbf{Type 11.1: \newline Strong Focus on\newline Image(s) Analysis}} 
    & What design element best describes the visuals? \\ 
    &  &  & A. Composition \\ 
    &  &  & B. Perspective \\ 
    &  &  & C. Balance \\ 
    &  &  & D. Shape \\ 
    &  &  & \textbf{**TASK:** Examine the \textcolor{steelblue}{image(s)}/\textcolor{burntorange}{video} meticulously, focusing on every detail and visual element to identify the correct answer.} \\ 
    &  &  & \textbf{**PRIMARY GOAL:** Focus strictly on the \textcolor{steelblue}{image(s)}/\textcolor{burntorange}{video} and you \textbf{**MUST**} base your analysis solely on the content.} \\ 
    &  &  & \textbf{Answer with the option letter from the given choices directly.} \\ 
    \hline
    &  &  \multirow{1}{4cm}{\textbf{Type 11.2: \newline Strong Focus on\newline Question and \newline Options}} 
    & What design element best describes the visuals? \\ 
    &  &  & A. Composition \\ 
    &  &  & B. Perspective \\ 
    &  &  & C. Balance \\ 
    &  &  & D. Shape \\ 
    &  &  & \textbf{**TASK:** Carefully scrutinize the question and each option provided, ensuring your focus remains on understanding the choices and selecting the correct one.} \\ 
    &  &  & \textbf{**PRIMARY GOAL:** Focus entirely on the question and the options to determine the correct answer. Your response **MUST** be based solely on this information.} \\ 
    &  &  & \textbf{Answer with the option letter from the given choices directly.} \\ 
    \hline

    &  &  \multirow{1}{4cm}{\textbf{Type 11.3: \newline Strong Focus on\newline Required Answer\newline Format}} 
    & What design element best describes the visuals? \\ 
    &  &  & A. Composition \\ 
    &  &  & B. Perspective \\ 
    &  &  & C. Balance \\ 
    &  &  & D. Shape \\ 
    &  &  & \textbf{**TASK:** Review the \textcolor{steelblue}{image(s)}/\textcolor{burntorange}{video} carefully and determine the correct answer with precision.} \\ 
    &  &  & \textbf{**PRIMARY GOAL:** Ensure your output format is correct and adheres to this structure: Answer: \$LETTER.} \\ 
    &  &  & \textbf{**NON-NEGOTIABLE:** Any deviation, extra content, or improper format will result in an invalid response.} \\ 
    \hline

    &  &  \multirow{1}{4cm}{\textbf{Type 11.4: \newline Observation-\newline Driven Analysis}} 
    & What design element best describes the visuals? \\ 
    &  &  & A. Composition \\ 
    &  &  & B. Perspective \\ 
    &  &  & C. Balance \\ 
    &  &  & D. Shape \\ 
    &  &  & \textbf{Carefully observe the video, focusing on the order and causes of events, the movement and details of objects, as well as the actions and poses of persons. Based on these observations, choose the option letter that best answers the question.} \\ 
    &  &  & \textbf{Based on your observations, select the best option letter that accurately addresses the question.} \\ 
    \hline

    &  &  \multirow{1}{4cm}{\textbf{Type 11.5: \newline Chronological\newline Frame Analysis}} 
    & What design element best describes the visuals? \\ 
    &  &  & A. Composition \\ 
    &  &  & B. Perspective \\ 
    &  &  & C. Balance \\ 
    &  &  & D. Shape \\ 
    &  &  & \textbf{You will be given a set of frames uniformly sampled from a video, presented in their chronological order. \newline
    Analyze these frames carefully and determine the correct answer to the question based on the video content. \newline
    Answer with the option letter from the given choices directly.  } \\ 
    &  &  & \textbf{Please analyze these images and provide the answer to the question about the video content.} \\ 
    &  &  & \textbf{Answer with the option letter from the given choices directly.} \\ 
    \hline
    & \multirow{1}{4cm}{\textbf{Category 12: \newline Answer Handling}} 
    & \multirow{1}{4cm}{\textbf{Type 12.1: \newline Answer Handler 1}} 
    & What design element best describes the visuals? \\ 
    &  &  & A. Composition \\ 
    &  &  & B. Perspective \\ 
    &  &  & C. Balance \\ 
    &  &  & D. Shape \\ 
    &  &  & \textbf{Answer the preceding multiple-choice question in the following format: “Answer: \$LETTER” (without quotes) where LETTER is one of the options.} \\ 
    \hline
    &  &  \multirow{1}{4cm}{\textbf{Type 12.2: \newline Answer Handler 2}} 
    & What design element best describes the visuals? \\ 
    &  &  & A. Composition \\ 
    &  &  & B. Perspective \\ 
    &  &  & C. Balance \\ 
    &  &  & D. Shape \\ 
    &  &  & \textbf{Respond only with: "\$LETTER" of the best answer choice.} \\ 
    \hline

    &  &  \multirow{1}{4cm}{\textbf{Type 12.3: \newline Answer Handler 3}} 
    & What design element best describes the visuals?  \\ 
    &  &  & A. Composition \\ 
    &  &  & B. Perspective \\ 
    &  &  & C. Balance \\ 
    &  &  & D. Shape \\ 
    &  &  & \textbf{Best Choice: \$LETTER} \\ 
    \hline

    &  &  \multirow{1}{4cm}{\textbf{Type 12.4: \newline Answer Handler 4}} 
    & What design element best describes the visuals?  \\ 
    &  &  & A. Composition \\ 
    &  &  & B. Perspective \\ 
    &  &  & C. Balance \\ 
    &  &  & D. Shape \\ 
    &  &  & \textbf{Respond only with the correct option, formatted as follows:} \textbf{\$LETTER.} \\ 
    \hline

    &  &  \multirow{1}{4cm}{\textbf{Type 12.5: \newline Answer Handler 5}} 
    & What design element best describes the visuals?  \\ 
    &  &  & A. Composition \\ 
    &  &  & B. Perspective \\ 
    &  &  & C. Balance \\ 
    &  &  & D. Shape \\ 
    &  &  & \textbf{The last line of your response should be of the following format:'Answer: \$LETTER' (without quotes) where LETTER is one of the options.} \\ 
    \hline
    \multirow{1}{4cm}{\textbf{Super\_Category6: \newline Performance,\newline Feedback, and \newline Penalty}} 
    & \multirow{1}{4cm}{\textbf{Category 13: \newline Penalty-Based Prompts}} 
    & \multirow{1}{4cm}{\textbf{Type 13.1: \newline Penalties for\newline Mistakes}} 
    & What design element best describes the visuals? \\ 
    &  &  & A. Composition \\ 
    &  &  & B. Perspective \\ 
    &  &  & C. Balance \\ 
    &  &  & D. Shape \\ 
    &  &  & \textbf{**Warning:** An **incorrect answer** will result in a **"strict penalty"** being applied.} \\ 
    &  &  & \textbf{Carefully examine all details in the \textcolor{steelblue}{image(s)}/\textcolor{burntorange}{video}, analyze the question thoroughly, and select your response with precision.} \\ 
    &  &  & \textbf{Accuracy is **"non-negotiable"**, so take your time and avoid errors.} \\ 
    &  &  & \textbf{Ensure your response follows this format: Answer: \$LETTER} \\ 
    \hline
    &  &  \multirow{1}{4cm}{\textbf{Type 13.2: \newline Severe Accuracy\newline Warning}} 
    & What design element best describes the visuals? \\ 
    &  &  & A. Composition \\ 
    &  &  & B. Perspective \\ 
    &  &  & C. Balance \\ 
    &  &  & D. Shape \\ 
    &  &  & \textbf{**Notice**: Failure to provide the **correct answer** will lead to a **loss of accuracy points**, which could significantly impact your overall score.} \\ 
    &  &  & \textbf{This is your **only warning**: scrutinize the \textcolor{steelblue}{image(s)}/\textcolor{burntorange}{video} and question carefully.} \\ 
    &  &  & \textbf{Avoid careless mistakes and respond in the correct format: Answer: \$LETTER.} \\ 
    \hline

    &  &  \multirow{1}{4cm}{\textbf{Type 13.3: \newline Credibility Loss \newline Warning}} 
    & What design element best describes the visuals? \\ 
    &  &  & A. Composition \\ 
    &  &  & B. Perspective \\ 
    &  &  & C. Balance \\ 
    &  &  & D. Shape \\ 
    &  &  & \textbf{**Attention**: An **incorrect answer** will cause you to **lose credibility**, which may affect future evaluations.} \\ 
    &  &  & \textbf{This is a **serious warning**: carefully review all details, analyze the context thoroughly, and double-check your reasoning.} \\ 
    &  &  & \textbf{Incorrect responses will not be tolerated.} \\ 
    &  &  & \textbf{Your answer must be in this exact format: Answer: \$LETTER.} \\ 
    \hline

    &  &  \multirow{1}{4cm}{\textbf{Type 13.4: \newline Penalties for \newline Incorrect Formatt-\newline ing}} 
    & What design element best describes the visuals? \\ 
    &  &  & A. Composition \\ 
    &  &  & B. Perspective \\ 
    &  &  & C. Balance \\ 
    &  &  & D. Shape \\ 
    &  &  & \textbf{**Critical Notice**: Responses that do not adhere to the required format will be **penalized without exception**.} \\ 
    &  &  & \textbf{Ensure that your answer is presented exactly as: Answer: \$LETTER.} \\ 
    &  &  & \textbf{Deviations from this format will result in **automatic deductions**, regardless of the correctness of the content.} \\ 
    &  &  & \textbf{Pay attention to **both accuracy and formatting** to avoid penalties.} \\ 
    \hline
    & \multirow{1}{4cm}{\textbf{Category 14: \newline Incentive-Based\newline Prompts}} 
    & \multirow{1}{4cm}{\textbf{Type 14.1: \newline Incentive-Based\newline Prompts}} 
    & What design element best describes the visuals? \\ 
    &  &  & A. Composition \\ 
    &  &  & B. Perspective \\ 
    &  &  & C. Balance \\ 
    &  &  & D. Shape \\ 
    &  &  & \textbf{Imagine you’re competing for a generous tip of \$100 for delivering a flawless and accurate answer.} \\ 
    &  &  & \textbf{Carefully analyze the \textcolor{steelblue}{image(s)}/\textcolor{burntorange}{video} and the question provided, paying attention to every detail and nuance.} \\ 
    &  &  & \textbf{\textcolor{forestgreen}{Respond only with: ‘\$LETTER’ of the best answer choice.}/\textcolor{plum}{The last line of your response should be of the following format: 'Answer: LETTER' (without quotes) where LETTER is one of the options.}} \\ 
    &  &  & \textbf{No explanations are needed—just focus on accuracy to secure the reward.} \\ 
    \hline
    &  &  \multirow{1}{4cm}{\textbf{Type 14.2: \newline Performance-Based \newline Rewards}} 
    & What design element best describes the visuals? \\ 
    &  &  & A. Composition \\ 
    &  &  & B. Perspective \\ 
    &  &  & C. Balance \\ 
    &  &  & D. Shape \\ 
    &  &  & \textbf{Your performance will be graded, and there’s an opportunity to earn extra credit for exceptional accuracy.} \\ 
    &  &  & \textbf{Examine the \textcolor{steelblue}{image(s)}/\textcolor{burntorange}{video} and question meticulously, ensuring your analysis is both thorough and logical.} \\ 
    &  &  & \textbf{\textcolor{forestgreen}{Respond only with: ‘\$LETTER’ of the best answer choice.}/\textcolor{plum}{The last line of your response should be of the following format: 'Answer: LETTER' (without quotes) where LETTER is one of the options.}} \\ 
    &  &  & \textbf{Precise and correct answers will demonstrate your capabilities and earn you the recognition you deserve.} \\ 
    \hline

    &  &  \multirow{1}{4cm}{\textbf{Type 14.3: \newline Encouraging Better\newline Solutions}} 
    & What design element best describes the visuals? \\ 
    &  &  & A. Composition \\ 
    &  &  & B. Perspective \\ 
    &  &  & C. Balance \\ 
    &  &  & D. Shape \\ 
    &  &  & \textbf{Only the top-quality answers will earn additional points and recognition.} \\ 
    &  &  & \textbf{This is your chance to stand out by providing an exceptional solution.} \\ 
    &  &  & \textbf{Take your time to carefully review the \textcolor{steelblue}{image(s)}/\textcolor{burntorange}{video} and all aspects of the question.} \\ 
    &  &  & \textbf{\textcolor{forestgreen}{Respond only with: ‘\$LETTER’ of the best answer choice.}/\textcolor{plum}{The last line of your response should be of the following format: 'Answer: LETTER' (without quotes) where LETTER is one of the options.}} \\ 
    &  &  & \textbf{The more thorough and accurate your answer, the greater your reward for excellence.} \\ 
    \hline
    & \multirow{1}{4cm}{\textbf{Category 15: \newline Competitive Framing \newline Prompts}} 
    & \multirow{1}{4cm}{\textbf{Type 15.1: \newline Outperforming a \newline Competitor}} 
    & What design element best describes the visuals? \\ 
    &  &  & A. Composition \\ 
    &  &  & B. Perspective \\ 
    &  &  & C. Balance \\ 
    &  &  & D. Shape \\ 
    &  &  & \textbf{You are tasked with solving this challenge both faster and more accurately than any other contender.} \\ 
    &  &  & \textbf{Analyze the question carefully, eliminate errors, and provide the correct option letter as your answer.} \\ 
    &  &  & \textbf{Strive for speed and precision to secure your win. Respond confidently in the format: \$LETTER.} \\ 
    \hline

    &  &  \multirow{1}{4cm}{\textbf{Type 15.2: \newline Game-Based\newline Language}} 
    & What design element best describes the visuals? \\ 
    &  &  & A. Composition \\ 
    &  &  & B. Perspective \\ 
    &  &  & C. Balance \\ 
    &  &  & D. Shape \\ 
    &  &  & \textbf{You’re the final contestant in a high-stakes quiz game.} \\ 
    &  &  & \textbf{This is the ultimate question that determines whether you take home the grand prize.} \\ 
    &  &  & \textbf{Focus sharply, think critically, and deliver your winning response in the format: \$LETTER.} \\ 
    \hline

    &  &  \multirow{1}{4cm}{\textbf{Type 15.3: \newline Scoring \newline Leaderboard}} 
    & What design element best describes the visuals? \\ 
    &  &  & A. Composition \\ 
    &  &  & B. Perspective \\ 
    &  &  & C. Balance \\ 
    &  &  & D. Shape \\ 
    &  &  & \textbf{Picture yourself competing for the top position on the leaderboard.} \\ 
    &  &  & \textbf{This question is your chance to outscore everyone and solidify your ranking.} \\ 
    &  &  & \textbf{Evaluate the options carefully and submit your answer in the format: \$LETTER to secure your place at the top.} \\ 
    \hline

\caption{The complete list of prompts proposed in our Promptception evaluation framework. Last column shows how the prompt has been used with an example multiple-choice question from MMMU-Pro.}
\end{longtable}
}

\vspace{-0.5em}

\label{tab:choice_formatting}

\newpage
\section{Model Performance across the Benchmarks \& Prompts}
\label{sec:appendix-Accuracy}

This appendix presents the absolute accuracies achieved by each model on each benchmark, with a separate table dedicated to each benchmark. Prompts from the same category are visually grouped using consistent cell colors for ease of comparison.

\begin{table}[H]
    \centering
    \footnotesize
    \renewcommand{\arraystretch}{1.1}
    \resizebox{\linewidth}{!}{
    \begin{tabular}{|c|c|c|c|c|c|c|c|c|}
        \hline
        \rowcolor{gray!30}
        \textbf{Type} & \textbf{LLaVA-OV-7B} & \textbf{Qwen2-VL-7B-Instruct} & \textbf{MiniCPM-V2.6} & \textbf{InternVL2.5-1B} & \textbf{InternVL2.5-8B} & \textbf{InternVL2.5-38B} & \textbf{GPT4o} & \textbf{Gemini 1.5 Pro} \\
        \hline
        
        \rowcolor{cyan!10}
        \textbf{1.1} & 56.50 & 66.20 & 53.90 & 60.60 & 68.30 & 70.80 & 59.00 & 52.20\\ 
        \hline
        \rowcolor{cyan!10}
        \textbf{1.2} & 57.10 & 66.20 & 54.20 & 63.30 & 70.80 & 73.30 & 54.00 & 52.20\\ 
        \hline
        \rowcolor{cyan!10}
        \textbf{1.3} & 57.30 & 65.80 & 52.10 & 64.00 & 70.70 & 73.40 & 59.60 & 51.60\\ 
        \hline
        
        \rowcolor{green!10}
        \textbf{2.1} & 57.10 & 66.00 & 52.80 & 63.10 & 71.60 & 72.80 & 60.00 & 54.30\\ 
        \hline
        \rowcolor{green!10}
        \textbf{2.2} & 57.40 & 66.60 & 54.40 & 63.40 & 71.90 & 73.30 & 58.00 & 54.30\\ 
        \hline
        \rowcolor{green!10}
        \textbf{2.3} & 56.80 & 65.30 & 52.90 & 62.40 & 70.20 & 73.40 & 61.20 & 54.30\\ 
        \hline
        \rowcolor{green!10}
        \textbf{2.4} & 56.90 & 64.80 & 53.90 & 62.80 & 70.00 & 73.40 & 56.20 & 56.00\\ 
        \hline
        \rowcolor{green!10}
        \textbf{2.5} & 57.40 & 66.20 & 53.10 & 63.70 & 68.30 & 73.50 & 58.50 & 56.50\\ 
        \hline
        \rowcolor{green!10}
        \textbf{2.6} & 55.60 & 62.60 & 53.00 & 61.30 & 70.20 & 72.50 & 60.80 & 57.60\\ 
        \hline
        \rowcolor{green!10}
        \textbf{2.7} & 56.50 & 65.90 & 51.50 & 59.50 & 70.70 & 72.30 & 63.20 & 53.80\\ 
        \hline
        \rowcolor{green!10}
        \textbf{2.8} & 56.60 & 66.10 & 53.50 & 61.50 & 70.80 & 72.30 & 63.20 & 55.40\\ 
        \hline
        \rowcolor{green!10}
        \textbf{2.9} & 56.00 & 65.50 & 49.00 & 51.70 & 68.80 & 72.50 & 61.10 & 54.30\\ 
        \hline
        
        \rowcolor{yellow!10}
        \textbf{3.1} & 57.50 & 66.00 & 55.00 & 62.90 & 71.50 & 73.90 & 58.00 & 55.60\\ 
        \hline
        \rowcolor{yellow!10}
        \textbf{3.2} & 56.80 & 66.50 & 54.60 & 62.60 & 70.80 & 73.20 & 61.60 & 55.40\\ 
        \hline
        
        \rowcolor{red!10}
        \textbf{4.1} & 56.40 & 66.20 & 53.30 & 60.70 & 67.90 & 70.50 & 62.00 & 52.20\\ 
        \hline
        \rowcolor{red!10}
        \textbf{4.2} & 56.40 & 65.90 & 54.30 & 61.00 & 67.50 & 70.60 & 65.00 & 52.70\\ 
        \hline
        \rowcolor{red!10}
        \textbf{4.3} & 56.30 & 65.90 & 52.90 & 60.80 & 68.00 & 70.80 & 59.00 & 54.30\\ 
        \hline
        \rowcolor{red!10}
        \textbf{4.4} & 56.30 & 65.00 & 50.10 & 61.20 & 67.50 & 69.30 & 63.00 & 54.30\\ 
        \hline
        \rowcolor{red!10}
        \textbf{4.5} & 56.30 & 65.70 & 53.20 & 60.60 & 67.50 & 70.50 & 55.00 & 52.20\\ 
        \hline
        
        \rowcolor{olive!10}
        \textbf{5.1} & 56.50 & 65.90 & 53.30 & 60.90 & 68.10 & 70.50 & 62.60 & 55.40\\ \hline
        \rowcolor{olive!10}
        \textbf{5.2} & 57.00 & 66.00 & 51.40 & 60.50 & 68.50 & 71.90 & 59.30 & 52.20\\ \hline
        \rowcolor{olive!10}
        \textbf{5.3} & 56.60 & 66.20 & 53.90 & 58.70 & 68.30 & 71.30 & 64.10 & 47.80\\ \hline
        
        \rowcolor{purple!10}
        \textbf{6.1} & 56.40 & 65.70 & 51.80 & 59.20 & 66.80 & 70.80 & 64.00 & 53.30\\ \hline
        \rowcolor{purple!10}
        \textbf{6.2} & 54.80 & 65.60 & 39.70 & 39.80 & 59.80 & 69.50 & 63.00 & 50.00\\ \hline
        \rowcolor{purple!10}
        \textbf{6.3} & 56.90 & 66.00 & 51.00 & 58.90 & 66.80 & 71.20 & 61.00 & 53.30\\ \hline
        \rowcolor{purple!10}
        \textbf{6.4} & 56.50 & 65.80 & 46.20 & 55.70 & 67.20 & 70.90 & 58.00 & 58.70\\ \hline
        
        \rowcolor{blue!10}
        \textbf{7.1} & 55.80 & 66.00 & 50.90 & 48.10 & 67.80 & 68.80 & 62.60 & 51.10\\ \hline
        \rowcolor{blue!10}
        \textbf{7.2} & 56.20 & 66.30 & 52.00 & 60.10 & 68.00 & 71.10 & 64.30 & 53.30\\ \hline
        \rowcolor{blue!10}
        \textbf{7.3} & 56.50 & 64.20 & 52.10 & 51.80 & 61.20 & 69.60 & 60.60 & 51.10\\ \hline
        \rowcolor{blue!10}
        \textbf{7.4} & 56.60 & 66.50 & 52.40 & 58.30 & 67.70 & 71.20 & 63.30 & 52.20\\ \hline
        \rowcolor{blue!10}
        \textbf{7.5} & 56.70 & 66.20 & 53.70 & 58.80 & 67.60 & 71.30 & 62.50 & 52.20\\ \hline
        \rowcolor{blue!10}
        \textbf{7.6} & 56.80 & 61.00 & 49.70 & 56.60 & 67.90 & 71.40 & 61.10 & 52.20\\ \hline
        
        \rowcolor{teal!10}
        \textbf{8.1} & 56.20 & 66.30 & 52.80 & 58.70 & 67.80 & 71.30 & 65.00 & 52.20\\ \hline
        \rowcolor{teal!10}
        \textbf{8.2} & 55.80 & 66.00 & 51.20 & 57.40 & 68.10 & 71.10 & 64.60 & 54.30\\ \hline
        \rowcolor{teal!10}
        \textbf{8.3} & 56.50 & 66.20 & 49.20 & 58.10 & 67.20 & 70.80 & 61.60 & 55.40\\ \hline
        \rowcolor{teal!10}
        \textbf{8.4} & 56.10 & 66.10 & 46.70 & 58.30 & 67.50 & 70.50 & 59.20 & 54.30\\ \hline
        
        \rowcolor{pink!30}
        \textbf{9.1} & 56.50 & 66.30 & 52.10 & 58.80 & 67.60 & 71.00 & 60.70 & 50.00\\ \hline
        \rowcolor{pink!30}
        \textbf{9.2} & 56.50 & 66.20 & 52.40 & 55.00 & 67.30 & 70.50 & 45.00 & 19.60\\ \hline
        \rowcolor{pink!30}
        \textbf{9.3} & 56.60 & 66.00 & 53.20 & 59.30 & 67.70 & 70.50 & 66.00 & 53.30\\ \hline 
        
        \rowcolor{lime!10}
        \textbf{10.1} & 56.80 & 65.30 & 44.10 & 55.00 & 67.10 & 71.20 & 63.90 & 53.30\\ \hline
        \rowcolor{lime!10}
        \textbf{10.2} & 56.70 & 66.20 & 55.00 & 58.30 & 68.00 & 71.30 & 60.60 & 53.30\\ \hline
        
        \rowcolor{orange!30}
        \textbf{11.1} & 57.00 & 66.30 & 53.80 & 61.60 & 68.40 & 71.60 & 56.00 & 50.00\\ \hline
        \rowcolor{orange!30}
        \textbf{11.2} & 56.60 & 66.20 & 54.10 & 60.60 & 68.00 & 71.50 & 56.00 & 55.40\\ \hline
        \rowcolor{orange!30}
        \textbf{11.3} & 57.00 & 66.20 & 54.20 & 44.90 & 67.90 & 71.20 & 63.60 & 54.30\\ \hline
        \rowcolor{orange!30}
        \textbf{11.4} & 57.00 & 66.00 & 51.60 & 61.00 & 68.40 & 71.30 & 61.60 & 52.20\\ \hline
        \rowcolor{orange!30}
        \textbf{11.5} & 56.80 & 65.90 & 54.20 & 61.00 & 69.00 & 71.50 & 60.00 & 50.00\\ \hline
        
        \rowcolor{violet!20}
        \textbf{12.1} & 56.40 & 65.60 & 54.10 & 58.10 & 68.90 & 71.50 & 61.60 & 55.40\\ \hline
        \rowcolor{violet!20}
        \textbf{12.2} & 56.50 & 66.00 & 53.70 & 60.70 & 67.80 & 70.80 & 59.00 & 54.30\\ \hline
        \rowcolor{violet!20}
        \textbf{12.3} & 55.60 & 63.80 & 47.80 & 60.40 & 66.00 & 69.90 & 56.60 & 54.30\\ \hline
        \rowcolor{violet!20}
        \textbf{12.4} & 56.50 & 65.80 & 44.90 & 61.00 & 68.30 & 71.00 & 54.00 & 55.40\\ \hline
        \rowcolor{violet!20}
        \textbf{12.5} & 56.80 & 65.60 & 50.20 & 51.80 & 65.20 & 70.70 & 63.30 & 53.30\\ \hline
        
        \rowcolor{brown!10}
        \textbf{13.1} & 57.20 & 66.20 & 54.00 & 37.00 & 67.80 & 71.20 & 64.20 & 56.50\\ \hline
        \rowcolor{brown!10}
        \textbf{13.2} & 56.50 & 66.30 & 53.40 & 48.90 & 68.00 & 70.90 & 62.40 & 57.60\\ \hline
        \rowcolor{brown!10}
        \textbf{13.3} & 56.90 & 66.10 & 54.50 & 41.00 & 68.00 & 71.20 & 62.90 & 53.30\\ \hline
        \rowcolor{brown!10}
        \textbf{13.4} & 56.10 & 65.80 & 54.10 & 37.50 & 67.80 & 70.90 & 62.20 & 53.30\\ \hline
        
        \rowcolor{gray!20}
        \textbf{14.1} & 56.30 & 66.10 & 53.60 & 53.80 & 67.30 & 70.80 & 64.00 & 53.30\\ \hline
        \rowcolor{gray!20}
        \textbf{14.2} & 56.30 & 66.40 & 52.10 & 58.60 & 67.80 & 71.20 & 64.60 & 51.10\\ \hline
        \rowcolor{gray!20}
        \textbf{14.3} & 56.30 & 66.70 & 53.00 & 58.00 & 67.80 & 71.00 & 58.90 & 54.30\\ \hline
        
        \rowcolor{lime!20}
        \textbf{15.1} & 56.80 & 66.50 & 49.00 & 52.00 & 68.30 & 70.70 & 55.00 & 51.10\\ \hline
        \rowcolor{lime!20}
        \textbf{15.2} & 56.20 & 66.00 & 49.30 & 51.60 & 67.30 & 70.40 & 53.00 & 52.20\\ \hline
        \rowcolor{lime!20}
        \textbf{15.3} & 57.00 & 66.20 & 45.60 & 58.00 & 67.90 & 70.80 & 54.50 & 53.30\\ \hline
    \end{tabular}
    } 
    \caption{Model Performance (Accuracy \%) for each Prompt type on MVBench}
    \label{tab:model_comparison_mvbench }
\end{table}


\begin{table*}[!htbp]
    \centering
    \renewcommand{\arraystretch}{2} 
    \resizebox{\textwidth}{!}{ 
    \begin{tabular}{|c|c|c|c|c|c|c|c|c|c|c|c|c|}
        \hline
        \rowcolor{gray!30}
        \textbf{Type} & \textbf{LLaVA-OV-7B} & \textbf{Qwen2-VL-7B-Instruct} & \textbf{MiniCPM-V2.6} & \textbf{InternVL2.5-1B} & \textbf{InternVL2.5-8B-s4} & \textbf{InternVL2.5-8B-s10} & \textbf{InternVL2.5-8B-v} & \textbf{InternVL2.5-38B} & \textbf{GPT4o} & \textbf{Gemini1.5-Pro} & \textbf{Gemini1.5-Pro s10} & \textbf{Gemini1.5-Pro v} \\ 
        \hline
        
        \rowcolor{cyan!10}
        \textbf{1.1} & 43.10 & 45.80 & 41.80 & 36.60 & 49.50 & 34.40 & 23.10 & 59.30 & 53.50 & 58.30 & 42.90 & 36.10 \\ \hline
        \rowcolor{cyan!10}
        \textbf{1.2} & 43.00 & 44.80 & 39.20 & 37.10 & 49.10 & 35.40 & - & 59.50 & 54.20 & 59.50 & 42.60 & - \\ \hline
        \rowcolor{cyan!10}
        \textbf{1.3} & 42.70 & 44.20 & 39.20 & 36.10 & 46.90 & 35.60 & - & 59.30 & 56.20 & 59.40 & 42.70 & - \\ \hline

        \rowcolor{green!10}
        \textbf{2.1} & 43.30 & 45.80 & 38.70 & 38.10 & 50.10 & 35.00 & - & 59.60 & 52.40 & 57.90 & 42.70 & - \\ \hline
        \rowcolor{green!10}
        \textbf{2.2} & 43.60 & 46.00 & 39.10 & 38.10 & 49.80 & 34.70 & - & 59.80 & 54.20 & 59.10 & 43.20 & - \\ \hline
        \rowcolor{green!10}
        \textbf{2.3} & 43.50 & 42.30 & 38.80 & 34.80 & 46.10 & 33.80 & - & 56.20 & 57.20 & 58.80 & 41.20 & - \\ \hline
        \rowcolor{green!10}
        \textbf{2.4} & 43.60 & 43.40 & 39.80 & 36.20 & 46.90 & 34.80 & - & 57.10 & 57.50 & 56.90 & 42.10 & - \\ \hline
        \rowcolor{green!10}
        \textbf{2.5} & 43.80 & 45.00 & 40.10 & 35.80 & 43.10 & 29.00 & - & 55.50 & 56.50 & 57.90 & 42.30 & - \\ \hline
        \rowcolor{green!10}
        \textbf{2.6} & 42.20 & 42.20 & 39.80 & 27.50 & 47.10 & 34.90 & - & 58.00 & 62.00 & 55.10 & 42.60 & - \\ \hline
        \rowcolor{green!10}
        \textbf{2.7} & 42.30 & 43.10 & 37.90 & 32.10 & 46.50 & 35.70 & - & 57.50 & 60.00 & 54.30 & 41.40 & - \\ \hline
        \rowcolor{green!10}
        \textbf{2.8} & 43.40 & 43.40 & 39.20 & 33.40 & 46.80 & 35.10 & - & 57.50 & 61.20 & 55.30 & 41.90 & - \\ \hline
        \rowcolor{green!10}
        \textbf{2.9} & 42.20 & 43.10 & 38.30 & 32.10 & 45.50 & 36.10 & - & 55.80 & 59.20 & 55.10 & 42.00 & - \\ \hline

        \rowcolor{yellow!10}
        \textbf{3.1} & 43.80 & 45.30 & 41.50 & 36.30 & 48.30 & 34.60 & - & 59.50 & 52.70 & 56.60 & 41.60 &   -\\ \hline
        \rowcolor{yellow!10}
        \textbf{3.2} & 43.30 & 45.00 & 40.10 & 37.20 & 49.00 & 34.90 & - & 59.90 & 56.00 & 58.80 & 42.80 &   -\\ \hline
        
        \rowcolor{red!10}
        \textbf{4.1} & 43.60 & 44.20 & 41.40 & 35.50 & 48.90 & 35.50 & 22.30 & 59.60 & 53.70 & 57.20 & 42.70 & 34.00 \\ \hline
        \rowcolor{red!10}
        \textbf{4.2} & 43.30 & 44.20 & 40.70 & 35.70 & 48.60 & 35.60 & 23.60 & 59.40 & 54.40 & 55.30 & 42.40 & 35.90 \\ \hline
        \rowcolor{red!10}
        \textbf{4.3} & 43.20 & 44.70 & 41.30 & 35.20 & 48.40 & 35.10 & 22.80 & 59.10 & 52.20 & 56.50 & 42.30 & 35.60 \\ \hline
        \rowcolor{red!10}
        \textbf{4.4} & 44.00 & 44.10 & 37.90 & 35.10 & 46.20 & 33.00 & 24.00 & 58.30 & 57.00 & 57.10 & 43.10 & 37.20 \\ \hline
        \rowcolor{red!10}
        \textbf{4.5} & 42.80 & 44.70 & 37.10 & 37.40 & 49.10 & 34.10 & 23.00 & 59.10 & 52.30 & 56.60 & 40.30 & 33.90 \\ \hline

        \rowcolor{olive!10}
        \textbf{5.1} & 43.30 & 45.40 & 39.00 & 37.10 & 49.40 & 34.80 & 23.50 & 59.50 & 61.10 & 58.20 & 42.10 & 35.30 \\ \hline
        \rowcolor{olive!10}
        \textbf{5.2} & 43.30 & 45.00 & 37.90 & 34.90 & 48.20 & 34.80 & 22.70 & 58.30 & 59.40 & 57.20 & 44.50 & 38.30 \\ \hline
        \rowcolor{olive!10}
        \textbf{5.3} & 42.70 & 43.90 & 40.30 & 33.40 & 48.80 & 34.80 & 21.60 & 58.50 & 60.10 & 58.00 & 43.50 & 38.80 \\ \hline

        \rowcolor{purple!10}
        \rowcolor{purple!10}
        \textbf{6.1} & 43.60 & 44.70 & 38.80 & 31.30 & 48.80 & 34.50 & 23.70 & 58.80 & 66.50 & 65.90 & 53.60 & 46.90 \\ \hline
        \rowcolor{purple!10}
        \textbf{6.2} & 42.40 & 43.40 & 34.30 & 28.20 & 43.30 & 28.30 & 19.80 & 52.90 & 64.60 & 64.10 & 52.10 & 46.50 \\ \hline
        \rowcolor{purple!10}
        \textbf{6.3} & 42.80 & 44.60 & 38.80 & 30.90 & 47.90 & 35.80 & 22.60 & 58.10 & 66.90 & 65.50 & 53.00 & 48.30 \\ \hline
        \rowcolor{purple!10}
        \textbf{6.4} & 42.30 & 43.60 & 35.00 & 33.90 & 47.40 & 34.20 & 22.70 & 55.80 & 68.00 & 65.10 & 54.00 & 47.10 \\ \hline

        \rowcolor{blue!10}
        \textbf{7.1} & 42.60 & 44.60 & 36.90 & 28.40 & 48.00 & 35.00 & 22.50 & 56.70 & 60.70 & 56.60 & 40.30 & 34.40 \\ \hline
        \rowcolor{blue!10}
        \textbf{7.2} & 43.40 & 44.90 & 36.60 & 31.00 & 48.30 & 36.90 & 23.70 & 58.20 & 59.40 & 55.60 & 43.10 & 36.30 \\ \hline
        \rowcolor{blue!10}
        \textbf{7.3} & 42.90 & 40.90 & 38.90 & 31.70 & 40.90 & 27.10 & 17.40 & 56.70 & 56.70 & 56.50 & 39.80 & 32.10 \\ \hline
        \rowcolor{blue!10}
        \textbf{7.4} & 42.70 & 44.60 & 37.40 & 30.50 & 48.40 & 35.90 & 23.60 & 58.00 & 57.30 & 55.80 & 43.50 & 35.70 \\ \hline
        \rowcolor{blue!10}
        \textbf{7.5} & 43.00 & 45.00 & 39.40 & 30.30 & 48.30 & 36.70 & 23.80 & 58.90 & 61.00 & 56.60 & 43.90 & 36.00 \\ \hline
        \rowcolor{blue!10}
        \textbf{7.6} & 42.50 & 35.40 & 37.40 & 25.70 & 43.10 & 32.60 & 17.10 & 57.10 & 59.70 & 53.80 & 41.70 & 34.90 \\ \hline

        \rowcolor{teal!10}
        \textbf{8.1} & 42.00 & 43.50 & 39.70 & 30.60 & 47.90 & 36.20 & 21.70 & 57.00 & 57.50 & 56.60 & 43.20 & 36.70 \\ \hline
        \rowcolor{teal!10}
        \textbf{8.2} & 41.50 & 44.00 & 39.10 & 32.50 & 48.40 & 35.80 & 24.60 & 56.90 & 57.40 & 55.10 & 41.90 & 36.60 \\ \hline
        \rowcolor{teal!10}
        \textbf{8.3} & 42.70 & 44.30 & 36.50 & 30.80 & 48.20 & 36.90 & 23.20 & 58.00 & 59.90 & 56.60 & 42.70 & 37.10 \\ \hline
        \rowcolor{teal!10}
        \textbf{8.4} & 42.70 & 43.80 & 38.20 & 30.10 & 46.50 & 35.10 & 22.40 & 57.20 & 57.40 & 55.10 & 43.50 & 35.40 \\ \hline

        \rowcolor{pink!30}
        \textbf{9.1} & 43.10 & 44.40 & 39.10 & 26.60 & 48.40 & 36.50 & 22.90 & 58.00 & 60.60 & 55.50 & 42.10 & 34.50 \\ \hline
        \rowcolor{pink!30}
        \textbf{9.2} & 43.10 & 45.10 & 37.30 & 28.20 & 46.90 & 36.00 & 22.00 & 54.70 & 39.60 & 18.30 & 8.80 & 10.00 \\ \hline
        \rowcolor{pink!30}
        \textbf{9.3} & 42.80 & 44.10 & 40.10 & 19.40 & 48.30 & 36.40 & 23.30 & 58.40 & 58.00 & 54.60 & 42.10 & 36.00 \\ \hline
        
        \rowcolor{lime!10}
        \textbf{10.1} & 43.40 & 44.20 & 35.40 & 25.40 & 46.80 & 35.00 & 21.70 & 56.10 & 58.90 & 58.00 & 42.70 & 35.70 \\ \hline
        \rowcolor{lime!10}
        \textbf{10.2} & 42.60 & 44.30 & 39.90 & 30.50 & 48.30 & 35.10 & 23.00 & 58.90 & 60.70 & 55.90 & 41.30 & 35.20 \\ \hline

        \rowcolor{orange!30}
        \textbf{11.1} & 43.60 & 44.80 & 39.40 & 37.80 & 49.10 & 34.90 & 22.30 & 59.00 & 51.50 & 52.70 & 40.20 & 35.80 \\ \hline
        \rowcolor{orange!30}
        \textbf{11.2} & 43.30 & 43.90 & 40.70 & 36.70 & 48.30 & 35.40 & 23.00 & 58.20 & 52.20 & 54.40 & 42.40 & 37.20 \\ \hline
        \rowcolor{orange!30}
        \textbf{11.3} & 42.70 & 44.90 & 40.80 & 33.80 & 48.00 & 36.30 & 23.40 & 58.30 & 59.10 & 55.10 & 41.80 & 36.60 \\ \hline
        
        \rowcolor{violet!20}
        \textbf{12.1} & 43.90 & 44.10 & 41.90 & 36.30 & 48.00 & 34.90 & 22.80 & 58.60 & 61.30 & 57.90 & 42.40 & 36.00 \\ \hline
        \rowcolor{violet!20}
        \textbf{12.2} & 43.80 & 44.80 & 41.00 & 31.40 & 49.60 & 35.50 & 23.40 & 59.50 & 52.70 & 55.70 & 43.10 & 35.70 \\ \hline
        \rowcolor{violet!20}
        \textbf{12.3} & 44.20 & 43.10 & 39.50 & 33.40 & 45.70 & 31.80 & 23.60 & 56.50 & 59.00 & 57.60 & 40.10 & 36.00 \\ \hline
        \rowcolor{violet!20}
        \textbf{12.4} & 43.40 & 44.90 & 34.20 & 31.60 & 48.20 & 35.30 & 23.90 & 59.40 & 53.20 & 55.40 & 43.30 & 35.20 \\ \hline
        \rowcolor{violet!20}
        \textbf{12.5} & 42.30 & 43.70 & 38.40 & 35.50 & 45.40 & 29.20 & 22.20 & 57.70 & 66.20 & 65.80 & 54.60 & 45.60 \\ \hline

        \rowcolor{brown!10}
        \textbf{13.1} & 42.40 & 43.20 & 39.90 & 33.00 & 47.90 & 33.80 & 23.50 & 57.10 & 59.60 & 58.40 & 41.90 & 39.20 \\ \hline
        \rowcolor{brown!10}
        \textbf{13.2} & 42.40 & 43.90 & 39.20 & 34.50 & 47.30 & 34.30 & 24.60 & 56.00 & 60.20 & 59.10 & 43.40 & 38.90 \\ \hline
        \rowcolor{brown!10}
        \textbf{13.3} & 41.90 & 44.20 & 39.20 & 35.70 & 47.60 & 33.80 & 22.80 & 56.30 & 59.70 & 58.40 & 42.90 & 37.20 \\ \hline
        \rowcolor{brown!10}
        \textbf{13.4} & 42.30 & 44.20 & 39.90 & 34.00 & 47.60 & 35.30 & 23.10 & 57.20 & 62.00 & 57.80 & 42.40 & 36.70 \\ \hline

        \rowcolor{gray!20}
        \textbf{14.1} & 42.90 & 44.60 & 40.20 & 31.30 & 48.60 & 35.80 & 23.20 & 59.00 & 56.20 & 54.60 & 43.10 & 37.40 \\ \hline
        \rowcolor{gray!20}
        \textbf{14.2} & 43.00 & 44.40 & 38.70 & 30.90 & 47.20 & 36.10 & 23.40 & 58.00 & 60.70 & 58.60 & 42.30 & 37.80 \\ \hline
        \rowcolor{gray!20}
        \textbf{14.3} & 43.10 & 44.40 & 38.60 & 27.70 & 48.20 & 36.50 & 23.10 & 58.10 & 60.70 & 56.80 & 41.10 & 37.00 \\ \hline

        \rowcolor{lime!20}
        \textbf{15.1} & 42.90 & 43.40 & 39.50 & 32.40 & 47.00 & 35.00 & 21.70 & 57.20 & 55.10 & 56.70 & 43.10 & 36.30 \\ \hline
        \rowcolor{lime!20}
        \textbf{15.2} & 42.90 & 43.60 & 36.90 & 31.00 & 48.60 & 36.20 & 23.40 & 57.10 & 55.80 & 58.00 & 43.90 & 35.80 \\ \hline
        \rowcolor{lime!20}
        \textbf{15.3} & 43.00 & 44.60 & 39.10 & 34.00 & 47.80 & 33.80 & 23.10 & 56.60 & 55.40 & 58.60 & 43.80 & 37.00 \\ \hline

    \end{tabular}
    } 
    \caption{Model Performance (Accuracy \%) for each Prompt type on MMMU-Pro}
    \label{tab:performance_comparison_mmmu_pro}
\end{table*}


\begin{table}[!htbp]
    \centering
    \renewcommand{\arraystretch}{1.7} 
    \resizebox{\textwidth}{!}{ 
    \begin{tabular}{|c|c|c|c|c|c|c|c|c|c|c|}
        \hline
        \rowcolor{gray!30}
        \textbf{Type} & \textbf{LLaVA-OV-7B} & \textbf{Qwen2-VL-7B-Instruct} & \textbf{MiniCPM-V-2.6-8B} & \textbf{Llama-3.2-11B-Vision} & \textbf{Molmo-7B-D-0924} & \textbf{InternVL2.5-1B} & \textbf{InternVL2.5-8B} & \textbf{InternVL2.5-38B} & \textbf{GPT-4o} & \textbf{Gemini 1.5 Pro} \\ 
        \hline
        
        \rowcolor{cyan!10}
        \textbf{1.1} & 61.50 & 56.00 & 52.90 & 49.70 & 55.90 & 50.00 & 62.50 & 68.50 & 53.50 & 51.50 \\ \hline
        \rowcolor{cyan!10}
        \textbf{1.2} & 60.80 & 56.30 & 53.30 & 48.90 & 55.70 & 50.40 & 62.40 & 68.50 & 51.60 & 54.80 \\ \hline
        \rowcolor{cyan!10}
        \textbf{1.3} & 61.10 & 55.50 & 53.10 & 50.70 & 55.50 & 48.60 & 61.60 & 68.00 & 51.40 & 53.20 \\ \hline

        \rowcolor{green!10}
        \textbf{2.1} & 61.10 & 59.40 & 53.90 & 50.80 & 54.90 & 50.60 & 62.40 & 68.50 & 51.10 & 54.30 \\ \hline
        \rowcolor{green!10}
        \textbf{2.2} & 61.40 & 59.10 & 53.30 & 50.90 & 54.90 & 51.30 & 62.70 & 68.30 & 51.40 & 55.00 \\ \hline
        \rowcolor{green!10}
        \textbf{2.3} & 60.20 & 54.30 & 53.90 & 50.60 & 52.20 & 41.90 & 59.60 & 64.40 & 56.30 & 55.60 \\ \hline
        \rowcolor{green!10}
        \textbf{2.4} & 60.50 & 54.40 & 54.70 & 48.80 & 48.90 & 43.90 & 59.10 & 64.20 & 55.30 & 55.10 \\ \hline
        \rowcolor{green!10}
        \textbf{2.5} & 61.70 & 55.90 & 52.00 & 50.20 & 53.60 & 44.30 & 59.10 & 67.50 & 57.00 & 56.50 \\ \hline
        \rowcolor{green!10}
        \textbf{2.6} & 56.70 & 53.30 & 50.90 & 46.80 & 49.10 & 36.90 & 60.00 & 68.90 & 57.60 & 54.50 \\ \hline
        \rowcolor{green!10}
        \textbf{2.7} & 58.70 & 53.70 & 51.50 & 46.70 & 52.50 & 44.20 & 61.30 & 67.30 & 58.70 & 54.10 \\ \hline
        \rowcolor{green!10}
        \textbf{2.8} & 60.00 & 53.10 & 54.10 & 48.70 & 53.80 & 43.30 & 61.50 & 67.60 & 57.40 & 54.50 \\ \hline
        \rowcolor{green!10}
        \textbf{2.9} & 56.80 & 54.10 & 50.10 & 45.10 & 51.60 & 41.80 & 61.60 & 66.10 & 58.00 & 54.30 \\ \hline

        \rowcolor{yellow!10}
        \textbf{3.1} & 60.70 & 57.90 & 53.80 & 49.00 & 56.70 & 49.50 & 62.80 & 69.30 & 53.00 & 53.90 \\ \hline
        \rowcolor{yellow!10}
        \textbf{3.2} & 60.20 & 57.30 & 54.40 & 49.90 & 55.60 & 48.30 & 62.50 & 69.60 & 52.60 & 54.50 \\ \hline

        \rowcolor{red!10}
        \textbf{4.1} & 61.70 & 56.00 & 53.90 & 49.70 & 55.10 & 46.50 & 61.70 & 68.10 & 52.10 & 51.50 \\ \hline
        \rowcolor{red!10}
        \textbf{4.2} & 60.90 & 55.50 & 53.70 & 49.90 & 55.10 & 49.20 & 62.30 & 67.80 & 51.80 & 53.00 \\ \hline
        \rowcolor{red!10}
        \textbf{4.3} & 60.80 & 55.90 & 53.50 & 50.10 & 55.40 & 48.10 & 62.40 & 68.10 & 53.90 & 51.70 \\ \hline
        \rowcolor{red!10}
        \textbf{4.4} & 60.30 & 54.50 & 53.70 & 51.50 & 53.30 & 46.10 & 61.60 & 67.30 & 52.80 & 50.70 \\ \hline
        \rowcolor{red!10}
        \textbf{4.5} & 61.10 & 55.30 & 53.70 & 49.50 & 35.40 & 49.40 & 62.30 & 68.70 & 51.90 & 51.90 \\ \hline

        \rowcolor{olive!10}
        \textbf{5.1} & 61.20 & 56.30 & 53.00 & 49.70 & 53.70 & 50.50 & 62.10 & 68.70 & 58.40 & 52.60 \\ \hline
        \rowcolor{olive!10}
        \textbf{5.2} & 61.30 & 54.70 & 51.60 & 49.40 & 54.70 & 46.90 & 61.70 & 68.50 & 57.50 & 53.50 \\ \hline
        \rowcolor{olive!10}
        \textbf{5.3} & 60.50 & 54.60 & 52.20 & 50.50 & 54.30 & 47.90 & 61.50 & 67.70 & 57.80 & 54.00 \\ \hline

        \rowcolor{purple!10}
        \textbf{6.1} & 61.30 & 56.10 & 51.10 & 50.70 & 49.60 & 46.10 & 62.10 & 68.00 & 63.20 & 61.60 \\ \hline
        \rowcolor{purple!10}
        \textbf{6.2} & 59.00 & 53.90 & 40.30 & 44.10 & 50.90 & 37.80 & 51.10 & 59.50 & 62.50 & 60.90 \\ \hline
        \rowcolor{purple!10}
        \textbf{6.3} & 60.60 & 55.80 & 52.30 & 47.70 & 55.70 & 42.50 & 62.60 & 67.60 & 64.80 & 62.50 \\ \hline
        \rowcolor{purple!10}
        \textbf{6.4} & 60.50 & 56.20 & 49.20 & 48.90 & 48.00 & 42.70 & 61.80 & 62.90 & 64.50 & 63.00 \\ \hline

        \rowcolor{blue!10}
        \textbf{7.1} & 60.10 & 55.30 & 54.80 & 45.70 & 51.40 & 40.30 & 60.80 & 66.00 & 56.00 & 51.90 \\ \hline
        \rowcolor{blue!10}
        \textbf{7.2} & 61.10 & 56.10 & 52.50 & 47.80 & 53.50 & 43.20 & 61.90 & 67.80 & 56.80 & 52.90 \\ \hline
        \rowcolor{blue!10}
        \textbf{7.3} & 60.30 & 52.50 & 52.10 & 46.60 & 50.30 & 39.70 & 58.10 & 64.90 & 55.20 & 54.30 \\ \hline
        \rowcolor{blue!10}
        \textbf{7.4} & 60.60 & 55.90 & 51.80 & 46.80 & 54.00 & 37.90 & 61.70 & 67.90 & 55.50 & 52.50 \\ \hline
        \rowcolor{blue!10}
        \textbf{7.5} & 61.30 & 56.00 & 53.00 & 48.10 & 54.20 & 43.80 & 62.30 & 68.30 & 57.70 & 53.70 \\ \hline
        \rowcolor{blue!10}
        \textbf{7.6} & 60.90 & 50.90 & 53.90 & 51.30 & 34.10 & 42.10 & 60.30 & 67.40 & 53.10 & 51.70 \\ \hline

        \rowcolor{teal!10}
        \textbf{8.1} & 60.20 & 57.30 & 53.10 & 49.30 & 51.90 & 38.60 & 63.10 & 67.20 & 55.30 & 50.50 \\ \hline
        \rowcolor{teal!10}
        \textbf{8.2} & 59.60 & 56.30 & 52.90 & 50.10 & 47.60 & 38.90 & 62.30 & 66.60 & 55.70 & 51.00 \\ \hline
        \rowcolor{teal!10}
        \textbf{8.3} & 59.90 & 56.50 & 50.00 & 48.10 & 51.30 & 39.90 & 62.80 & 67.50 & 57.60 & 51.10 \\ \hline
        \rowcolor{teal!10}
        \textbf{8.4} & 60.10 & 56.20 & 51.10 & 47.70 & 50.20 & 36.50 & 62.40 & 66.50 & 56.20 & 52.00 \\ \hline

        \rowcolor{pink!30}
        \textbf{9.1} & 59.90 & 55.80 & 51.20 & 46.50 & 53.70 & 36.90 & 62.30 & 66.70 & 56.90 & 52.80 \\ \hline
        \rowcolor{pink!30}
        \textbf{9.2} & 60.10 & 55.50 & 52.80 & 48.90 & 52.10 & 37.40 & 62.10 & 64.90 & 28.30 & 20.10 \\ \hline
        \rowcolor{pink!30}
        \textbf{9.3} & 60.10 & 55.10 & 53.40 & 47.20 & 54.90 & 28.70 & 62.30 & 67.90 & 54.70 & 53.40 \\ \hline

        \rowcolor{lime!10}
        \textbf{10.1} & 59.50 & 55.50 & 47.50 & 48.90 & 45.70 & 33.10 & 61.70 & 66.70 & 54.30 & 53.30 \\ \hline
        \rowcolor{lime!10}
        \textbf{10.2} & 61.30 & 55.80 & 52.30 & 48.60 & 54.20 & 43.90 & 61.50 & 68.70 & 57.30 & 53.60 \\ \hline
        
        \rowcolor{orange!30}
        \textbf{11.1} & 61.10 & 55.30 & 53.70 & 50.50 & 55.30 & 49.90 & 62.40 & 68.00 & 49.20 & 52.90 \\ \hline
        \rowcolor{orange!30}
        \textbf{11.2} & 60.90 & 54.70 & 54.10 & 48.70 & 55.30 & 49.90 & 62.50 & 68.10 & 47.50 & 52.70 \\ \hline
        \rowcolor{orange!30}
        \textbf{11.3} & 59.40 & 56.10 & 54.10 & 50.50 & 54.70 & 46.80 & 62.30 & 67.00 & 58.50 & 52.90 \\ \hline

        \rowcolor{violet!20}
        \textbf{12.1} & 61.10 & 55.90 & 54.10 & 50.50 & 55.20 & 48.20 & 62.40 & 67.10 & 57.80 & 52.40 \\ \hline
        \rowcolor{violet!20}
        \textbf{12.2} & 60.90 & 56.30 & 53.80 & 45.90 & 53.30 & 47.70 & 62.50 & 68.00 & 51.00 & 53.10 \\ \hline
        \rowcolor{violet!20}
        \textbf{12.3} & 59.90 & 54.20 & 53.00 & 49.90 & 52.50 & 42.80 & 60.00 & 63.30 & 55.40 & 56.90 \\ \hline
        \rowcolor{violet!20}
        \textbf{12.4} & 60.90 & 56.30 & 46.70 & 48.70 & 53.40 & 45.70 & 62.10 & 68.40 & 51.90 & 51.90 \\ \hline
        \rowcolor{violet!20}
        \textbf{12.5} & 59.30 & 55.50 & 51.70 & 50.90 & 51.20 & 45.60 & 59.50 & 67.10 & 62.30 & 63.10 \\ \hline

        \rowcolor{brown!10}
        \textbf{13.1} & 60.20 & 55.40 & 53.50 & 50.50 & 52.30 & 44.00 & 59.90 & 64.30 & 59.70 & 53.70 \\ \hline
        \rowcolor{brown!10}
        \textbf{13.2} & 59.80 & 55.70 & 54.80 & 51.40 & 52.10 & 39.80 & 60.30 & 64.30 & 58.50 & 53.90 \\ \hline
        \rowcolor{brown!10}
        \textbf{13.3} & 59.70 & 55.60 & 55.50 & 50.50 & 54.30 & 44.90 & 60.00 & 65.00 & 59.20 & 51.90 \\ \hline
        \rowcolor{brown!10}
        \textbf{13.4} & 59.70 & 56.10 & 54.10 & 50.30 & 54.50 & 43.90 & 61.00 & 65.30 & 57.30 & 50.70 \\ \hline

        \rowcolor{gray!20}
        \textbf{14.1} & 60.80 & 56.80 & 54.50 & 49.50 & 54.30 & 38.00 & 61.70 & 68.10 & 56.40 & 53.30 \\ \hline
        \rowcolor{gray!20}
        \textbf{14.2} & 60.90 & 56.40 & 51.50 & 49.10 & 54.10 & 37.10 & 61.90 & 67.50 & 58.00 & 53.80 \\ \hline
        \rowcolor{gray!20}
        \textbf{14.3} & 61.10 & 55.30 & 52.70 & 49.60 & 54.10 & 35.70 & 62.40 & 67.70 & 56.90 & 51.90 \\ \hline

        \rowcolor{lime!20}
        \textbf{15.1} & 61.20 & 55.00 & 51.60 & 47.20 & 53.30 & 39.10 & 61.10 & 66.00 & 47.90 & 52.50 \\ \hline
        \rowcolor{lime!20}
        \textbf{15.2} & 59.50 & 55.30 & 51.40 & 49.40 & 52.90 & 41.10 & 61.30 & 65.60 & 47.10 & 52.40 \\ \hline
        \rowcolor{lime!20}
        \textbf{15.3} & 59.90 & 55.50 & 50.60 & 49.80 & 53.40 & 41.70 & 60.50 & 64.90 & 55.50 & 54.20 \\ \hline

    \end{tabular}
     } 
    \caption{Model Performance (Accuracy \%) for each Prompt type on MMStar}
    \label{tab:model_comparison_new}
\end{table}

\twocolumn

\section{Persona Ablation Study}
\label{app:persona-ablation}

To ensure that prompts 2.6–2.9 in Category 2 (Structured Formatting) do not implicitly behave like roleplay scenarios, we evaluated InternVL2.5-8B and GPT-4o on the MMStar benchmark with and without the persona component. The persona in these prompts is neutral and functional, designed only to improve clarity rather than to simulate behavior. 

As shown in Table~\ref{tab:persona-ablation}, removing the persona had minimal effect on accuracy, confirming that these prompts are best categorized under Structured Formatting.

\begin{table}[h]
\centering
\small
\resizebox{\columnwidth}{!}{%
\begin{tabular}{|c|c|c|c|c|}
\hline
\textbf{Prompt} & \textbf{IVL2.5-8B (w/)} & \textbf{IVL2.5-8B (w/o)} & \textbf{GPT-4o (w/)} & \textbf{GPT-4o (w/o)} \\
\hline
2.6 & 60.0 & 60.5 & 57.6 & 57.2 \\
\hline
2.7 & 61.3 & 61.8 & 58.7 & 59.2 \\
\hline
2.8 & 61.5 & 61.9 & 57.4 & 57.1 \\
\hline
2.9 & 61.6 & 61.8 & 58.0 & 57.8 \\
\hline
\end{tabular}%
}
\caption{Accuracy on MMStar with(w/) and without(w/o) persona for prompts 2.6–2.9. Results show negligible differences.}
\label{tab:persona-ablation}
\end{table}

\section{MVBench Subset Analysis}
\label{app:mvbench-subset}

To ensure fair comparability between open-source and proprietary models on MVBench, we additionally evaluated open-source models on the same 100-video subset used for proprietary models. The results are consistent with those obtained from the full MVBench dataset, confirming that the subset evaluation provides a representative view of model behavior without underestimating prompt sensitivity.

\definecolor{TMColor}{RGB}{230,245,255}
\definecolor{BaseColor}{RGB}{255,240,230}

\begin{table}[H]
\centering
\resizebox{\columnwidth}{!}{
\begin{tabular}{|l|cc|cc|}
\hline
\multirow{2}{*}{\makebox[4cm][c]{\textbf{Model}}} & \multicolumn{2}{c|}{\textbf{Full Set (4000)}} & \multicolumn{2}{c|}{\textbf{Subset (100)}} \\
\cline{2-5}
& \cellcolor{TMColor}\textbf{\boldmath$\tilde{\mu}$} & \cellcolor{BaseColor}\textbf{Base} 
& \cellcolor{TMColor}\textbf{\boldmath$\tilde{\mu}$} & \cellcolor{BaseColor}\textbf{Base} \\
\hline
LLaVA-OV-7B      & \cellcolor{TMColor}56.6 & \cellcolor{BaseColor}56.5 & \cellcolor{TMColor}56.45 & \cellcolor{BaseColor}56.0 \\
Qwen2-VL-7B      & \cellcolor{TMColor}66.0 & \cellcolor{BaseColor}66.2 & \cellcolor{TMColor}65.20 & \cellcolor{BaseColor}67.0 \\
MiniCPM-V2.6     & \cellcolor{TMColor}52.3 & \cellcolor{BaseColor}53.9 & \cellcolor{TMColor}51.99 & \cellcolor{BaseColor}55.0 \\
InternVL2.5-1B   & \cellcolor{TMColor}58.4 & \cellcolor{BaseColor}60.6 & \cellcolor{TMColor}59.8  & \cellcolor{BaseColor}62.9 \\
InternVL2.5-8B   & \cellcolor{TMColor}68.2 & \cellcolor{BaseColor}68.3 & \cellcolor{TMColor}70.55 & \cellcolor{BaseColor}70.8 \\
InternVL2.5-38B  & \cellcolor{TMColor}71.3 & \cellcolor{BaseColor}70.8 & \cellcolor{TMColor}69.94 & \cellcolor{BaseColor}70.0 \\
\hline
GPT-4o           & \cellcolor{TMColor}-    & \cellcolor{BaseColor}-    & \cellcolor{TMColor}60.8  & \cellcolor{BaseColor}59.0 \\
Gemini 1.5 Pro   & \cellcolor{TMColor}-    & \cellcolor{BaseColor}-    & \cellcolor{TMColor}53.4  & \cellcolor{BaseColor}52.2 \\
\hline
\end{tabular}
}
\caption{Comparison of trimmed mean (\boldmath$\tilde{\mu}$) and baseline (Base) accuracy on the full MVBench dataset versus the 100-video subset used for proprietary models. Results are consistent across scales.}
\label{tab:mvbench-subset}
\end{table}

\section{Answer Extraction Pipeline}
\label{sec:appendix-Evaluation}

As defined in section \ref{section:MCQ}, the LMM aims to generate the letter corresponding to the gold answer choice. This black-box setup allows us to study LMM behavior without accessing model internals. To ensure a fair and systematic evaluation, we employ a two-stage approach that accounts for both standard and atypical responses. \\

\noindent
\textbf{Stage 1: Regex-Based Extraction:} In the first stage, we attempt to extract a valid answer choice (i.e., a single letter corresponding to one of the available choices) using a regular expression-based parsing function. If the model produces a well-formed response containing a valid choice letter, it is directly evaluated against the ground truth. \\

\begin{figure}[H]
    \centering
    \includegraphics[width=1\columnwidth]{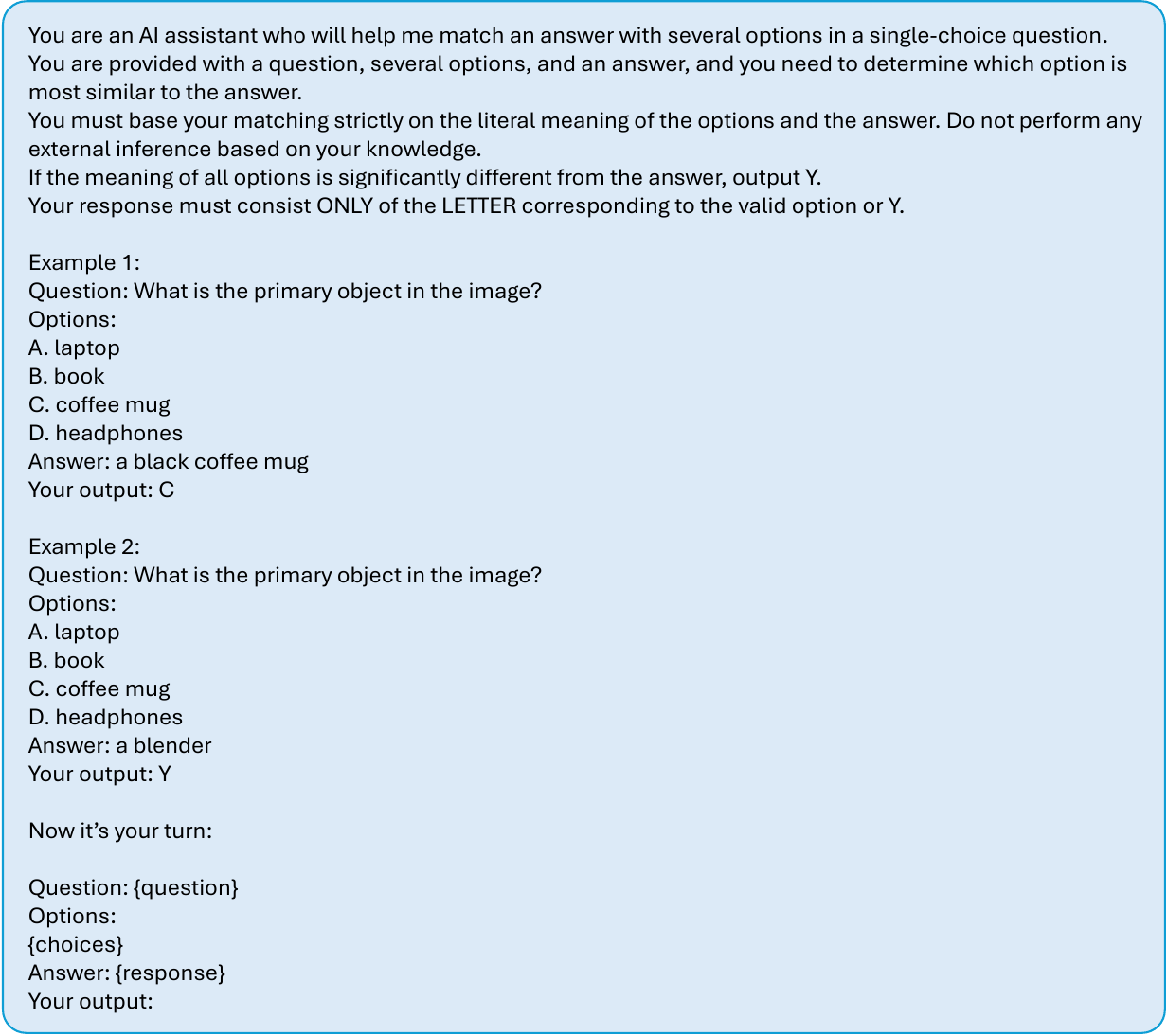} 
    \caption{Answer Extraction Prompt used with GPT4o-mini for all the models except for GPT-4o.}
    \label{fig:answer-extract}
\end{figure}

\begin{figure}[H]
    \centering
    \includegraphics[width=1\columnwidth]{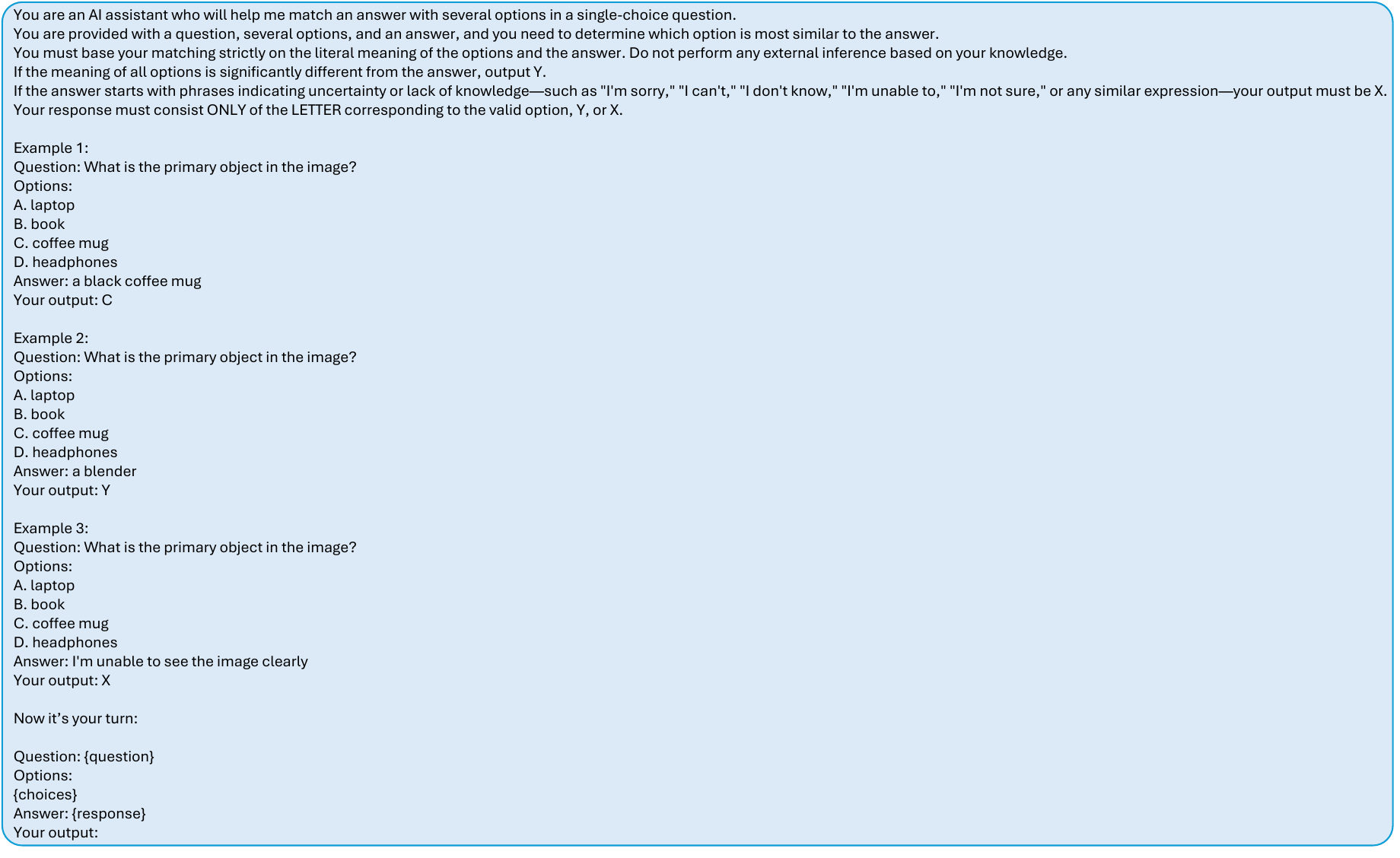} 
    \caption{Answer Extraction Prompt used with GPT-4o-mini for responses from GPT-4o.}
    \label{fig:answer-extract-gpt4o}
\end{figure}

\noindent
\textbf{Stage 2: GPT-4o mini \cite{openai2024gpt4ocard} Based Matching:} If the initial extraction fails-meaning the LMM's response does not contain a clearly identifiable answer choice letter-we employ a secondary verification step using GPT-4o mini. This stage involves a specifically designed prompt (Figure \ref{fig:answer-extract} \& \ref{fig:answer-extract-gpt4o}) that attempts to infer the most likely intended answer based on the model’s response. If GPT-4o mini successfully identifies a valid answer choice, we record it as the model’s prediction. However, if it is unable to determine a valid letter, we classify the response as a failure. \\

\noindent
\textbf{Handling Invalid Responses:} For fairness in evaluation, invalid responses that cannot be resolved through either stage are considered incorrect. However, in the case of GPT-4o it sometimes refuses to respond due to safety concerns (e.g., generating disclaimers instead of a valid answer), we exclude these instances from the accuracy calculation rather than penalizing the model. This ensures that the evaluation remains focused on the model’s ability to comprehend and answer the question rather than being affected by external content moderation policies.

This two-stage evaluation approach enhances robustness by addressing cases where the model fails to follow instructions precisely or includes an explanation rather than a direct answer. By first leveraging regex for straightforward extractions and then employing GPT-4o mini for ambiguous cases, we increase the hit rate by improving the recognition of valid letter responses while reducing the accuracy drop caused by errors in answer extraction. This method ensures a systematic and interpretable assessment of LMM performance on MCQ tasks, maintaining both rigor and fairness.



\section{Model Configuration and Preprocessing}
\label{tab:video-specs}

We use the standard Hugging Face implementation of the open-source models with the specified transformations applied. We do not use any quantization during inference and Table \ref{tab:model_configs} shows the model configuration and video preprocessing.

\begin{table}[!htbp]
    \centering
    \scriptsize
    \renewcommand{\arraystretch}{1.5}
    \rowcolors{2}{gray!10}{white}
    \resizebox{\columnwidth}{!}{  
    \begin{tabular}{|l|l|c|c|}
        \hline
        \rowcolor{teal!20}
        \textbf{Model Name} & \textbf{Variant/Checkpoint} & \textbf{FPS} & \textbf{Frames} \\
        \hline
        Llava-OV-7B & llava-hf/llava-onevision-qwen2-7b-ov-hf  & 1 & 32 \\
        InternVL2.5-8B & OpenGVLab/InternVL2\_5-8B & 1 & 16 \\
        InternVL2.5-1B & OpenGVLab/InternVL2\_5-1B & 1 & 16 \\
        InternVL2.5-38B & OpenGVLab/InternVL2\_5-38B  & 1 & 16  \\
        MiniCPM-V 2.6 & openbmb/MiniCPM-V-2\_6  & 1 & 64  \\
        Qwen2-VL-7B & Qwen/Qwen2-VL-7B-Instruct  & 2 & 64  \\
        GPT-4o & gpt-4o-2024-08-06  & 1 & 64  \\
        Gemini 1.5 Pro & gemini-1.5-pro-latest  & 1 & 64  \\
        \hline
    \end{tabular}
    }
    \caption{Model Configurations and Video Preprocessing}
    \label{tab:model_configs}
\end{table}



\section{Best and Worst Prompts}
\label{sec:appendix-best_worst}
\begin{figure}[!htbp]
    \centering
    \includegraphics[width=1\columnwidth]{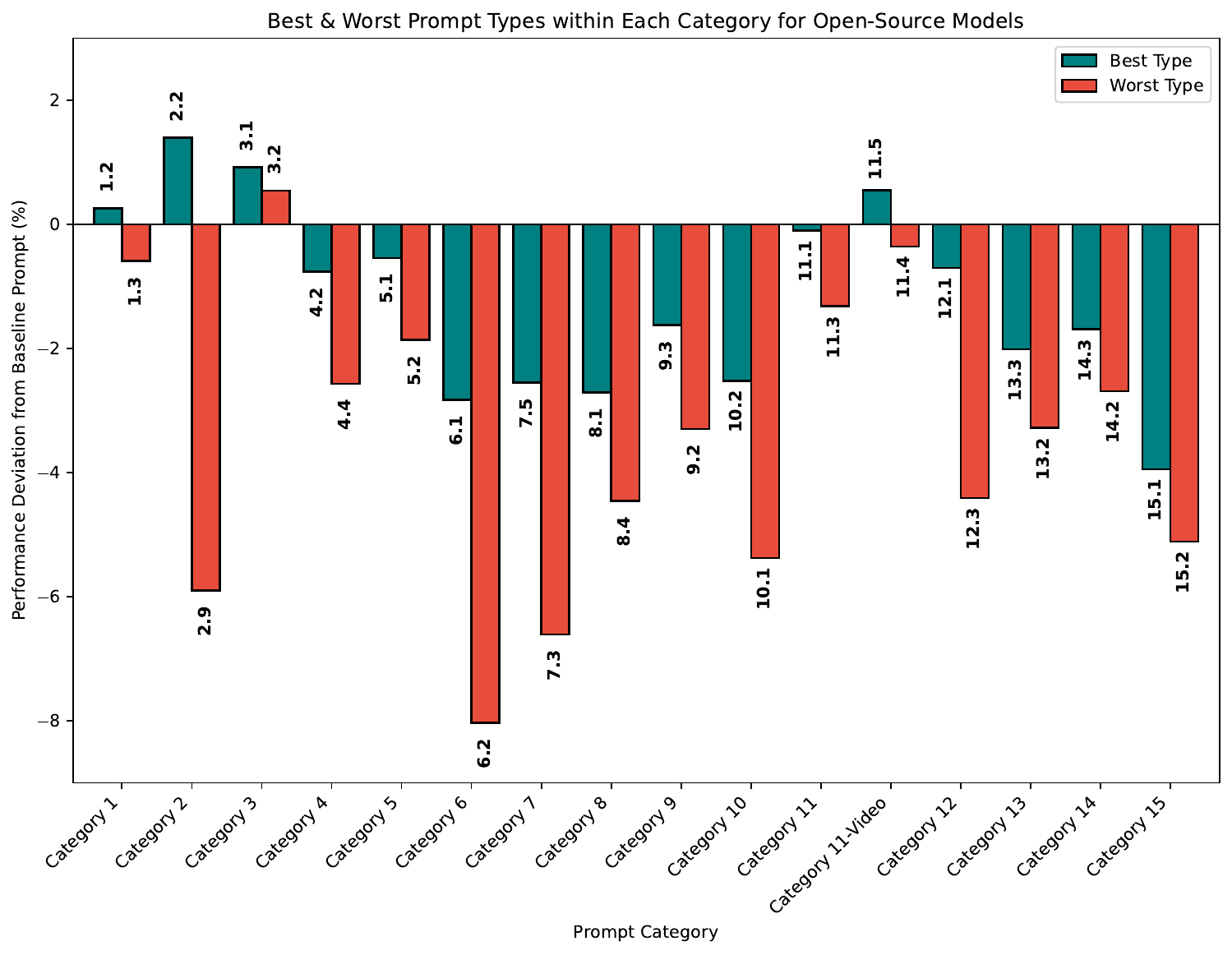}
    \caption{Best \& Worst Prompts within each category for Open-source models. The Deviation of Relative Accuracy (PRAD) with respect to the Baseline Prompt Accuracy is averaged across Open-source Models and the 3 Benchmarks (MMStar, MMMU-Pro \& MVBench) for each Prompt Type.}
    \label{fig:Best-Worst-Open-Baseline}
    \vspace{-1.5em}  
\end{figure}

\begin{figure}[!htbp]
    \centering
    \includegraphics[width=1\columnwidth]{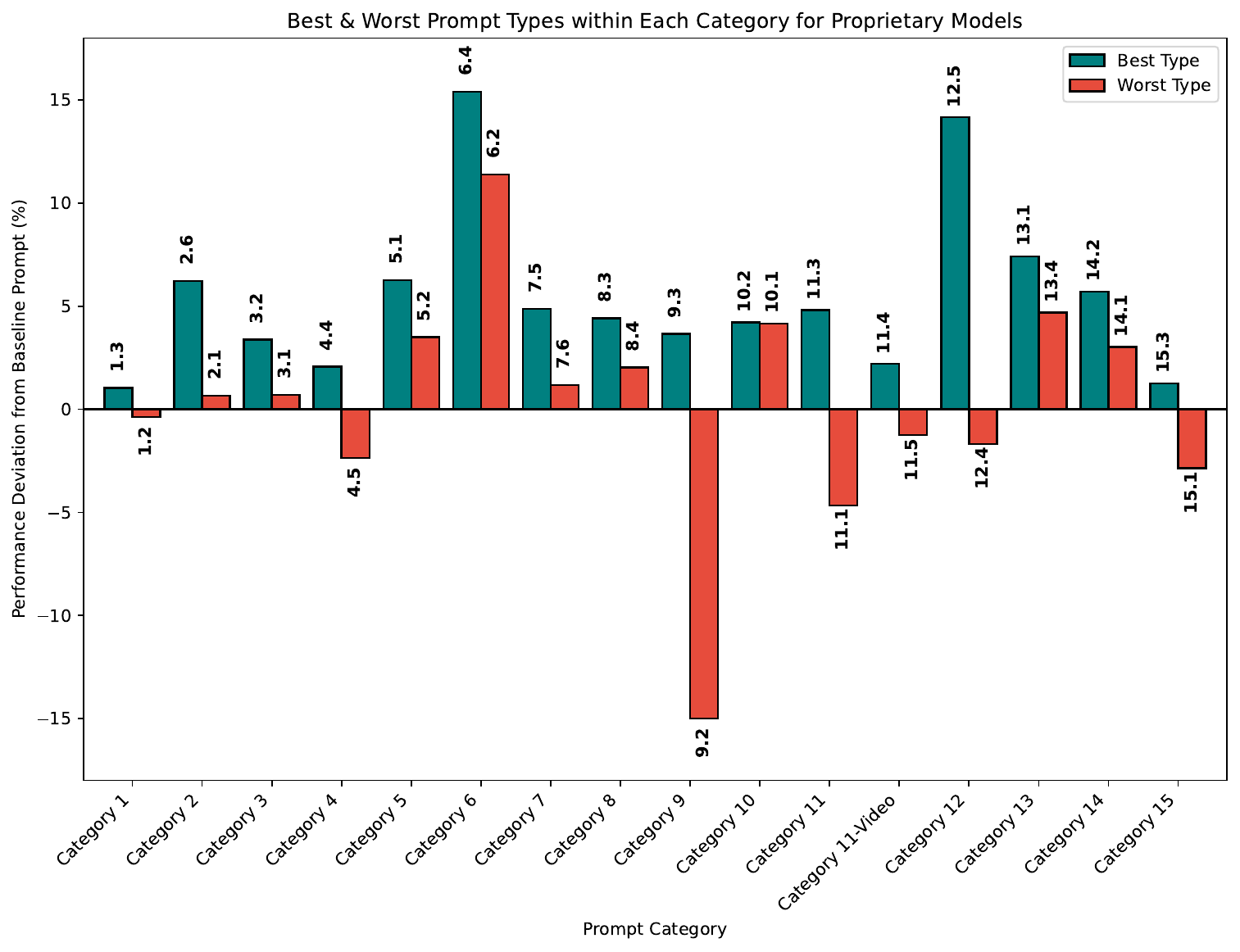}
    \caption{Best \& Worst Prompts within each category for Proprietary models. The Deviation of Relative Accuracy (PRAD) with respect to the Baseline Prompt Accuracy is averaged across Proprietary Models and the 3 Benchmarks (MMStar, MMMU-Pro \& MVBench) for each Prompt Type.}
    \label{fig:Best-Worst-Closed-Baseline}
\end{figure}

The deviation from the baseline (Equation \ref{eq:prad_baseline}) is used to obtain the Figure \ref{fig:Best-Worst-Open-Baseline} \& \ref{fig:Best-Worst-Closed-Baseline}, which highlight the best and worst-performing prompts within each category. 

\section{Which Model is Sensitive?}
\label{sec:appendix-model Sensitivity}

\begin{figure}[!htbp]
\centering
\includegraphics[width=1\columnwidth]{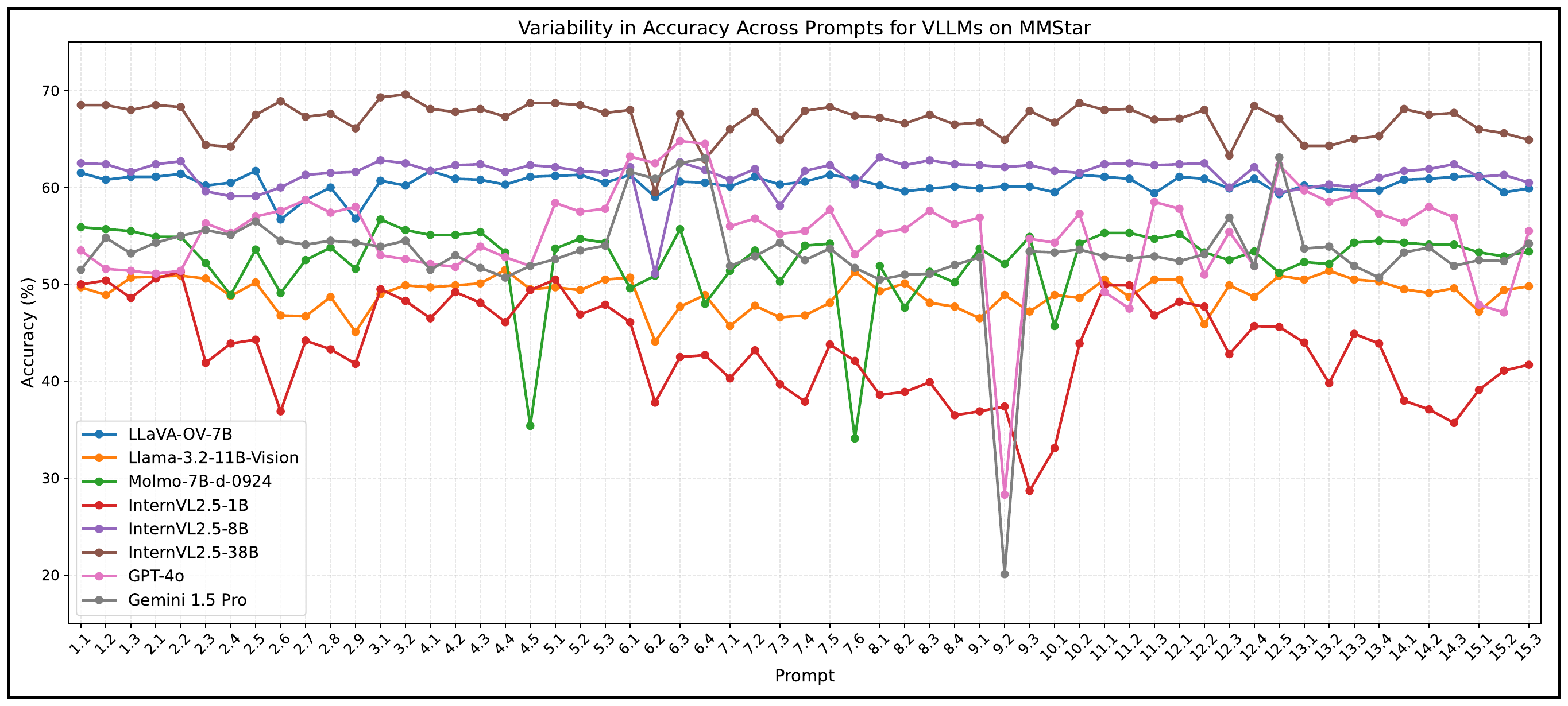}
\caption{Accuracy fluctuations across different textual prompts for models on the MMStar benchmark. These results highlight the varying degrees of prompt sensitivity among models.}
\label{fig:fluctuation line chart MMStar}
\end{figure}

\begin{figure}[!htbp]
\centering
\includegraphics[width=1.05\columnwidth]{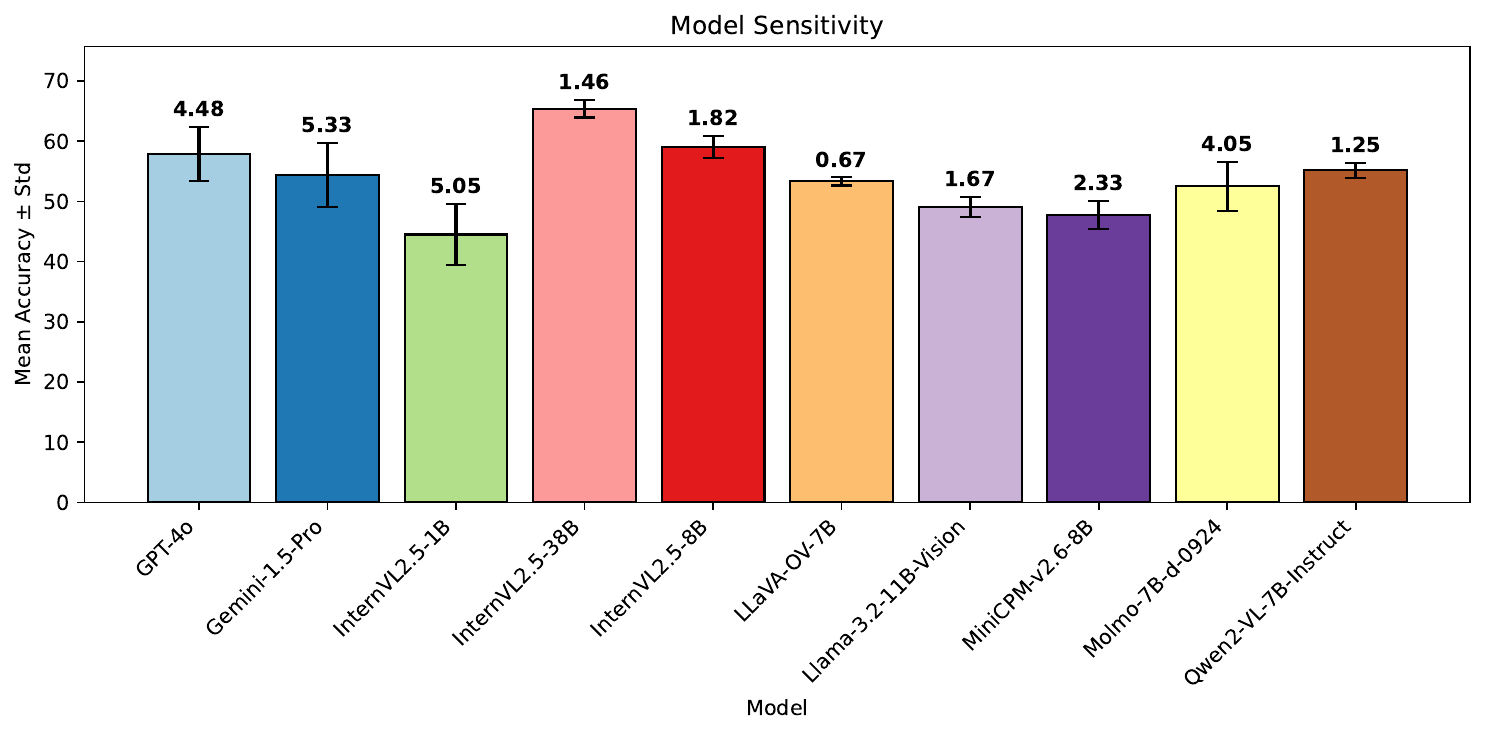}
\caption{Model sensitivity across all prompts, showing mean accuracy and standard deviation averaged over the three benchmarks: MMStar, MMMU-Pro, MVBench.}
\label{fig:model_sensitivity(All)}
\end{figure}

Figure \ref{fig:fluctuation line chart MMStar} illustrates how model performance on the MMStar benchmark fluctuates across different prompt formulations, indicating that some models are more sensitive to prompt changes than others.

To explore this further, Figure \ref{fig:model_sensitivity(All)} summarizes the mean accuracy and standard deviation for each model across all prompts, averaged over the three benchmarks. A higher standard deviation denotes greater prompt sensitivity, as the model's performance varies more significantly depending on the phrasing. Conversely, a lower standard deviation indicates more stable and consistent behavior across prompts.

To better understand model sensitivity, we categorize prompts based on their instructional intent into three groups: positive, neutral, and negative as shown in Section \ref{sec:prompt_cat_sens}.





This classification helps clarify model behavior. Ideally, a model should excel with positive prompts, perform reasonably under neutral conditions, and struggle with negative prompts. Analyzing all prompts together can conflate these effects: an ideal model might exhibit high standard deviation simply due to following expected behaviors across prompt types. Therefore, it is important to evaluate sensitivity within each category.

\begin{figure}[h]
    \centering
    \includegraphics[width=1\columnwidth]{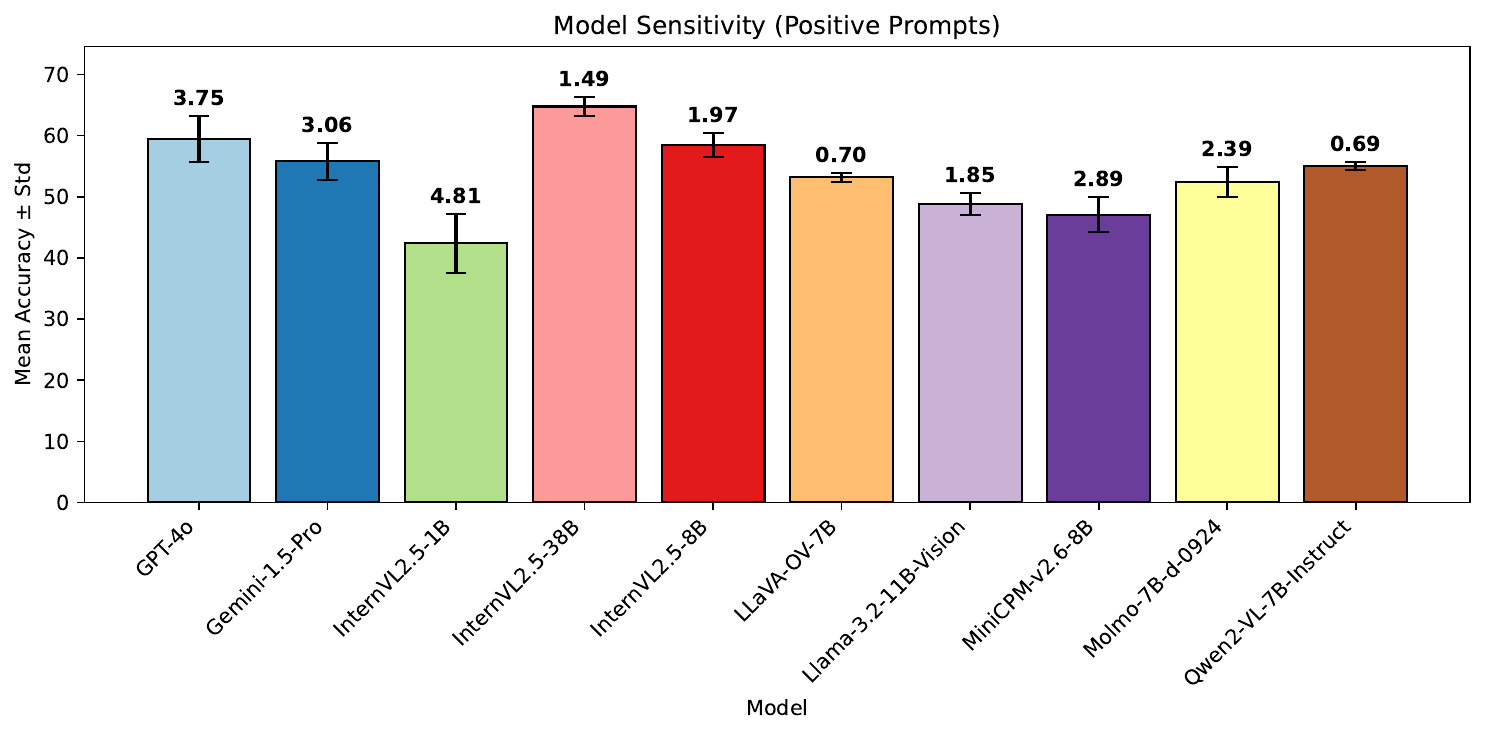} 
    \caption{Model Sensitivity (Positive Prompts). Shows mean accuracy and standard deviation averaged over the three benchmarks: MMStar, MMMU-Pro, MVBench considering only the positive prompts.}
    \label{fig:model_sensitivity(Pos)}
\end{figure}

\begin{figure}[h]
    \centering
    \includegraphics[width=1\columnwidth]{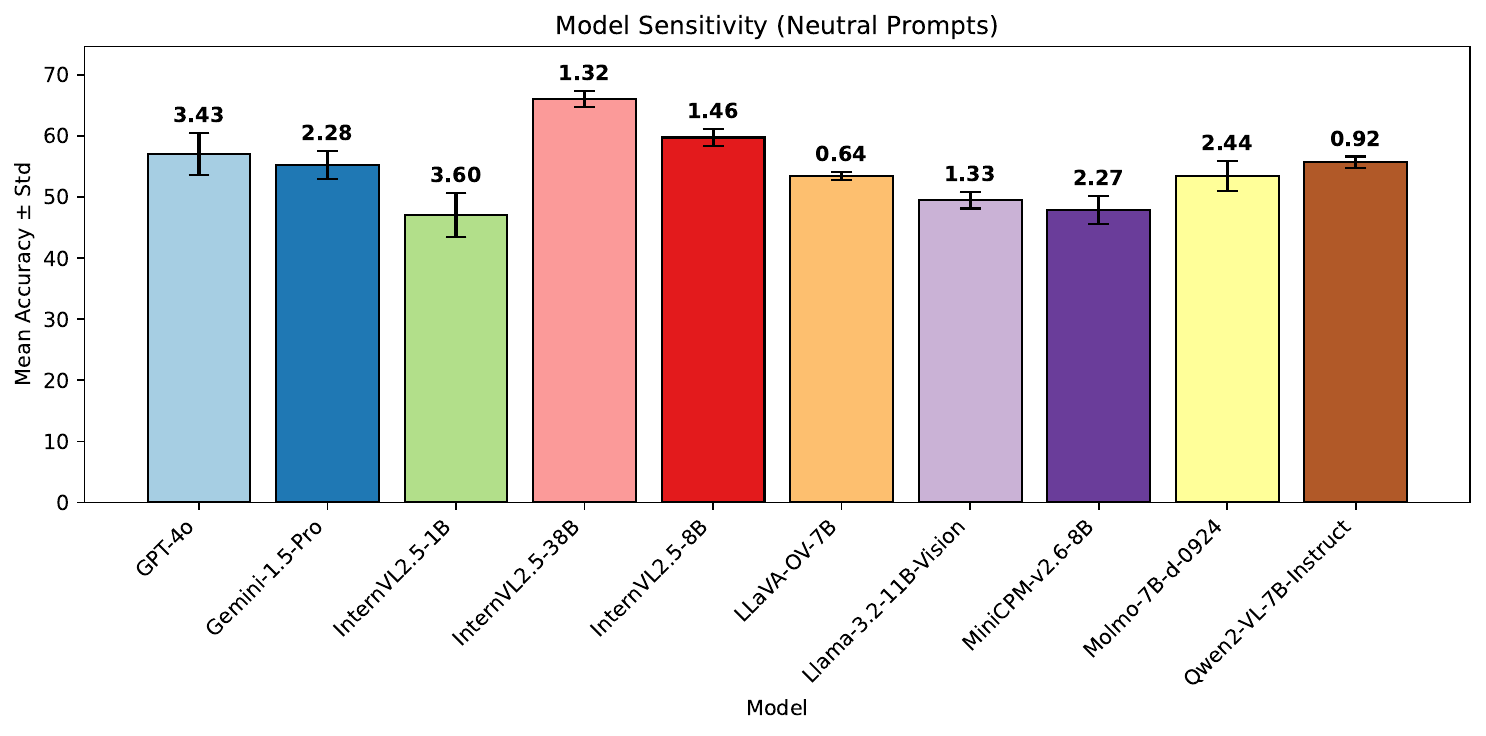} 
    \caption{Model Sensitivity (Neutral Prompts). Shows mean accuracy and standard deviation averaged over the three benchmarks: MMStar, MMMU-Pro, MVBench considering only the neutral prompts.}
    \label{fig:model_sensitivity(Neu)}
\end{figure}

\begin{figure}[h]
    \centering
    \includegraphics[width=1\columnwidth]{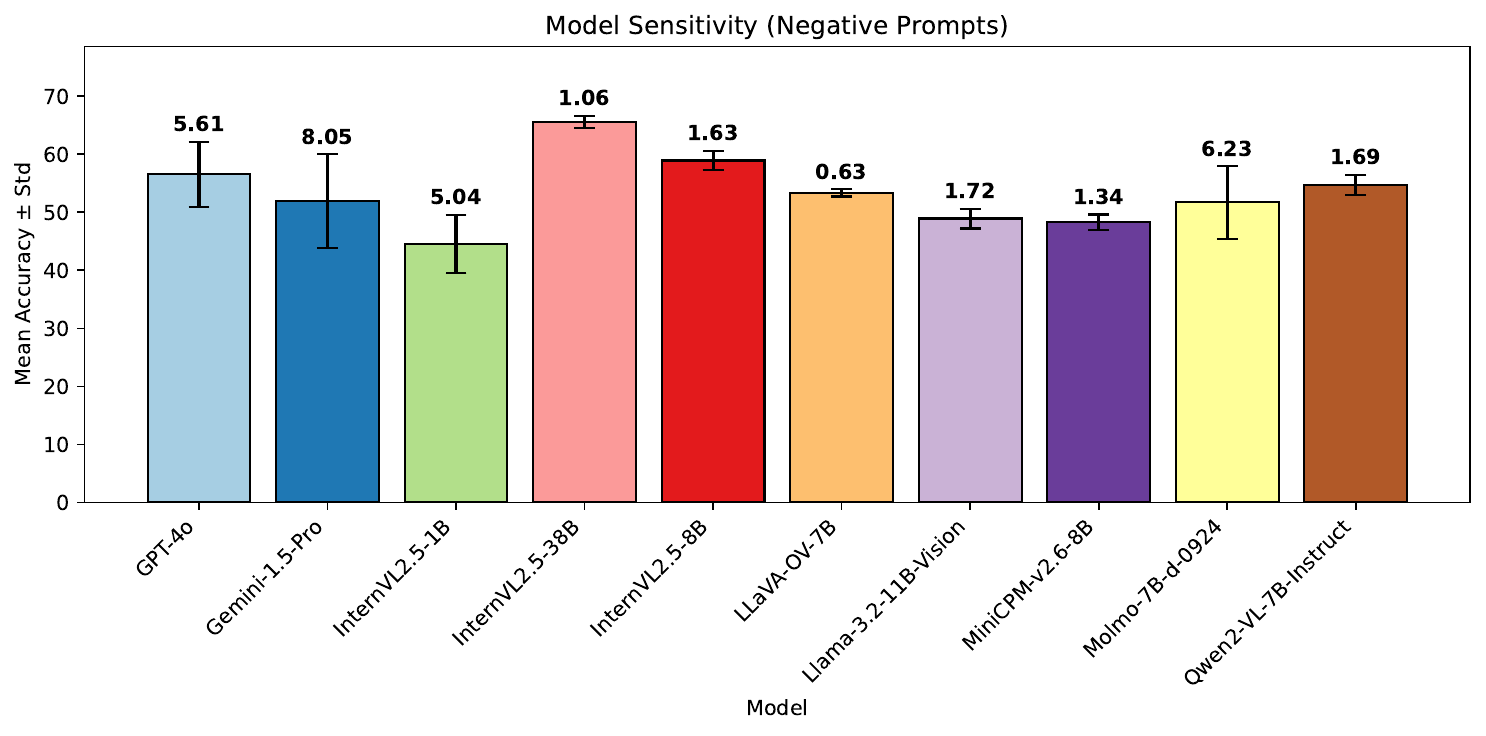} 
    \caption{Model Sensitivity (Negative Prompts). Shows mean accuracy and standard deviation averaged over the three benchmarks: MMStar, MMMU-Pro, MVBench considering only the negative prompts.}
    \label{fig:model_sensitivity(Neg)}
\end{figure}

Interestingly, the observed trend of model sensitivity remained consistent across all categories, Positive (Figure~\ref{fig:model_sensitivity(Pos)}), Neutral (Figure~\ref{fig:model_sensitivity(Neu)}), and Negative (Figure~\ref{fig:model_sensitivity(Neg)}). This indicates that the relative robustness and variability of models are preserved regardless of prompt intent.

Closed-source models, such as GPT-4o and Gemini 1.5 Pro, exhibit higher sensitivity. This could be due to refined instruction tuning, structured optimization for user queries, and meta-prompting mechanisms. These models are fine-tuned to strictly adhere to instructions, making them more responsive to prompt variations and less robust to deviations in phrasing.

Mid-to-large open-source models (7B–38B) demonstrate lower prompt sensitivity. This is beacause they are generally trained with weaker instruction adherence, enabling them to respond more consistently across diverse prompts. Their tendency toward overgeneralization helps mitigate prompt dependency, making them more robust in handling input variations. However, Molmo-7B deviates from this trend, showing higher variability likely due to a lack of fine-tuning on VQA objective and exposure to diverse training tasks such as grounding, which has increased prompt sensitivity.

Smaller open-source models (1B) exhibit greater prompt sensitivity. This could be due to their limited model capacity and weaker context retention abilities. With fewer parameters, these models struggle to generalize effectively, making them highly dependent on structured input formats. While they also exhibit weaker instruction following, their constrained ability to retain context results in higher reactivity to prompt phrasing. Consequently, smaller models show greater fluctuations in performance, reinforcing the trend that model size and instruction fine-tuning influence robustness significantly.

For the open-source case, a comparison of InternVL 1B, 8B, and 38B further supports this trend. The 1B model is highly sensitive to prompt variations, while 8B and 38B exhibit similar levels of sensitivity. This suggests that beyond a certain model size, increasing parameters does not significantly impact prompt stability.

\section{Benchmark Level in-depth Analysis}
\label{sec:appendix-BM Level Analysis}

\subsection{Which Benchmark is Sensitive?}

\begin{figure}[h]
    \centering
    \includegraphics[width=0.8\columnwidth]{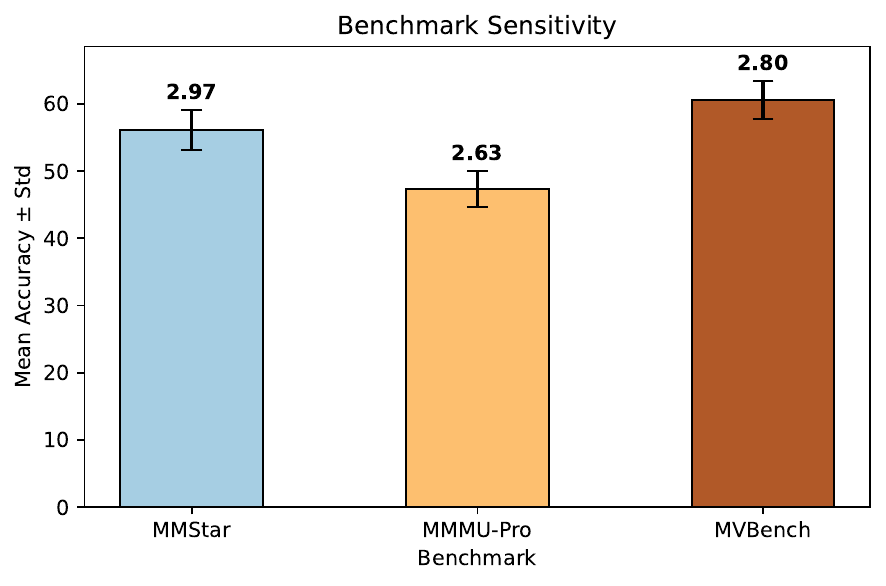} 
    \caption{Benchmark Sensitivity}
    \label{fig:BM_sensitivity}
\end{figure}

The bar chart (Figure \ref{fig:BM_sensitivity}) presents the mean accuracy and standard deviation of model performance across prompts, averaged per benchmark. The benchmarks MMStar, MMMU-Pro, and MVBench exhibit comparable levels of sensitivity, with MVBench showing the highest variability. This suggests that models experience similar fluctuations in performance across these benchmarks, implying no single benchmark is significantly more robust than the others. The slight differences indicate that while all benchmarks maintain a consistent evaluation framework, some may introduce more variability in responses due to task diversity or complexity.

\subsection{Which Question Type in MMMU-Pro is Sensitive?}

\begin{figure}[!htbp]
    \centering
    \includegraphics[width=1\columnwidth]{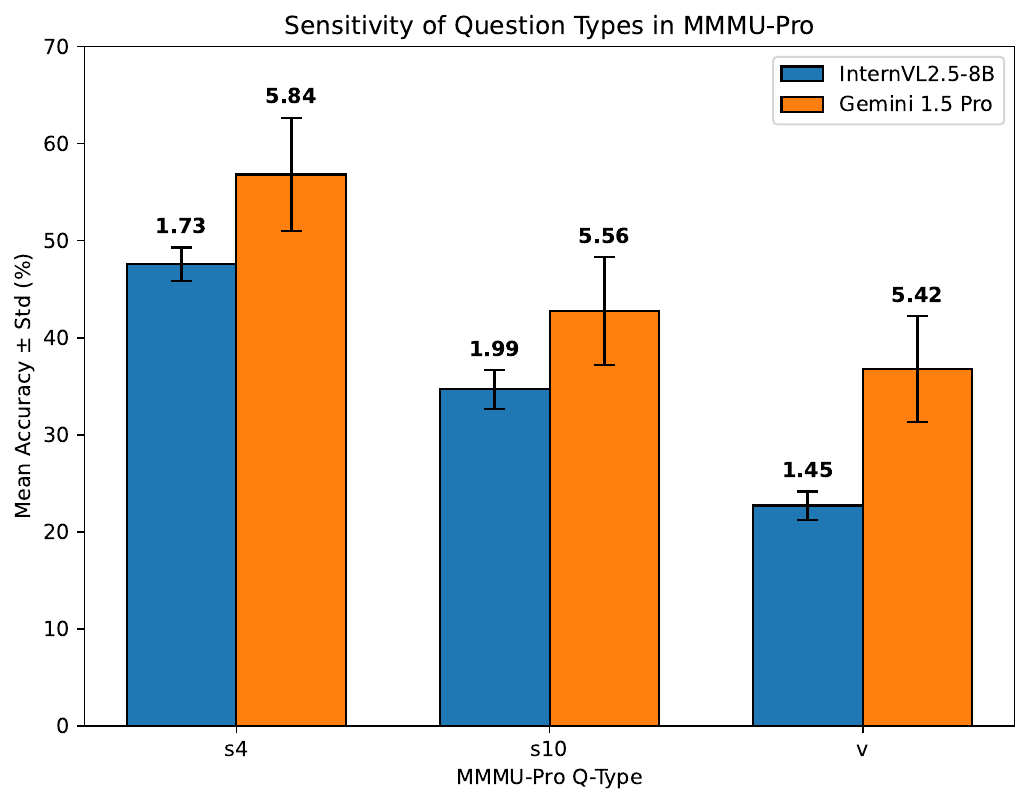} 
    \caption{Sensitivity of Question Types in MMMU-Pro. The Figure represents the Average Accuracy and Standard Deviation averaged across the models.}
    \label{fig:Q-type-Std}
\end{figure}

The MMMU-Pro benchmark introduces three distinct types of multiple-choice questions: (1) S4, the standard format with four answer choices; (2) S10, an extended version with ten answer choices; and (3) V, a vision-based setting where the question is embedded within an image, with no explicit text input provided to the model.

Among these, the S4 setting achieves the highest accuracy (Figure \ref{fig:Q-type-Std}). Both S4 and S10 demonstrate comparable levels of robustness, as indicated by their similar standard deviations, suggesting that increasing the number of answer choices does not significantly impact robustness. In contrast, the V setting, despite yielding the lowest accuracy, exhibits the highest robustness. This indicates that while this setting pose greater challenges for models, their performance remains relatively stable across different prompts.

\subsection{Which Benchmark in MMStar is Sensitive?}

\begin{figure}[!htbp]
    \centering
    \includegraphics[width=1\columnwidth]{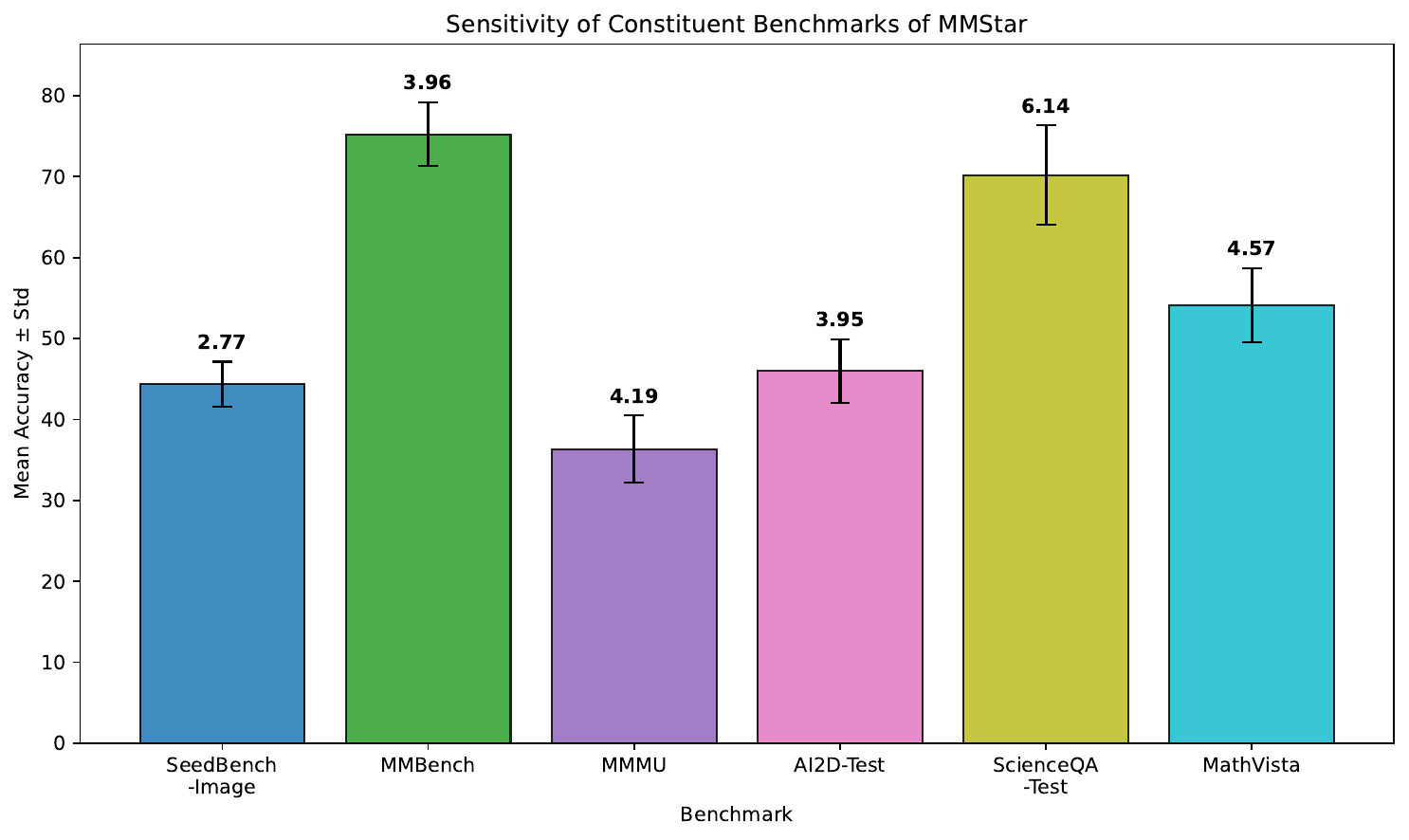} 
    \caption{Sensitivity of Constituent Benchmarks of MMStar. The Figure shows Average Accuracy and Standard Deviation averaged across Models.}
    \label{fig:MMStar-BM}
\end{figure}

MMStar is constructed by aggregating a subset of questions from six existing benchmarks. Among these, ScienceQA-Test \cite{lu2022learnexplainmultimodalreasoning} exhibits the highest sensitivity to prompt variations, while SeedBench-Image \cite{li2023seedbenchbenchmarkingmultimodalllms} demonstrates the least (Figure \ref{fig:MMStar-BM}). The remaining four benchmarks, MMBench \cite{liu2024mmbenchmultimodalmodelallaround}, MMMU \cite{yue2024mmmumassivemultidisciplinemultimodal}, AI2D-Test \cite{kembhavi2016diagramworthdozenimages}, and MathVista \cite{lu2024mathvistaevaluatingmathematicalreasoning}, display comparable levels of sensitivity.

\subsection{Which Core Capability in MMStar is Sensitive?}

\begin{figure}[!htbp]
    \centering
    \includegraphics[width=1\columnwidth]{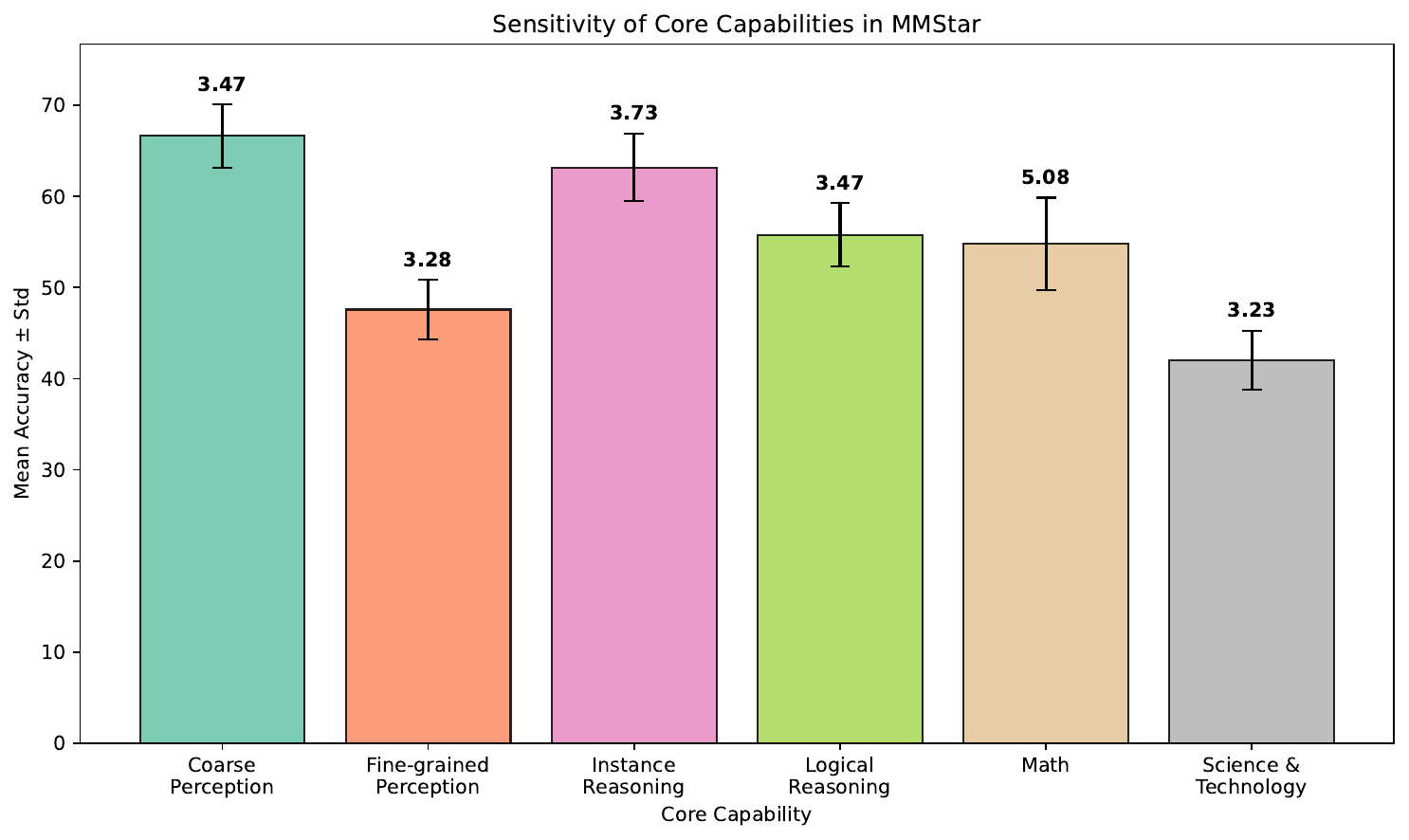} 
    \caption{Sensitivity of Core Capabilities of MMStar. The Figure shows Average Accuracy and Standard Deviation averaged across Models.}
    \label{fig:MMStar CC}
\end{figure}

MMStar evaluates six core capabilities: Coarse Perception, Fine-grained Perception, Instance Reasoning, Logical Reasoning, Math, and Science \& Technology. Among these, Math exhibits the highest sensitivity to prompt variations. The remaining five capabilities show comparable levels of sensitivity. (Figure \ref{fig:MMStar CC})

\subsection{Which Subject in MMMU-Pro is Sensitive?}

\begin{figure}[!htbp]
    \centering
    \includegraphics[width=1\columnwidth]{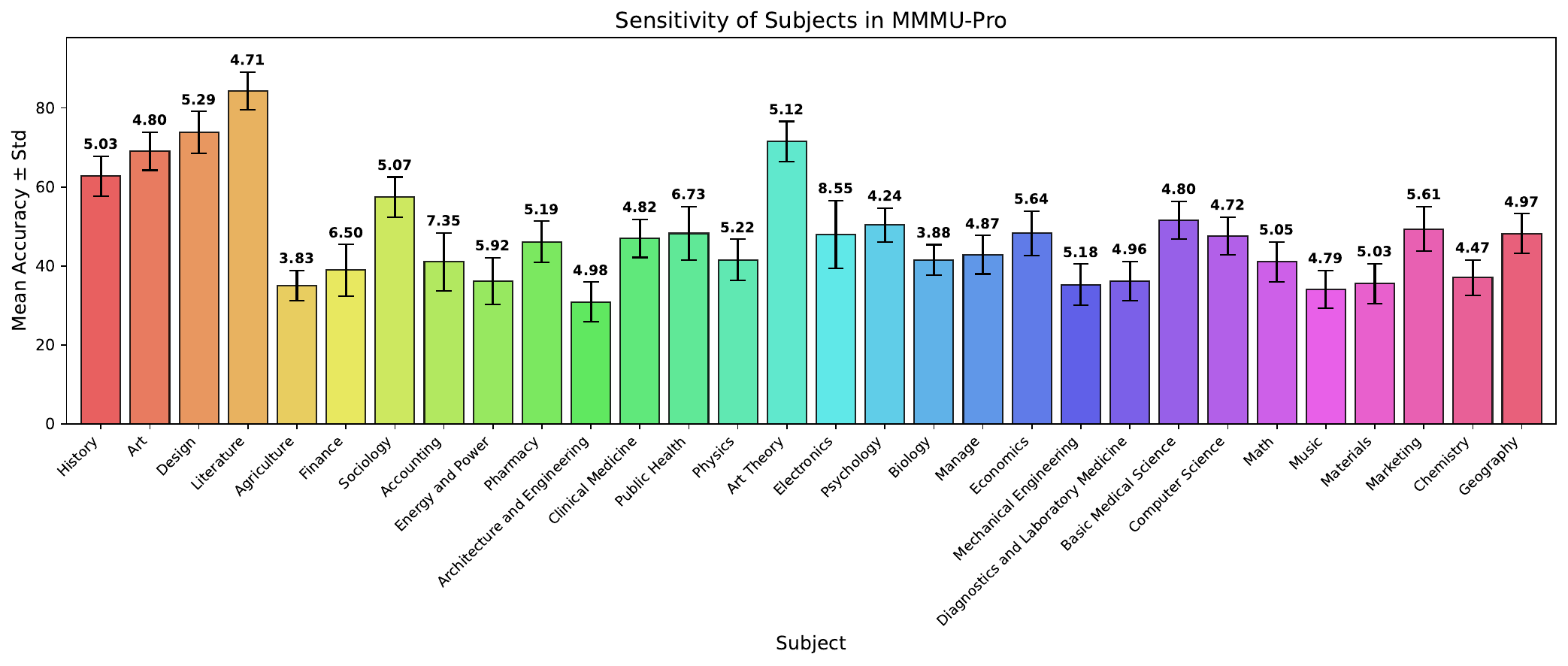} 
    \caption{Sensitivity of Subjects in MMMU-Pro. The Figure shows Average Accuracy and Standard Deviation averaged across Models.}
    \label{fig:MMMU-Pro types}
\end{figure}

The sensitivity analysis of subjects in MMMU-Pro (Figure \ref{fig:MMMU-Pro types}) reveals that Electronics exhibits the highest variation in performance across different prompts, followed by Accounting, Public Health, Finance, and Energy and Power. These subjects are more susceptible to changes in prompt phrasing, indicating a higher reliance on specific wording for model accuracy. In contrast, Management, Biology, Economics, Architecture and Engineering, and Clinical Medicine show the least sensitivity, suggesting that prompt variations have a minimal effect on model performance in these domains.

\subsection{Which Temporal Task in MVBench is Sensitive?}

\begin{figure}[!htbp]
    \centering
    \includegraphics[width=1\columnwidth]{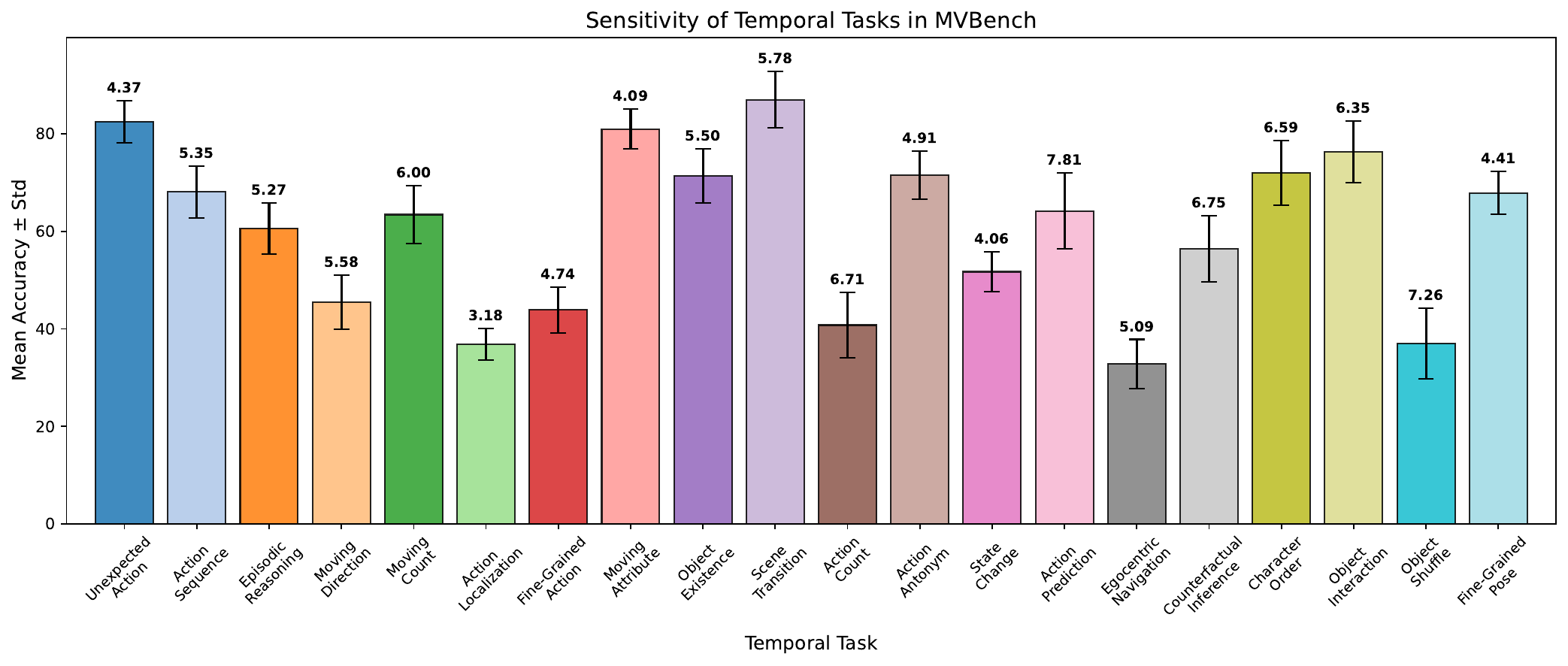} 
    \caption{Sensitivity of Temporal Tasks in MVBench. The Figure shows Average Accuracy and Standard Deviation averaged across Models.}
    \label{fig:MVBench types}
\end{figure}

The sensitivity analysis of temporal tasks in MVBench (Figure \ref{fig:MVBench types}) reveals that Action Prediction exhibits the highest variation in performance across different prompts, followed by Object Shuffle, Action Count, Counterfactual Inference, and Character Order. These tasks are particularly sensitive to prompt phrasing, suggesting that slight modifications in wording significantly impact model responses. In contrast, State Change, Moving Attribute, Unexpected Action, Fine-Grained Pose, and Fine-Grained Action demonstrate the least sensitivity, indicating more stable performance across different prompt formulations. These findings provide insights into which temporal reasoning tasks require more careful prompt engineering for consistent model evaluation.

\section{Model-Specific Anomalies}
\label{subsec:Model specific interesting cases}

\begin{figure}[!htbp]
    \centering
    \includegraphics[width=1\columnwidth]{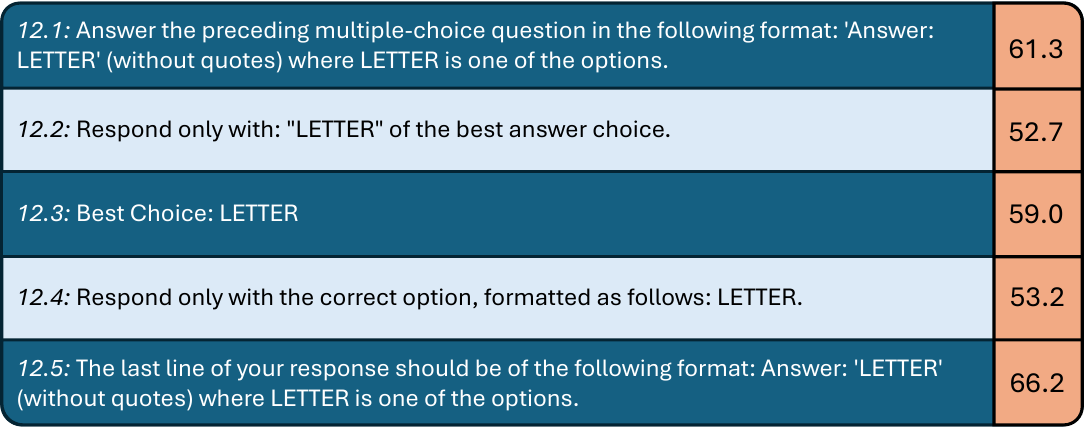} 
    \caption{GPT-4o Performance (Absolute Accuracy) for Category-12 Prompts on MMMU-Pro. This shows how even the slightest difference in how the answer is expected can result in significant fluctuations in performance.}
    \label{fig:Cat12-GPT4o}
\end{figure}

This study yielded some surprising findings. Notably, a straightforward variation in phrasing for Category 12 led to accuracy shifts of up to 15\% in proprietary models (Figure \ref{fig:Cat12-GPT4o}). Additionally, Prompt 12.5 performed comparably to the Chain-of-Thought (CoT) prompt. Category 6 CoT prompts led to an approximate 10\% accuracy improvement for proprietary models, while no noticeable gains were observed for open-source models. Another noteworthy finding is that the "Act as a Computer Vision Professor" prompt (9.1) resulted in a slight accuracy decrease, whereas the "Act as a Careless Student" prompt (9.2) caused a dramatic 40\% drop in accuracy for Gemini 1.5 Pro on MMMU-Pro. This pattern was consistently observed across all datasets for both GPT-4o and Gemini 1.5 Pro.

Unexpected accuracy drops were observed in open-source models, as shown in Table \ref{tab:open_accuracy_drop}. The table presents the absolute accuracy drop relative to the baseline, along with the corresponding model responses for each instance. Another notable observation was that when the prompt included the \$ symbol (e.g. 12.3: Best Choice: \$LETTER), GPT-4o more frequently refused to respond due to safety concerns, often generating disclaimers instead of valid answers. Consequently, the \$ symbol was omitted from all prompts for GPT-4o.

\begin{table}[t]
\centering
\small
\resizebox{\columnwidth}{!}{%
\begin{tabular}{|l|l|l|c|>{\raggedright\ttfamily\arraybackslash}p{3.2cm}|}
    \hline
    \rowcolor{teal!20}
    \multicolumn{1}{|c|}{\textbf{Dataset}} &
    \multicolumn{1}{c|}{\textbf{Type}} &
    \multicolumn{1}{c|}{\textbf{Model}} &
    \multicolumn{1}{c|}{\textbf{\(\Delta\) Accuracy}} &
    \multicolumn{1}{c|}{\textbf{Response}} \\
    \hline\hline

    \rowcolor{gray!10}
    MMStar     & Type6.2  & MiniCPM-v2.6-8B  & -12.6 & \texttt{"\$LETTER"} \\
    \rowcolor{gray!10}
               & Type4.5  & Molmo-7B-d-0924  & -20.5 & \texttt{"A/A/A/A"} \\
    \rowcolor{gray!10}
               & Type7.6  & Molmo-7B-d-0924  & -21.8 & \texttt{"E" or "F"} \\
    \rowcolor{gray!10}
               & Type9.3  & InternVL2.5-1B   & -21.3 & \texttt{"\$LETTER"} \\
    \rowcolor{gray!10}
               & Type10.1 & InternVL2.5-1B   & -16.9 & \texttt{"\$LETTER"} \\
    \hline

    \rowcolor{cyan!10}
    MMMU-Pro   & Type7.6  & InternVL2.5-1B   & -10.9 & \texttt{"E" or "F" or "G"} \\
    \rowcolor{cyan!10}
               & Type9.3  & InternVL2.5-1B   & -17.2 & \texttt{"\$LETTER"} \\
    \rowcolor{cyan!10}
               & Type10.1 & InternVL2.5-1B   & -11.2 & \texttt{"\$LETTER"} \\
    \hline

    \rowcolor{orange!10}
    MVBench    & Type6.2  & MiniCPM-v2.6-8B  & -14.2 & \texttt{"\$ LETTER"} \\
    \rowcolor{orange!10}
               & Type6.2  & InternVL2.5-1B   & -20.8 & \texttt{"\$ LETTER"} \\
    \rowcolor{orange!10}
               & Type11.3 & InternVL2.5-1B   & -15.7 & \texttt{"\$NON-NEGOTIABLE", "\$ERROR", "\$NON\_EXISTENT"} \\
    \rowcolor{orange!10}
               & Type13.1 & InternVL2.5-1B   & -23.6 & \texttt{"Answer: \$LETTER"} \\
    \rowcolor{orange!10}
               & Type13.3 & InternVL2.5-1B   & -19.6 & \texttt{"Incorrect"} \\
    \rowcolor{orange!10}
               & Type13.4 & InternVL2.5-1B   & -23.1 & \texttt{"Answer: \$LETTER. Deviations from this format will result in automatic deductions"} \\
    \hline
\end{tabular}%
} 
\caption{Instances of Significant Accuracy Drops and Corresponding Model Responses}
\label{tab:open_accuracy_drop}
\end{table}


\normalsize

\newpage

\section{Distribution of Standard Deviation within Prompt Categories}
\label{fig:std dist}

\begin{figure}[H]
    \centering
    \includegraphics[width=1\columnwidth]{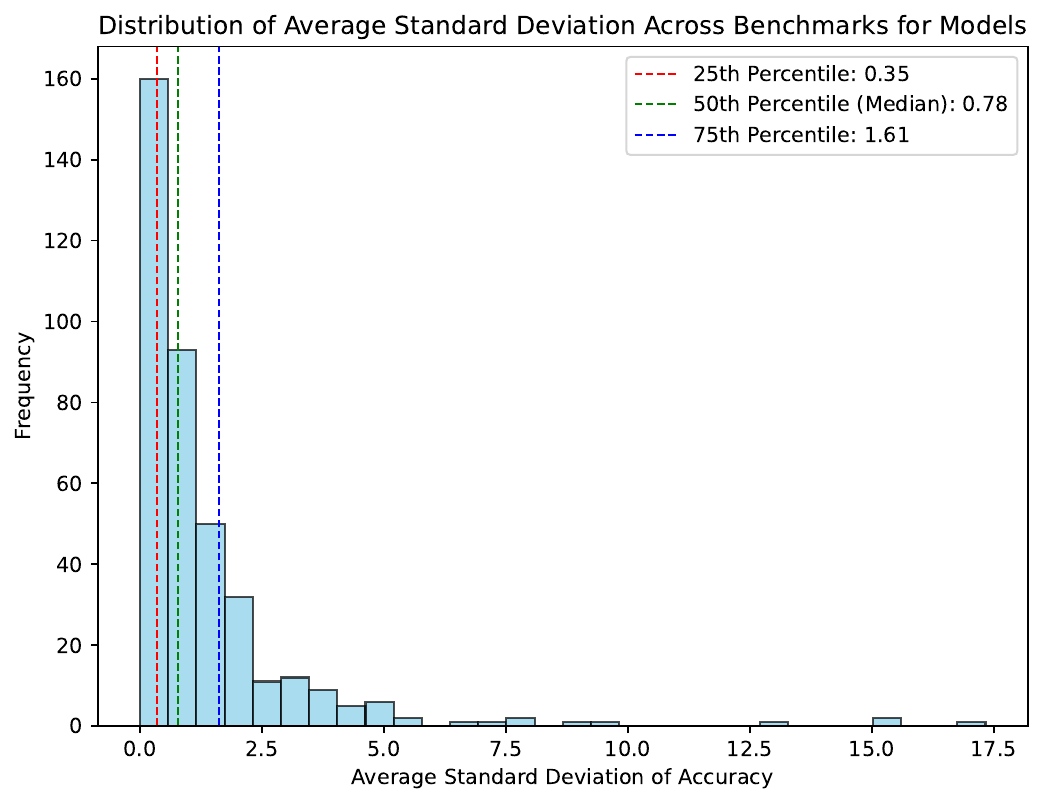} 
    \caption{Distribution of standard deviation values computed across prompt categories and models. Each value represents the variability in accuracy within a single prompt category for a given model. The aggregated distribution is used to define a threshold for high sensitivity, with the median standard deviation of 0.78 serving as the cutoff between low- and high-sensitivity prompt categories.}
    \label{fig:std-dist}
\end{figure}

\end{document}